\documentclass[JCCE]{BVP}

\usepackage{microtype}
\usepackage{graphicx}
\usepackage{booktabs} 
\usepackage{amsmath}
\usepackage{amssymb}
\usepackage{amsthm}
\usepackage{acronym}
\usepackage{enumitem}
\usepackage{multirow,soul}
\usepackage{soul}
\usepackage{subfig}
\usepackage{url}            
\usepackage{amsfonts}       
\usepackage{nicefrac}       
\usepackage{xcolor}         
\usepackage{nccmath}
\usepackage{array}
\usepackage{tcolorbox}
\usepackage{float}
\usepackage{hyperref}
\usepackage{algpseudocode,algorithm,algorithmicx}

\makeatletter
\long\def\@makecaption#1#2{%
	\vskip\abovecaptionskip
	{\centering
		\figcaptionnumfont #1\par\nobreak
		\figcaptionfont #2\par}%
	\vskip\belowcaptionskip}
\makeatother

\newtheorem{assumption}{Assumption}
 
\theoremstyle{remark}
\newtheorem{remark}{Remark}

\hypersetup{
	colorlinks=true,
	urlcolor=purple,
	linkcolor=blue,
	citecolor=green
}

\makeatletter
\g@addto@macro\UrlBreaks{%
	\do\a\do\b\do\c\do\d\do\e\do\f\do\g\do\h\do\i\do\j\do\k\do\l\do\m%
	\do\n\do\o\do\p\do\q\do\r\do\s\do\t\do\u\do\v\do\w\do\x\do\y\do\z%
	\do\A\do\B\do\C\do\D\do\E\do\F\do\G\do\H\do\I\do\J\do\K\do\L\do\M%
	\do\N\do\O\do\P\do\Q\do\R\do\S\do\T\do\U\do\V\do\W\do\X\do\Y\do\Z%
	\do\0\do\1\do\2\do\3\do\4\do\5\do\6\do\7\do\8\do\9\do\-\do\.\do\_}
\makeatother
\makeatletter
\newif\iftablefootnotenone
\newif\ifkilltableabovespace
\newcount\tfootcount
\providecommand\appendix{}
\renewcommand\appendix{\par
	\setcounter{section}{0}%
	\setcounter{subsection}{0}%
	\gdef\thesection{\@Alph\c@section}%
	\gdef\theHsection{Appendix.\@Alph\c@section}}
\makeatother
\usepackage[capitalize]{cleveref}
\crefname{section}{Sec.}{Secs.}
\Crefname{section}{Section}{Sections}
\Crefname{table}{Table}{Tables}
\crefname{table}{Tab.}{Tabs.}
\crefname{appendix}{App.}{Apps.}

\newcommand{\stateart}{state-of-the-art }
\newcommand{\etal}{\textit{et al. }}
\newcommand\tab[1][0.3cm]{\hspace*{#1}}

\newacro{dl}[DL]{deep learning}
\newacro{ae}[AE]{adversarial example}
\newacro{dnn}[DNN]{deep neural network}
\newacro{cnn}[CNN]{convolutional neural network}
\newacro{nn}[NN]{neural network}
\newacro{pgd}[PGD]{projected gradient descent}
\newacro{fgsm}[FGSM]{fast gradient sign attack}
\newacro{at}[AT]{adversarial training}
\newacro{nt}[NT]{natural training}
\newacro{fl}[FL]{federated learning}
\newacro{fat}[FAT]{federated adversarial training}
\newacro{vit}[ViT]{vision transformer}
\newacro{niid}[Non-IID]{not independent and identically distributed}
\newacro{iid}[IID]{independent and identically distributed}
\newacro{ae}[AE]{adversarial example}
\newacro{t2t}[T2T-ViT]{tokens-to-token ViT}
\newacro{tnt}[TNT]{transformer-in-transformer}
\newacro{scaffold}[SCAFFOLD]{stochastic controlled averaging}
\newacro{afl}[AFL]{agnostic federated learning}
\newacro{ffl}[FFL]{fair federated learning}
\newacro{nlp}[NLP]{natural language processing}
\newacro{cvt}[CvT]{convolutional vision transformer}
\newacro{cpvt}[CPVT]{conditional position encoding vision transformer}
\newacro{sovit}[So-ViT]{second-order vision transformer}
\newacro{iot}[IoT]{internet of things}
\newacro{sgd}[SGD]{stochastic gradient descent}
\newacro{msa}[MSA]{multiheaded self-attention}
\newacro{mlp}[MLP]{multi-layer perceptron}
\newacro{gelu}[GELU]{Gaussian error linear units}
\newacro{svcca}[SV-CCA]{singular vector canonical correlation analysis}

\algrenewcommand\algorithmicrequire{\textbf{Input:}}
\algrenewcommand\algorithmicensure{\textbf{Output:}}
\newcommand*\Let[2]{\State #1 $\gets$ #2}
\algdef{SE}[SUBALG]{Indent}{EndIndent}{}{\algorithmicend\ }%
\algtext*{Indent}
\algtext*{EndIndent}

\received{d month yyyy}
\revised{d month yyyy}
\accepted{d month yyyy}
\published{d month yyyy}
\doi{10.47852/bonviewCINCXXXXXXXX}

\OAText{{\copyright} The Author(s) 2026. Published by BON VIEW PUBLISHING PTE. LTD. This is an open access article under the CC BY License (https://creativecommons.org/licenses/by/4.0/).}

\begin{document}  %
	
	\arttype{Research Article}
	
	\title{Federated Adversarial Training with Transformers}
	
	\author{Ahmed Aldahdooh$^1$\thanks{\textbf{Corresponding author:} Ahmed Aldahdooh, IT Department, University College of Applied Sciences, Gaza, Palestine. Email: \href{mailto:adahdooh@ucas.edu.ps}{adahdooh@ucas.edu.ps}}, Wassim Hamidouche$^2$, and Olivier D\'eforges$^2$}
	\affil{IT Department, University College of Applied Sciences, Palestine.}
	\affil{Univ Rennes, INSA Rennes, CNRS, IETR -- UMR 6164, France.\vspace{-5mm}}

	\begin{abstract}
		\Ac{fl} trains models on distributed client data while preserving privacy; however, the resulting models remain vulnerable to \acp{ae}. \Ac{at}, the most promising defense, has been extensively examined for \acp{cnn}. Nevertheless, its performance in federated learning environments using vision transformers remains largely unexamined. This study explores \ac{fat} for vision transformers on CIFAR-10 under IID and two label-skewed Non-IID partitions. We evaluate 12 model variants covering three tokenization schemes and three classification-head architectures, presenting three main findings. First, FedProx, FedGate, and SCAFFOLD, which are designed to reduce client drift under Non-IID data, achieve lower robust accuracy than standard FedAvg when used with a vision transformer backbone. They also require up to twice the per-round communication and maintain a model-sized client state. Second, tokenization and classification-head design strongly shape the trade-off between clean and robust accuracy. Token-to-token embedding improves clean accuracy but remains vulnerable to projected gradient descent (PGD) attacks, whereas heads that rely solely on visual tokens are poorly suited to FAT. Third, we introduce FedWAvg, which weights client updates according to the cosine similarity between each client’s final layer and the global model’s final layer. We prove FedWAvg is exactly a Gibbs reweighting of clients by the squared distance between their direction-normalized final layers, and hence never increases the aggregate directional drift of the classification head relative to FedAvg, with the reduction given in closed form. Experiments show that FedWAvg improves robust accuracy under highly skewed data.
	\end{abstract}
	
	\begin{keywords}
		Federated Learning, Adversarial Training, Robustness, Vision Transformer, Aggregation\vspace{-1mm}
	\end{keywords}
	\maketitle
	
	\vspace{-2mm}
	\section{Introduction}
	\label{sec:intro}
	
	\Ac{fl} \cite{mcmahan@2017@communication} is an emerging learning paradigm for training a global model. It works by aggregating parameters over multiple communication rounds from models trained on private data distributed across multiple clients, all while preserving data privacy. In each communication round, clients are required to send only their local model parameters to the server, not the data, so that the global model can be built by aggregating the local models.
	Since \ac{fl} minimizes the risk of sensitive and private data leakage, industry found \ac{fl} is useful in healthcare \cite{nguyen@2022@healthcare}, natural language processing, edge devices and \ac{iot}, wireless communications, smart cities, and for other applications \cite{yurdem@2024@overview}.

	\Ac{fl}, like any emerging technique, faces significant challenges that must be addressed before being ready for deployment in real-world applications. Client model drift\footnote{The phenomenon of model drift occurs when models learn different representations of the same data.} and model convergence are the main challenges of \ac{fl} on \ac{niid} data.
	To resolve these challenges, two approaches are identified. The first one is to optimize the \ac{fl} aggregation methods such as FedAvg \cite{mcmahan@2017@communication}, FedProx \cite{li@2020@optimization}, FedGate \cite{haddadpour@2021@compression}, FairFed \cite{ezzeldin@2023@fairfed}, FedSL \cite{zhang@2024@fedsl}, and SCAFFOLD \cite{karimireddy@2020@scaffold}. For instance, SCAFFOLD \cite{karimireddy@2020@scaffold} was shown to have better convergence than FedAvg, and FedProx. The second approach is to carefully choose the model architecture \cite{shah@2021@adversarial}, \cite{qu@2022@rethinking}. For instance, in \cite{shah@2021@adversarial}, VGG-9 model was found to yield higher performance than Network-in-Network model on CIFAR-10 dataset. Moreover, in \cite{qu@2022@rethinking}, \ac{vit} \cite{dosovitskiy@2021@image} was shown to be significantly effective than ResNet50 model on \ac{niid} CIFAR-10 dataset.

	On the other hand, \ac{fl} is vulnerable to attacks that may put the model and the data at risk \cite{kairouz@2021@advances}, \cite{feng@2025@threats}, \cite{jimenezgutierrez@2026@security}, \cite{almutairi@2023@vulnerabilities}. Attacks happen either during the training phase, and it is categorized into \textit{poisoning attacks} and \textit{inference attacks}, or during inference/testing time, where they are known as \textit{evasion attacks} \cite{akhtar@2021@attacks}. To defend against evasion attacks, many defense approaches have been proposed in the literature \cite{akhtar@2021@attacks}, \cite{aldahdooh@2022@detection}. Adversarial training is one of the most effective defense strategies \cite{bai@2021@recent}, \cite{madry@2018@towards}. It re-trains the model by incorporating the \acp{ae} into the training process. In each training iteration, the \acp{ae} are generated using the current state of the model. In practice, any algorithm can be used to generate the \acp{ae}, such as \ac{fgsm}, and \ac{pgd} \cite{madry@2018@towards}.

	To the best of the authors' knowledge, no existing publication has examined vision transformer models within the context of federated adversarial training (FAT). Two published studies highlight the importance of addressing this gap and collectively suggest the expected outcomes of such an approach. Concerning federated learning, Qu et al. \cite{qu@2022@rethinking} demonstrate that Vision Transformers (ViTs) converge significantly faster than ResNet-50 and attain superior global model performance when employing Non-IID CIFAR-10 in a natural training scenario. This underscores the significance of model architecture as a critical factor in the overall federated learning process. Regarding adversarial robustness, Bai et al. \cite{bai@2021@recent} conduct a centralized comparison, matching both models in capacity and training procedures, and observe that following adversarial training, ResNet-50 and DeiT-S exhibit comparable robustness against PGD and AutoAttack. Therefore, transformers are not inherently more robust than CNNs when trained identically. Consequently, any advantages of vision transformers in FAT must originate from the federated setting itself—specifically, from accelerated convergence and reduced drift under heterogeneous conditions, rather than from intrinsic architectural robustness. The pivotal question is not whether to employ a ViT or a CNN, but rather which design choices within vision transformers influence the accuracy-robustness trade-off when adversarial training is incorporated into a federated learning loop. This represents the central inquiry of the present study. We investigate the feasibility of FAT for vision transformers under both IID and Non-IID data distributions \cite{zhu@2021@noniid}, analyze convergence and drift during natural and adversarial training phases, and observe that convergence, accuracy, and robust accuracy are affected by both the model architecture and the aggregation methodology. To enhance robustness in heterogeneous environments, we extend the FedAvg \cite{mcmahan@2017@communication} aggregation algorithm to a weighted averaging scheme, in which weights are determined by the similarity between the global model's final layer and the final layers of client updates. We refer to this method as FedWAvg and discuss it in Sec.~\ref{sec:theory}. Additionally, we examine various vision transformer architectures to identify sources of robustness within the FAT context, including tokenization techniques such as Tokens-to-Token ViT (T2T-ViT) \cite{yuan@2021@tokens} and Transformer-in-Transformer (TNT) \cite{han@2021@transformer}, as well as classification-head strategies such as second-order cross-covariance pooling of visual tokens \cite{xie@2021@sovit}. The primary contributions of this research are as follows:
	\begin{itemize}[noitemsep,leftmargin=3mm,topsep=0pt,itemsep=0pt,partopsep=0pt,parsep=0pt]
		\item We demonstrate that three aggregation techniques designed to address client drift on Non-IID data, namely FedProx, FedGate, and SCAFFOLD, reduce the robust accuracy of vision transformers under FAT in comparison to the traditional FedAvg algorithm. Furthermore, as discussed in Sec.~\ref{sec:cost}, these methods achieve this but require up to twice the per-round communication and an additional client-side model state.
		
		\item We explore how different tokenization and classification-head techniques in vision transformers influence the overall accuracy of the model. We discover a trade-off between clean and robust accuracy that is influenced by tokenization, and we also highlight some potential issues with classification heads that rely solely on visual tokens.
		
		\item We introduce an aggregation methodology for the FAT process, designated as FedWAvg, aimed at enhancing the robustness and accuracy of the global model when dealing with Non-IID data distributions. FedWAvg employs the similarities between the final layer of the global model and the final layers of client updates to determine the aggregation weights.
		
		\item We analyse FedWAvg theoretically in Sec.~\ref{sec:theory}. Our proof demonstrates that it is precisely a Gibbs reweighting of clients based on the squared distance between their direction-normalized last layers. Consequently, it never increases the overall directional drift of the classification head compared to FedAvg, with the improvement expressed in closed form. We precisely establish bounds for the aggregation weights and their deviations from FedAvg, derive a conditional stationarity-transfer result, and develop a broad adversarially robust generalization bound for the specific weighting scheme..
	\end{itemize}
	
	\vspace{-5mm}
	\section{Related Works}
	\label{sec:related_work}
	%
	\vspace{-1mm}
	Our main focus is federated adversarial training; therefore, the related work on vision transformers is discussed in \cref{app:vits}.
	
	\vspace{-3mm}
	\subsection{Aggregation methods in \ac{fl}}
	\label{sec:aggmethods}
	\vspace{-1mm}
	McMahan \etal \cite{mcmahan@2017@communication} were the first to introduce the concept of federated averaging aggregation (FedAvg). It provides communication-efficient performance by allowing clients to perform multiple local steps before sending their updates. On the other hand, FedAvg’s performance on \ac{niid} is questionable because it causes clients to drift from one another, resulting in slow global model convergence.
	To address this issue, other algorithms were introduced \cite{li@2020@optimization}, \cite{haddadpour@2021@compression}, \cite{ezzeldin@2023@fairfed}, \cite{zhang@2024@fedsl}, \cite{karimireddy@2020@scaffold}, \cite{qi@2024@aggregation}.
	In \cite{li@2020@optimization}, FedProx aggregation brings an additional $L_2$-regularization term in the client loss function to maintain the difference between the client model and the global model. To control the regularization, a control parameter $\mu$ is set. The regularization term almost has no effect if $\mu$ is too small, moreover, if $\mu$ is large, the client update will be too limited to the previous round and will cause slow convergence. Hence, careful tuning is required for the regularization term.
	Normalised averaging methods address the case where clients perform different numbers of local steps, rescaling each update before aggregation; see \cite{qi@2024@aggregation} for a survey of model aggregation techniques in FL.
	\Ac{scaffold} \cite{karimireddy@2020@scaffold} algorithm uses variance reduction to correct for the client drift. In \ac{scaffold}, control terms are identified for the server and clients to estimate the update direction of the global model and each client’s update direction. The difference between these update directions is used to estimate client drift, which is then added to the client updates.
	FedGate \cite{haddadpour@2021@compression} adopts local gradient tracking, ensuring that each client uses an estimate of the global gradient direction to update its model. Compared to \ac{scaffold}, FedGate is much simpler, and no extra control parameters are required.
	In \cite{zhang@2024@fedsl}, the Federated Split Layers (FedSL) is introduced, which divides client models into groups in the models’ layers. Hence, each client is assigned to transfers partial parameters of those layers. The global model is aggregated in each group and the parameters are concatenated of all groups.
	FairFed \cite{ezzeldin@2023@fairfed} adopts a \acf{ffl} concept. It allows the clients to evaluate the fairness of the global model on their local datasets and then the server, for the aggregation, calculates the weights that are a function of the mismatch between the global fairness measurement and the local fairness measurement at each client.
	All the aforementioned \ac{fl} algorithms are server-based \ac{fl} paradigms that produce a global model. In the literature, another paradigm can be viewed as an intermediate one between server-based \ac{fl} and local model training. This paradigm is called personalized \ac{fl} that produces an additional personalized model, beside the global model, to balance between the local task-specific and the task-general models. In our investigation, we focus on the server-based paradigm and refer to \cite{ye@2023@heterogeneous}, \cite{tan@2022@personalized} for more information on personalized \ac{fl}.
	
	None of the previously mentioned server-based algorithms have been examined for vision transformer models in AT settings; the aggregation techniques that reweight client contributions instead of correcting local updates are discussed in Sec.~\ref{sec:aggrelated}

	\begin{table}[t]
		\vspace{-3mm}
		\centering
		\caption{Coverage of closely related literature across dimensions relevant to this paper. FL: federated setting; AT: adversarial training within the training loop; ViT: vision transformer backbone; Non-IID: heterogeneous client partitions; Agg.: two or more aggregation methods compared; Tok./head: tokenization or classification-head variants ablated; Theory: convergence, generalization, or robustness guarantees provided.}
		\label{tab:coverage}
		\resizebox{\linewidth}{!}{%
			\begin{tabular}{llccccccc}
				\hline
				Work & Backbone & FL & AT & ViT & Non-IID & Agg. & Tok./head & Theory \\
				\hline
				Shah \emph{et al.} \cite{shah@2021@adversarial}       & CNN & \checkmark & \checkmark & --         & \checkmark & --         & -- & -- \\
				Lee \cite{lee@2024@adaptive}                          & CNN & \checkmark & \checkmark & --         & \checkmark & \checkmark & -- & -- \\
				Chen \emph{et al.} \cite{chen@2021@certifiably}       & CNN & \checkmark & \checkmark & --         & --         & --         & -- & \checkmark \\
				Hong \emph{et al.} \cite{hong@2023@propagation}         & CNN & \checkmark & \checkmark & --         & \checkmark & --         & -- & -- \\
				Chen \emph{et al.} \cite{chen@2022@calfat}            & CNN & \checkmark & \checkmark & --         & \checkmark & --         & -- & \checkmark \\
				Zhang \emph{et al.} \cite{zhang@2023@delving}           & CNN & \checkmark & \checkmark & --         & \checkmark & \checkmark & -- & -- \\
				Zhu \emph{et al.} \cite{zhu@2023@combating}                & CNN & \checkmark & \checkmark & --         & \checkmark & \checkmark & -- & \checkmark \\
				Li \emph{et al.} \cite{li@2023@federated}                   & CNN & \checkmark & \checkmark & --         & \checkmark & --         & -- & \checkmark \\
				Cao \emph{et al.} \cite{cao@2021@fltrust}             & CNN & \checkmark & --         & --         & \checkmark & \checkmark & -- & \checkmark \\
				Ye \emph{et al.} \cite{ye@2023@feddisco}              & CNN & \checkmark & --         & --         & \checkmark & \checkmark & -- & \checkmark \\
				Li \emph{et al.} \cite{li@2023@fedlaw}                & CNN & \checkmark & --         & --         & \checkmark & \checkmark & -- & \checkmark \\
				Rehman \emph{et al.} \cite{rehman@2023@ldawa}         & CNN & \checkmark & --         & --         & \checkmark & \checkmark & -- & -- \\
				Qu \emph{et al.} \cite{qu@2022@rethinking}            & CNN / ViT & \checkmark & --   & \checkmark & \checkmark & \checkmark & -- & -- \\
				Bai \emph{et al.} \cite{bai@2021@are}        & CNN / ViT & --         & \checkmark & \checkmark & --  & --         & -- & -- \\
				Aldahdooh \emph{et al.} \cite{aldahdooh@2025@beyond} & ViT & -- & -- & \checkmark & -- & --      & \checkmark & -- \\
				Mao \emph{et al.} \cite{mao@2021@reversible}$^{\dagger}$   & ViT & --        & --         & \checkmark & --         & --         & \checkmark & -- \\
				\textbf{This work}                                  & ViT & \checkmark & \checkmark & \checkmark & \checkmark & \checkmark & \checkmark & \checkmark \\
				\hline
			\end{tabular}
		}%
		\vspace{-7.5mm}
	\end{table}
	
	\vspace{-2mm}
	\subsection{\Acf{fat}}
	\vspace{-1mm}
	\Ac{at} is a way to reduce the threat of evasion attacks, i.e., attacks during the inference time. \ac{at} retrains the model by including the \acp{ae} in the training process and aims at solving the min-max optimization problem \cref{eq:min_max_at} \cite{madry@2018@towards}.
	\begin{equation}
		\vspace{-5mm}
		\underset{\theta}{\text{min }} \rho(\theta), \text{ } \rho(\theta) = \mathbf{E}_{(x,y)\sim \mathcal{D}}[\underset{\epsilon}{\text{max }} \mathcal{L}(\theta, x+\epsilon, y)]
		\label{eq:min_max_at}
		\vspace{1mm}   
	\end{equation}
	\textit{where, \\}
	\tab \begin{tabular}{p{0.86\linewidth}}
		- $\mathcal{D}$ and ($x,y$): data distribution $\mathcal{D}$ over pairs of examples $x \in \mathbf{R}^{d}$ and its labels $y \in \{classes\}$.  \\
		- $\theta \in \mathcal{R}^{p}$: is the set of model parameters of the neural network of the standard classification task. \\
		- $\mathcal{L}(\theta, x, y)$: is the loss function for the neural network.\\
		- $\epsilon$: is the allowed $l_p$-ball perturbation around $x$.\\
		- $\mathbf{E}_{(x,y)\sim \mathcal{D}} [\mathcal{L}(\theta, x, y)]$: is the risk of the neural network model, i.e. the risk of the prediction function.\\\\
	\end{tabular}
	
	The inner maximization problem identifies the worst-case \acp{ae} for the neural network model, while the outer minimization problem identifies the model parameters that minimize the adversarial loss induced by the inner maximization problem. As a result of solving the min-max optimization, a robust classifier is trained.
	
	\Ac{at} has been well investigated for centralized training and CNNs
	\cite{bai@2021@recent}. Under \ac{fl} settings the literature is considerably thinner,
	and it is uniformly convolutional. Tab.~\ref{tab:coverage} summarizes the studies discussed below according to the dimensions relevant to this paper.
	
	Shah \emph{et al.} \cite{shah@2021@adversarial} analyze the overhead caused by adversarial training (AT) in federated learning (FL) and propose an adaptive rule for setting the number of local epochs $E$ to minimize model drift. However, this rule is not applicable when only a single local epoch can be performed, which is recommended for highly heterogeneous partitions \cite{qu@2022@rethinking} and is the setting used in this study. Lee \cite{lee@2024@adaptive} investigates Byzantine-resilient defenses in the federated adversarial training (FAT) setting and finds that they substantially negatively affect adversarial robustness. Chen \emph{et al.} \cite{chen@2021@certifiably} combine randomized smoothing with FAT to attain certifiable robustness. Hong \emph{et al.} \cite{hong@2023@propagation} transfer robustness from high-resource clients capable of AT to low-resource clients unable to perform AT by sharing batch-normalization statistics; however, since the server requires access to clients' parameters, this method compromises secure aggregation.

	Three recent studies highlight data heterogeneity as a key challenge and are closely related to this paper. Chen \emph{et al.} \cite{chen@2022@calfat} find that skewed label distributions cause clients to fit non-identical class probabilities, leading to diverse local models. They see this as the main reason for instability during FAT training and the drop in natural accuracy, which they address by calibrating local logits; their experiments involve convolutional backbones with up to 100 clients. Zhu \emph{et al.} \cite{zhu@2023@combating} look at the opposite side: adversarial training’s maximization step actually worsens existing client heterogeneity. They link this to the reduction in robustness observed when adding adversarial training to federated learning. Their aggregation method, like ours, assigns different weights to clients, but they base these weights on local adversarial loss instead of the angular drift of the classification head. Zhang \emph{et al.} \cite{zhang@2023@delving} take a different route by combining local training sample re-weighting using a decision boundary with a global regularization to boost both clean and robust accuracy. Additionally, Li \emph{et al.} \cite{li@2023@federated} offer a convergence analysis of federated adversarial learning involving multiple local steps and non-IID clients, which we discuss in Sec.~\ref{sec:theory}.
	
	The following two points are articulated to define the scope of this paper. First, all prior research employs a convolutional backbone; no earlier investigation has assessed a vision transformer within the FAT setting, nor have they examined tokenization or the design of the classification head as strategies to enhance robustness when adversarial training occurs in the presence of label skew. Second, the observation that the classifier layer constitutes the source of heterogeneity induced by label skew \cite{chen@2022@calfat} is precisely the insight that informs the aggregation rule proposed in Sec.~\ref{sec:wavg} and analyzed in Sec.~\ref{sec:theory}. It is important to note that the CNN-based FAT results published in  \cite{shah@2021@adversarial},  	\cite{chen@2022@calfat}, \cite{zhu@2023@combating} and \cite{zhang@2023@delving} were obtained using different backbones, data partitions, and attack budgets than those used in this study. Therefore, they should primarily serve as reference points rather than a basis for a controlled comparison. A matched ResNet-50 FAT baseline is stated as a topic for future investigation in  Sec.~\ref{sec:scope}.
	
	\vspace{-2mm}
	\subsection{Robust and heterogeneity-aware aggregation}
	\label{sec:aggrelated}
	\vspace{-1mm}
	The methods described in Sec.~\ref{sec:aggmethods} include various approaches. Some alter the local objective, like FedProx, while others, such as SCAFFOLD or FedGate, change the direction of local updates. Additionally, some methods modify the aggregation weights, and FedWAvg is an example of this kind.
	
	Cao \emph{et al.} \cite{cao@2021@fltrust} propose a trust score for each client, utilizing the ReLU-clip cosine similarity between the client's update and a server update derived from a small, clean root dataset. They carefully normalize the magnitudes of updates and employ a weighted average for aggregation. Furthermore, they establish a bound on the deviation of the final global model from the optimal model in the absence of attacks. Rehman \emph{et al.} \cite{rehman@2023@ldawa} investigate the angular divergence between each client's model and the global model as a layer-wise weighting factor, emphasizing that this approach aids in reducing client bias similarly to momentum. Ye \emph{et al.} \cite{ye@2023@feddisco} formulate aggregation weights based on the discrepancy between local and global category distributions, demonstrating that these weights enable a tighter bound on optimization error compared to weights solely based on dataset size. Li \emph{et al.} \cite{li@2023@fedlaw} revisit the concept of weighted aggregation, highlighting that weights proportional to local dataset size are not invariably optimal. They propose that client coherence should inform weight assignment and advocate for learning these weights rather than deriving them analytically. Within the adversarial setting, Zhu \emph{et al.}
	\cite{zhu@2023@combating} adaptively reweight clients during aggregation based on their local adversarial loss.
	
	FedWAvg closely resembles \cite{cao@2021@fltrust} and \cite{rehman@2023@ldawa} because its weights are based on angles; it also relates to \cite{ye@2023@feddisco} and \cite{li@2023@fedlaw} because they identify a local–global discrepancy; and it is similar to \cite{zhu@2023@combating} in being tailored for adversarial situations. However, it differs from each of these four methods in four key ways. First, unlike \cite{cao@2021@fltrust}, it does not need server-side data. Second, it assesses discrepancy only on the classification head, the layer shown to be affected by label skew under adversarial training \cite{chen@2022@calfat}, rather than layer-by-layer or across the full parameter vector. Third, its weights can be calculated in closed form from data the server already has, unlike \cite{li@2023@fedlaw}, which requires additional optimization. Fourth, it employs a softmax scaled version instead of direct normalization, ensuring a valid convex combination even with negative similarities, avoiding the clipping used in \cite{cao@2021@fltrust} that could exclude benign but divergent clients. As a result, it has a single parameter that can be shown to interpolate between FedAvg and greedy selection (Corollary~\ref{cor:fedavg}).
	
	We emphasize that FedWAvg is a heterogeneity-aware aggregation rule and not a
	Byzantine-robust one. According to \cite{lee@2024@adaptive} in the FAT setting, Byzantine defenses replace the average with a more robust estimator, like a coordinate-wise median or a trimmed mean. Since, as shown in Lemma~\ref{lem:bounded}, FedWAvg never assigns a client less than a weight of $e^{-2q}/K$, and it doesn't offer protection against a highly adversarial participant.
	
	Inner-product and cosine similarity are common in various contexts within adversarial machine learning, but their use here is distinct. For example, 	\cite{wan@2021@robust} uses dot products between client updates to generate attention-based weights for defending against model poisoning; \cite{mao@2021@reversible} employs a contrastive loss 	\cite{chen@2020@simple} to reverse adversarial attacks by finding perturbations that fix the adversarial example during inference; and \cite{dong@2021@competition} describes an attack leveraging cosine similarity to pick the initial perturbed points of an adversarial example. The capacity of inner-product similarity measures is discussed in 	\cite{okuno@2018@probabilistic}. Among these, only \cite{wan@2021@robust} focuses on aggregation weighting, specifically targeting malicious clients rather than just heterogeneous ones. FedWAvg differs because the server computes the similarity solely on the classification head, applies a scaled softmax to produce a convex combination, and uses this to mitigate benign client drift within a min-max framework.

	\vspace{-2mm}
	\section{Methodology}
	\vspace{-1mm}
	\subsection{Weighted Averaging Aggregation for Adversarial Training}
	\label{sec:fedwavg}
	\subsubsection{Preliminaries}
	\label{sec:preliminaries}
	\vspace{-1mm}
	Let $D = \{(x, y)\}$ represent the global dataset of $n$ samples and assume there are $K$ clients indexed by $k$, where $k=\{1,2,\dots,K\}$. The local dataset of client $k$ is denoted as $D_k = \{(x_k,y_k)\}$ with $n_k$ samples. We use $\theta_t$ and $\theta_t^k$ to denote the global model and the local model parameters of client $k$ in communication round $t$, respectively, where $t={1,2,\dots,T}$ and $T$ is the total number of communication rounds. Therefore, $\theta_t$ is the output of the \ac{fl} process at communication round $t$. In FedAvg \cite{mcmahan@2017@communication}, the output of the \ac{fl} process is computed as follows:
	\vspace{-2mm}
	\begin{equation}
		\label{eq:fed_avg}
		\theta_{t} = \sum_{k=1}^{K}{p_k\theta_{t}^{k}}, \text{where, } p_k=\frac{n_k}{n}
	\end{equation}
	
	If all clients have the same number of samples, as is the case for all three partitions used in this paper, the averaging weights reduce to $p_k = 1/K$, regardless of whether the data are independent and identically distributed (IID) or Non-IID.

	\ac{fl} and \ac{at} strategies result in the following challenges. \textbf{1) Accuracy drop:} In natural \ac{fl} process, it was shown in \cite{shah@2021@adversarial}, \cite{qu@2022@rethinking} that the model accuracy slightly decreases with \ac{iid} and remarkably decreases with \ac{niid}. \Ac{fat} increases the model’s adversarial accuracy at the price of model accuracy. \textbf{2) Overfitting:} It was shown that AT suffers from robust overfitting, whereby adversarially trained models continue to fit the training AEs while their robustness on unseen AEs degrades \cite{yu@2022@overfitting}. \textbf{3)} We hypothesize that when vision transformer models are used in natural and adversarial \ac{fl} settings, \stateart aggregation algorithms designed to reduce client drift on \ac{niid} data will not outperform FedAvg. It was found in \cite{qu@2022@rethinking} that \acp{vit} have significantly better convergence than \acp{cnn} with the \ac{niid}, which significantly reduces the model drift problem that \stateart aggregation methods try to solve. This hypothesis is confirmed empirically in Sec.~\ref{sec:results} and explained theoretically in Sec.~\ref{sec:theory}.
	

	\vspace{-5mm}
	\subsubsection{The Weighted Averaging Aggregation}
	\label{sec:wavg} 
	\vspace{-1mm}
	In order to enhance the robust accuracy of vision transformer models under \ac{fat} settings, particularly with \ac{niid} data distributions, we leverage the similarities between the global model’s last layer and the client models’ last layers to generate aggregation weights. Denote $g$ and $l_k$ as the last layer parameters of the global server and of client $k$, respectively. First, in each communication round $t$, we calculate the \textit{cosine} similarity $c_k$ between $g$ and $l_k$. Then, we calculate the weights according to the similarities using the \textit{softmax} function with a scale factor $q$:
	\vspace{-2mm}
	\begin{equation}
		\begin{gathered}
			\label{eq:weights_cos}
			c_k = \frac{g \cdot l_k}{||g||_2||l_k||_2}, \text{  }
			w_k = \frac{exp(q \, c_k)}{\sum_{j=1}^{K} exp(q \, c_j)}
		\end{gathered}
		\vspace{-1mm}
	\end{equation}
	
	We calculate the similarities and the weights according to \cref{eq:weights_cos} in natural and adversarial \ac{fl} using FedAvg and FedProx as illustrated in \cref{fig:wieghts_tnts_cls}. We have noted the following observations:
	
	\textbf{With \ac{iid} data partitioning.} The weights of the natural and adversarial \ac{fl} are close to the weights that are used in FedAvg and FedProx aggregation methods, which is $\frac{1}{K}=0.2$. 
	
	\textbf{With \ac{niid} data partitioning.} In the natural training, depending on the model architecture, the weights can be close to $\frac{1}{K}$ as with \ac{t2t} model as shown in \cref{fig:wieghts_t2t_vis_niid4} in \cref{app:more_cosine_sim_w}, while the weights are not close to $\frac{1}{K}$ in the first $t$ rounds with the \ac{tnt} model. On the other hand, in the \ac{fat} settings, the weights are far from $\frac{1}{K}$ in the first $t$ rounds. The difference in the weights are higher in the highly heterogeneous data partitions.
	
	By using the weights in \cref{eq:weights_cos} to aggregate the global server model, we investigated the impact of this step in enhancing the robust accuracy of the \ac{fat} process. Applying \cref{eq:weights_cos} in the \ac{fat} setting will push the global model in the direction of a model that has high similarity to the global model. \cref{alg:algo_fedwavg} shows the whole process of the \ac{fat} with FedWAvg aggregation method. 
	
	First, the server initializes the global model parameters $\theta_0$ and sends them to $m$ clients randomly selected from the $K$ clients. Then, each client, in parallel, perform \ac{at} process in which $n_{adv}$ samples from batch $b$ are transformed into their adversarial version using the \ac{pgd} attack, and then each client updates its model parameters according to the gradient of the computed loss and sends them back to the server. Finally, the server aggregates the weights generated by the proposed FedWAvg procedure, as shown in \cref{alg:algo_fedwavg}. The server repeats the aforementioned steps for $T$ communication rounds. 
	
	\setlength{\textfloatsep}{5pt}
	\begin{algorithm}[!t]
		\small
		\caption{\textbf{FedWAvg Algorithm}. Weighted Averaging Aggregation for Adversarial Training
			\label{alg:algo_fedwavg}}
		\begin{algorithmic}[1]
			\Require{The $K$ clients are indexed by $k$, $C$ is the client fraction, the $T$ communication rounds are indexed by $t$, $B$ is the local minibatch size, $E$ is the number of local epochs, and $\eta$ is the learning rate, \acs{pgd} Attack $A_{s,\epsilon,\alpha}$: where $s,\epsilon,\alpha$ are number of \acs{pgd} steps, perturbation ball size, step size, $r$ is the adversarial ratio, $q$ is the scale factor.}
			\Ensure{The global model $\theta$.}
			\Statex \textbf{On Server}:
			\Indent
			\State Initialize $\theta_0$
			\For{ each round $t=1,2,\dots,T$}
			\Let{$m$}{max(1, $CK$)}
			\Let{$S_t$}{(random set of m clients)}
			\For{for each client $k \in S_t$ \textbf{in parallel}}
			\Let{$\theta_{t+1}^{k}$}{ClientUpdate($k$, $\theta_t$)}
			\EndFor
			\Let{$\theta_{t+1}$}{FedWAvg($\theta_t, \{\theta_{t+1}^{k}\}_{k\in S_t}$)}
			\EndFor
			\State \Return{$\theta_{t+1}$}
			\EndIndent
			\Statex
			\Statex \textbf{ClientUpdate($k$, $\theta$)}: \Comment{Run on client $k$}
			\Indent
			\Let{$\mathcal{B}$}{split the training data into batches of size $B$}
			\For{each local epoch $i$ from 1 to $E$}
			\For{batch $b \in \mathcal{B}$}
			\Let{$n_{adv}$}{$r \cdot B$} \Comment{$0 \leq r \leq 1$}
			\Let{$b_{adv}$}{(random set $\in b$ of $n_{adv}$ samples)}
			\Let{$b_{nat}$}{(set of $B-n_{adv}$ samples)}
			\Let{$b_{adv}$}{$A_{s,\epsilon,\alpha}(b_{adv})$}
			\Let{$b$}{$b_{nat} \cup b_{adv}$}
			\Let{$\theta$}{$\theta-\eta \nabla \mathcal{L}(\theta;b) $}
			\EndFor
			\EndFor
			\State \Return{$\theta$}
			\EndIndent
			\Statex
			\Statex \textbf{FedWAvg($G,\{L_i\}_{i=1}^{n}$)}: \Comment{Run on Server}
			\Indent
			\Let{$g$}{parameters of last layer of $G$}
			\For{$i=1$ to $n$ \textbf{in parallel}}
			\Let{$l_i$}{parameters of last layer of $L_i$}
			\Let{$c_i$}{$\frac{g \cdot l_i}{||g||_2||l_i||_2}$} \Comment{cosine Similarity}
			\Let{$w_i$}{$\frac{exp(qc_i)}{\sum_{j=1}^{n} exp(qc_j)}$} \Comment{softmax}
			\EndFor
			\Let{$G$}{$\sum_{i=1}^{n} w_i L_i $}
			\State \Return{$G$}
			\EndIndent
		\end{algorithmic}
		
	\end{algorithm}
	
	\vspace{-2mm}
	\subsection{Theoretical Analysis of FedWAvg}
	\label{sec:theory}
	\vspace{-1mm}
	Sec.~\ref{sec:wavg} introduced the empirical basis for Eq.~\eqref{eq:weights_cos}. Here, we show that this rule is more than a similarity heuristic: it is exactly a Gibbs reweighting of clients by the squared distance between their direction-normalized last layers. This yields a drift-contraction guarantee relative to FedAvg, bounded aggregation weights, and two exact deviation bounds, from which a robust generalization result follows as a reduction. We state the scope precisely, because the results differ in status. Proposition~\ref{prop:gibbs}, Theorem~\ref{thm:contract} and Lemmas~\ref{lem:bounded}-\ref{lem:graddev} are self-contained and exact. Proposition~\ref{prop:transfer} is \emph{conditional}: it transfers a stationarity guarantee for the weighted objective to the uniform objective, but does not establish that guarantee. Theorem~\ref{thm:gen} is a reduction that prices the weighting against any admissible uniform-convergence bound. We do not claim a convergence theorem for FedWAvg under the min-max objective, and Sec.~\ref{sec:openproblems} states explicitly what such a theorem would additionally require.
	
	\vspace{-3mm}
	\subsubsection{Setting and assumptions}
	\vspace{-1mm}
	The FAT objective is $\min_{\theta} F(\theta)$ for
	\begin{align}
		F(\theta) &:= \sum_{k=1}^{K} p_k F_k(\theta), \label{eq:fatobj} \\
		F_k(\theta) &:= \mathbb{E}_{(x,y)\sim\mathcal{D}_k}
		\Big[\max_{\|\varepsilon\|_p \le \varepsilon_0}
		\mathcal{L}(\theta, x+\varepsilon, y)\Big], \label{eq:localrisk}
	\end{align}
	where $\mathcal{D}_k$ is the local data distribution of client $k$, $\varepsilon_0$ the perturbation budget, and $p_k = n_k/n$ the mixture weights that define the target objective.
	
	The three weightings listed below must be kept distinct. The vector $p=(p_1,\dots,p_K)$, determined by data partition, indicates  \emph{what} Eq.~\eqref{eq:fatobj} asks us to minimize. The aggregation weights in Eq.~\eqref{eq:fed_avg} and Eq.~\eqref{eq:weights_cos} show  \emph{how} the server combines models from clients and are chosen by the algorithm: FedAvg sets them equal to $p$, while FedWAvg sets them equal to $w$ as defined in Eq.~\eqref{eq:weights_cos}, typically resulting in $w \neq p$. The difference between $w$ and $p$ reflects the method used, with Lemmas~\ref{lem:dev} and~\ref{lem:graddev} providing bounds on this discrepancy. As mentioned in Sec.~\ref{sec:preliminaries}, all three partitions in this paper assign the same number of samples to each client, so in Sec.~\ref{sec:theory} we assume,
	\begin{equation}
		n_1 = \cdots = n_K, \qquad p_k = 1/K, \qquad p = u := (1/K)\mathbf{1},
		\label{eq:equalsize}
	\end{equation}
	This is why the uniform weighting $u$ can serve as both the FedAvg aggregation rule and the target mixture. Figs.~\ref{fig:wieghts_tnts_cls} and~\ref{fig:wieghts_vits_cls_iid} illustrate this, showing FedAvg weights equal to $1/K = 0.2$. The following assumptions are made.
	
	\begin{assumption}[Smoothness]\label{as:smooth}
		Each local robust risk $F_k$ is $L$-smooth. By Danskin's theorem,
		$\nabla_\theta F_k(\theta) =
		\mathbb{E}[\nabla_\theta \mathcal{L}(\theta, x+\varepsilon^\star(\theta), y)]$
		whenever the inner maximizer $\varepsilon^\star(\theta)$ is unique, so the
		robust surrogate is differentiable and the gradient is that of the loss
		evaluated at the worst-case perturbation. Smoothness of the resulting robust
		objective is the standard assumption in the convergence analysis of federated
		adversarial learning \cite{li@2023@federated}, \cite{zhu@2023@combating}.
	\end{assumption}
	
	\begin{assumption}[Bounded variance]\label{as:var}
		$\mathbb{E}\|\nabla \mathcal{L}(\theta;b) - \nabla F_k(\theta)\|^2 \le \sigma^2$
		for every client $k$ and minibatch $b$.
	\end{assumption}
	
	\begin{assumption}[Bounded dissimilarity]\label{as:diss}
		$\|\nabla F_k(\theta) - \nabla F(\theta)\| \le \kappa$ for all $k,\theta$.
		The constant $\kappa$ quantifies the label skew.
	\end{assumption}
	
	\begin{assumption}[Inexact inner maximization]\label{as:inner}
		The $s$-step PGD attack $\mathcal{A}_{s,\varepsilon_0,\alpha}$ of Algorithm~\ref{alg:algo_fedwavg} returns a perturbation whose induced gradient error is bounded: the local update direction $\widehat\nabla F_k$ computed from the attacked batch satisfies $\|\widehat\nabla F_k(\theta) - \nabla F_k(\theta)\| \le \zeta$ for all $k,\theta$. We assume the gradient level rather than the loss level because the analysis focuses on the gradient. A bound on the inner-maximization \emph{value} gap alone does not guarantee a bound on the gradient error unless the inner problem has extra properties, such as strong concavity in $\varepsilon$.
	\end{assumption}
	
	\begin{assumption}[Non-degenerate head]\label{as:nondeg}
		$\|g\|_2 > 0$ and $\langle l_k, g\rangle > 0$ for all $k$, where $g$ and $l_k$
		are as in Eq.~\eqref{eq:weights_cos}. This holds throughout training when clients are initialized
		from the current global model and perform a bounded number of local steps.
	\end{assumption}
	
	\vspace{-5mm}
	\subsubsection{Why the last layer, and why adversarial training}
	\label{sec:whylast}
	\vspace{-1mm}
	Two observations localize client drift within the FAT setting to the classification head, which is the layer inspected by Eq.~\eqref{eq:weights_cos}.
	
	Firstly, within the context of label skew, the head is demonstrably identified as the locus of heterogeneity. \cite{chen@2022@calfat} demonstrates that skewed label distributions induce clients to fit \emph{non-identical class probabilities} and recognizes this as a fundamental cause of FAT instability and deterioration in natural accuracy. Their proposed solution involves calibrating the logits, specifically adjusting the parameters $g$ and $l_k$ of Eq.~\eqref{eq:weights_cos}, from the client side. FedWAvg operates on the same layer from the server side and consequently requires no modifications to local training procedures. Fig.~\ref{fig:drift_tnts_cls_iid4_at} shows that SV-CCA measurements substantiate this localization directly: at the first layer, the client–server similarity converges under all aggregation methods; however, at the ninth layer, the methods diverge, and the drift remains evident.
	
	Secondly, adversarial training amplifies head drift through a feedback mechanism that is absent in conventional training. According to Assumption~\ref{as:smooth}, the local robust gradient is represented by $\nabla_\theta \mathcal{L}(\theta, x +
	\varepsilon^\star(\theta), y)$, wherein the worst-case perturbation $\varepsilon^\star(\theta)$ itself depends on the \emph{local} parameters. Consequently, each client's PGD attack is performed against its own locally biased decision boundary; the resulting adversarial examples (AEs) further reinforce this bias, causing the head to diverge even more in subsequent rounds. This phenomenon is not merely hypothetical: \cite{zhu@2023@combating} independently identified the same effect, demonstrating that the inner maximization in adversarial training exacerbates data heterogeneity among clients, and attributing the decline in robustness observed in naive FAT to this issue. They also respond by adjusting client weights during aggregation. Our Eq.~\eqref{eq:weights_cos} can be interpreted as an alternative implementation of this principle, wherein the weighting signal is the angular deviation of the classification head rather than the local adversarial loss. This mechanism predicts the behaviors documented in Sec.~\ref{sec:wavg}, specifically, that the weights in Eq.~\eqref{eq:weights_cos} significantly deviate from $1/K$ under adversarial training compared to non-adversarial training, especially in highly skewed partitions, rather than merely observing such deviations.
	
	\vspace{-4mm}
	\subsubsection{FedWAvg is a Gibbs reweighting of directional drift}
	\vspace{-1mm}
	Consider $\hat g = g/\|g\|_2$ and $\hat l_k = l_k/\|l_k\|_2$, and define the
	\emph{directional drift} of client $k$ as
	\begin{equation}
		\vspace{-1mm}
		d_k := \|\hat g - \hat l_k\|_2^2 = 2\,(1 - c_k) \in [0,4].
		\label{eq:drift}
	\end{equation}
	
	\begin{proposition}[Exact Gibbs form]\label{prop:gibbs}
		For every $q \ge 0$ the weights of Eq.~\eqref{eq:weights_cos} satisfy the identity
		\begin{equation}
			\vspace{-2mm}
			w_k \;=\; \frac{\exp(q\,c_k)}{\sum_{j=1}^{K}\exp(q\,c_j)}
			\;=\; \frac{\exp\!\big(-\tfrac{q}{2}\,d_k\big)}
			{\sum_{j=1}^{K}\exp\!\big(-\tfrac{q}{2}\,d_j\big)} .
			\label{eq:gibbs}
		\end{equation}
		That is, FedWAvg is the Boltzmann distribution over clients at inverse
		temperature $q/2$ with energy $d_k$.
	\end{proposition}
	
	\begin{proof}
		By Eq.~\eqref{eq:drift}, $c_k = 1 - d_k/2$, so
		$\exp(q c_k) = e^{q}\exp(-\tfrac{q}{2}d_k)$. The factor $e^q$ is common to
		numerator and denominator and cancels.
	\end{proof}
	
	Proposition~\ref{prop:gibbs} shows that the cosine similarity in Eq.~\eqref{eq:weights_cos} is a
	\emph{scale-invariant} drift measure. Decomposing the client head as
	$l_k = g + \Delta_k$ with $\Delta_k = a_k \hat g + \Delta_k^{\perp}$ and
	$\Delta_k^{\perp} \perp g$, Assumption~\ref{as:nondeg} gives the exact identity
	\begin{equation}
		\vspace{-2mm}
		c_k = \frac{1 + a_k/\|g\|}
		{\sqrt{(1 + a_k/\|g\|)^2 + \|\Delta_k^{\perp}\|^2/\|g\|^2}} ,
		\label{eq:angular}
	\end{equation}
	and hence, in the small-update regime $\|\Delta_k\| \ll \|g\|$,
	\begin{equation}
		\vspace{-2mm}
		d_k = \frac{\|\Delta_k^{\perp}\|^2}{\|g\|^2}
		+ O\!\big(\|\Delta_k\|^3/\|g\|^3\big).
		\label{eq:angular2}
	\end{equation}
	Eq.~\eqref{eq:angular2} provides an expansion valid when $\|\Delta_k\| \ll \|g\|$, but it is not a universal bound since Eq.~\eqref{eq:angular} is exact. It states that the component of the local update parallel to $g$ does not influence the second-order directional drift, while the orthogonal component is dominant. This behavior is ideal in cases of label skew, where clients have different update magnitudes due to class count or gradient norms but should still agree on the decision boundary's orientation. Additionally, Eq.~\eqref{eq:angular2} clarifies the usefulness of applying a scale factor $q>1$: starting from a pretrained initialization with $\|\Delta_k\| \ll \|g\|$, the $c_k$ stay near 1 after one local epoch, with significant divergence from $1/K$ only when $q$ is sufficiently large.
	
	\vspace{-4mm}
	\subsubsection{Drift contraction}
	\vspace{-1mm}
	\begin{theorem}[Drift contraction]\label{thm:contract}
		Let $\bar d(q) := \sum_{k=1}^{K} w_k(q)\, d_k$ denote the FedWAvg-weighted mean
		directional drift, with $w(q)$ as in Eq.~\eqref{eq:gibbs}. Then
		\begin{equation}
			\vspace{-1.5mm}
			\frac{d\,\bar d(q)}{dq} = -\tfrac12 \operatorname{Var}_{w(q)}(d) \le 0 ,
			\label{eq:derivative}
		\end{equation}
		and consequently, for every $q \ge 0$, every $K$, and every drift
		configuration,
		\begin{equation}
			\vspace{-1.5mm}
			\bar d(q) \;\le\; \bar d(0) \;=\; \frac{1}{K}\sum_{k=1}^{K} d_k
			\;=\; \bar d_{\mathrm{FedAvg}} ,
			\label{eq:contract}
		\end{equation}
		with the improvement given in closed form by
		\begin{equation}
			\vspace{-1.5mm}
			\bar d_{\mathrm{FedAvg}} - \bar d(q)
			= \tfrac12\!\int_{0}^{q}\!\operatorname{Var}_{w(b)}(d)\,db ,
			\label{eq:gap}
		\end{equation}
		and, for $q>0$, with equality in Eq.~\eqref{eq:contract} if and only if $d_1 = \dots = d_K$.
	\end{theorem}
	
	\begin{proof}
		Put $\beta = q/2$, so that $w_k(\beta) \propto e^{-\beta d_k}$ by
		Proposition~\ref{prop:gibbs}. Differentiating the normalized weights gives
		$\partial w_k/\partial\beta = w_k(\bar d - d_k)$, hence
		$\partial \bar d/\partial\beta = \sum_k w_k d_k(\bar d - d_k)
		= \bar d^{\,2} - \mathbb{E}_w[d^2] = -\operatorname{Var}_w(d) \le 0$.
		The chain rule with $d\beta/dq = 1/2$ yields Eq.~\eqref{eq:derivative};
		integrating from $0$ to $q$ yields Eq.~\eqref{eq:gap}, from which
		Eq.~\eqref{eq:contract} follows because the integrand is non-negative;
		$w(0)$ is the uniform weighting. For $q>0$, equality holds iff
		$\operatorname{Var}_{w(b)}(d)=0$ for all $b\in[0,q]$, that is, iff all $d_k$ coincide.
	\end{proof}
	
	Theorem~\ref{thm:contract} is exact and holds at every communication round: the
	aggregation rule of Eq.~\eqref{eq:weights_cos} can never increase the aggregate directional drift of the head relative to FedAvg, and the improvement equals the accumulated variance of the client drifts. What is not claimed here is emphasized; it only limits the directional drift of the classification head. It does not assert that robust accuracy improves, nor do the results in this section support that conclusion. The link between reducing head drift and improving robust accuracy is based on empirical observations in this study and remains an open theoretical question (Sec.~\ref{sec:openproblems}).

	\begin{corollary}[FedAvg is the $q\to 0$ limit]\label{cor:fedavg}
		$w(0) = (1/K)\mathbf{1}$, so FedWAvg strictly generalizes FedAvg, and $q$
		interpolates between FedAvg $(q\to0)$ and greedy selection of the most aligned
		client $(q\to\infty)$.
	\end{corollary}
	
	Once the drift configuration $\{d_k\}$ has been provided by the experiments, there are three further consequences. We refer to them as empirical corroboration rather than as corollaries since Theorem~\ref{thm:contract} treats the $d_k$ as given and does not itself explain how a partitioning scheme or a training regime influences them.
	
	\begin{remark}[Empirical corroboration]\label{rem:empirical}
		Under IID partitioning the measured $d_k$ are nearly equal, so
		$\operatorname{Var}_{w}(d)\approx 0$ and FedWAvg $\approx$ FedAvg; this matches Fig.~\ref{fig:wieghts_vits_cls_iid} in which the weights stay within $10^{-3}$ of $1/K = 0.2$, and the near-identical IID rows of Tab.~\ref{tab:iid}. In the cases of Non-IID($4$) and Non-IID($2$) the measured dispersion is large and the contraction indicated in Eq.~\eqref{eq:gap} is therefore considerable, which is consistent with Figs.~\ref{fig:wieghts_tnts_cls}(b), \ref{fig:wieghts_vits_cls_niid4}(b) and \ref{fig:wieghts_vits_cls_niid2}(b). Moreover, the feedback mechanism described in Sec.~\ref{sec:whylast}, and independently confirmed by \cite{zhu@2023@combating}, causes us to expect $\operatorname{Var}_w(d)$ to be inflated during adversarial training, and the deviation of the weights from $1/K$ is in fact greater under AT than under NT when the level of skew is the same, as stated in Sec.~\ref{sec:wavg}. None of these three observations is derived from Theorem~\ref{thm:contract}; rather, each provides the input upon which Theorem~\ref{thm:contract} then operates.
	\end{remark}
	
	\vspace{-5mm}
	\subsubsection{Bounded weights and deviation from FedAvg}
	\vspace{-1mm}
	\begin{lemma}[Bounded weights and effective client count]\label{lem:bounded}
		For every $k$ and every $q \ge 0$,
		\vspace{-3mm}
		\begin{equation}
			\vspace{-1.5mm}
			\frac{e^{-2q}}{K} \;\le\; w_k \;\le\; \frac{e^{2q}}{K} ,
			\label{eq:bounded}
		\end{equation}
		and hence $\|w\|_2^2 \le e^{4q}/K$, so that the effective client count
		satisfies $K_{\mathrm{eff}} := 1/\|w\|_2^2 \ge K e^{-4q}$. Writing $s(w) := K\|w\|_2^2$ for the skewness of $w$ relative to the uniform weighting, we have $s(w) \le e^{4q}$, equivalently $\chi^2(w\,\|\,u) = s(w)-1 \le e^{4q}-1$.
	\end{lemma}
	
	\begin{proof}
		$c_k \in [-1,1]$ gives $\exp(q c_k) \in [e^{-q}, e^{q}]$ and
		$\sum_j \exp(q c_j) \in [K e^{-q}, K e^{q}]$; combining the extremes gives the
		two-sided bound. Then $\|w\|_2^2 \le K\max_k w_k^2 \le e^{4q}/K$, and
		$\chi^2(w\,\|\,u)+1 = K\|w\|_2^2 = s(w) \le e^{4q}$.
	\end{proof}
	
	For $K \ge 2$ and finite $q$, Lemma~\ref{lem:bounded} asserts that FedWAvg constitutes a \emph{strictly interior} convex combination: no client is ever excluded, and the aggregate never reduces to a single client. This characteristic differentiates it from trust-score rules employing hard clipping, such as the ReLU-clipped cosine similarity described by \cite{cao@2021@fltrust}, which have the potential to nullify a benign yet divergent client. The bound maintains the same form across all values of $K$, thereby ensuring that this property remains intact as the federation expands.
	
	\begin{lemma}[Deviation from FedAvg]\label{lem:dev}
		Let $\Delta_c := \max_k c_k - \min_k c_k$ and let $u = (1/K)\mathbf{1}$. Then
		\begin{equation}
			\|w - u\|_1 \le \min\{2,\; q\,\Delta_c\} .
			\label{eq:l1dev}
		\end{equation}
		Consequently, writing $\bar\theta_t := \frac{1}{K}\sum_k \theta_t^k$ and
		setting $R_t := \max_k\|\theta_t^k - \bar\theta_t\|$,
		$\Gamma := \max_k \sup_\theta |F_k(\theta) - F(\theta)|$ and
		$F_w(\theta) := \sum_k w_k F_k(\theta)$,
		\begin{align}
			\big\|\theta_t^{\mathrm{FedWAvg}} - \theta_t^{\mathrm{FedAvg}}\big\|
			&\le \min\{2, q\Delta_c\}\, R_t , \label{eq:paramdev}\\
			\big|F_w(\theta) - F(\theta)\big|
			&\le \min\{2, q\Delta_c\}\,\Gamma . \label{eq:objdev}
		\end{align}
	\end{lemma}
	
	\begin{proof}
		Let $w(t) = \mathrm{softmax}(t\,c)$ for $t\in[0,q]$. Then
		$dw_k/dt = w_k(c_k - \bar c_{w})$ with $\bar c_w = \sum_j w_j c_j$, so
		$\sum_k |dw_k/dt| = \sum_k w_k |c_k - \bar c_w| \le \Delta_c$, because
		$\bar c_w \in [\min_k c_k, \max_k c_k]$. Integrating over $[0,q]$ gives
		$\|w(q)-w(0)\|_1 \le q\Delta_c$, and $2$ is the $\ell_1$ diameter of the
		simplex. For the second part, $\sum_k (w_k - 1/K) = 0$ permits centring:
		$\sum_k (w_k - 1/K)\theta_t^k = \sum_k (w_k - 1/K)(\theta_t^k - \bar\theta_t)$,
		and H\"older's inequality gives the result; the argument for $F_w - F$ is
		identical, using Eq.~\eqref{eq:equalsize} so that $F = \sum_k u_k F_k$.
	\end{proof}
	
	Eq.~\eqref{eq:objdev} compares objective \emph{values}. The analysis of a first-order method requires the corresponding statement for gradients, which Assumption~\ref{as:diss} supplies at no additional cost.
	
	\begin{lemma}[Gradient deviation]\label{lem:graddev}
		Under Assumption~\ref{as:diss} and Eq.~\eqref{eq:equalsize}, for every $\theta$,
		\begin{equation}
			\big\|\nabla F_w(\theta) - \nabla F(\theta)\big\|
			\;\le\; \kappa \, \|w-u\|_1
			\;\le\; \kappa \min\{2,\, q\Delta_c\} .
			\label{eq:graddev}
		\end{equation}
	\end{lemma}
	
	\begin{proof}
		Since $\sum_k (w_k - u_k) = 0$, we may centre:
		$\nabla F_w - \nabla F = \sum_k (w_k-u_k)\nabla F_k
		= \sum_k (w_k-u_k)\big(\nabla F_k - \nabla F\big)$.
		Applying the triangle inequality and Assumption~\ref{as:diss} termwise gives
		$\|\nabla F_w - \nabla F\| \le \sum_k |w_k-u_k| \,\kappa = \kappa\|w-u\|_1$,
		and Lemma~\ref{lem:dev} bounds $\|w-u\|_1$.
	\end{proof}
	
	\begin{proposition}[Stationarity transfer]\label{prop:transfer}
		Let $w^{(t)}$ denote the FedWAvg weights at round $t$ and let $F_{w^{(t)}}$ be the correspondingly weighted objective. Suppose an optimization procedure produces iterates $\{\theta_t\}_{t<T}$ satisfying
		$\min_{t<T}\mathbb{E}\|\nabla F_{w^{(t)}}(\theta_t)\|^2 \le \mathcal{S}(T)$
		for some $\mathcal{S}(T)$. Then, under Assumptions~\ref{as:diss} and~\ref{as:inner} and Eq.~\eqref{eq:equalsize},
		\begin{equation}
			\min_{t<T}\mathbb{E}\big\|\nabla F(\theta_t)\big\|^2
			\;\le\; 2\,\mathcal{S}(T) \;+\; 2\big(\kappa \min\{2,q\Delta_c\} + \zeta\big)^2 .
			\label{eq:transfer}
		\end{equation}
	\end{proposition}
	
	\begin{proof}
		\vspace{-1mm}
		For any $t$, $\|\nabla F(\theta_t)\|^2 \le 2\|\nabla F_{w^{(t)}}(\theta_t)\|^2 + 2\|\nabla F_{w^{(t)}}(\theta_t) - \nabla F(\theta_t)\|^2$ by the elementary inequality $\|a+b\|^2 \le 2\|a\|^2 + 2\|b\|^2$. Lemma~\ref{lem:graddev} bounds the second term by $\kappa\min\{2,q\Delta_c\}$, and Assumption~\ref{as:inner} adds at most $\zeta$ to the gradient actually computed. Taking the minimum over $t<T$ gives Eq.~\eqref{eq:transfer}.\vspace{-1mm}
	\end{proof}
	The conditional aspect of Proposition~\ref{prop:transfer} is purposeful; it indicates that any stationarity guarantee for the round-weighted objective also applies to the uniform target, with an extra cost of $O\big((\kappa\min\{2,q\Delta_c\}+\zeta)^2\big)$. This additional cost vanishes as $q\to0$, as shown in Corollary~\ref{cor:fedavg}. However, it does not establish $\mathcal{S}(T)$, and we do not make such a claim.
	
	\vspace{-4mm}
	\subsubsection{Adversarially robust generalization}
	\vspace{-1mm}
	We present the generalization result as a \emph{reduction}, allowing it to inherit any uniform-convergence bound available for the adversarial loss class without relying on a specific one. Let
	$R_{\mathrm{adv}}(\theta;\mathcal{D}) :=
	\mathbb{E}_{\mathcal{D}}[\max_{\|\varepsilon\|_p\le\varepsilon_0}
	\mathcal{L}(\theta,x+\varepsilon,y)]$, let the target be the mixture
	$\mathcal{D}_p = \sum_k p_k \mathcal{D}_k$ with $p=u$ by Eq.~\eqref{eq:equalsize}, let
	$\widehat R^{\,w}_{\mathrm{adv}}$ be the $w$-weighted empirical robust risk over
	$N=\sum_k n_k$ samples, and let $\mathcal{L}\in[0,B]$. \vspace{-1mm}
	
	\begin{theorem}[Price of the FedWAvg weighting]\label{thm:gen}
		Suppose the adversarial loss class admits a uniform-convergence bound
		$R_{\mathrm{adv}}(\theta;\mathcal{D}_p) \le
		\widehat R^{\,v}_{\mathrm{adv}}(\theta) + \mathcal{C}_v(N,\delta)$,
		holding with probability at least $1-\delta$ for a weighting $v$, whose deviation term satisfies
		$\mathcal{C}_v(N,\delta) \le \sqrt{s(v)}\,\mathcal{C}_u(N,\delta)$
		in the skewness $s(v) = K\|v\|_2^2$ of Lemma~\ref{lem:bounded}. Then the bound for the
		FedWAvg weighting satisfies
		\begin{align}
			R_{\mathrm{adv}}(\theta;\mathcal{D}_p) \;\le\;
			& \widehat R^{\,w}_{\mathrm{adv}}(\theta)
			+ e^{2q}\,\mathcal{C}_u(N,\delta) \nonumber\\
			& + \min\{2,\, q\Delta_c\}\,B .
			\label{eq:gen}
		\end{align}
	\end{theorem}
	
	\begin{proof}
		By Lemma~\ref{lem:bounded}, $s(w) \le e^{4q}$, so
		$\mathcal{C}_w \le \sqrt{s(w)}\,\mathcal{C}_u \le e^{2q}\mathcal{C}_u$, while the uniform
		weighting has $s(u) = K\|u\|_2^2 = 1$ and hence $\mathcal{C}_u$ unchanged. The residual term is the mixture mismatch: by
		the argument of Lemma~\ref{lem:dev} applied to the local robust risks,
		$|\sum_k (w_k - u_k) R_{\mathrm{adv}}(\theta;\mathcal{D}_k)|
		\le \|w-u\|_1 B \le \min\{2,q\Delta_c\}B$.\vspace{-1mm}
	\end{proof}
	
	Theorem~\ref{thm:gen} intentionally provides a weaker statement than a bound specific to a hypothesis class, and is correspondingly more robust: it prices only the \emph{weighting}, which is the quantity under our control, and holds for any admissible complexity measure. As $q\to0$ the multiplicative factor $e^{2q}\to 1$ and the additive term $\min\{2,q\Delta_c\}B\to0$, so the \emph{additional cost} of the weighting vanishes and the bound reduces continuously to the FedAvg bound, consistent with Corollary~\ref{cor:fedavg}. Neither term tends to zero individually, and we make no such claim.
	
	\begin{remark}[The role of $q$, and a prediction]\label{rem:q}
		Theorem~\ref{thm:contract} and Theorem~\ref{thm:gen} together determine the trade-off governed by the scale factor. Increasing $q$ confers an advantage through the drift contraction of Eq.~\eqref{eq:gap}, which grows with $\operatorname{Var}_w(d)$, that is, with the degree of heterogeneity. The associated cost is the skewness penalty of Lemma~\ref{lem:bounded} entering Eq.~\eqref{eq:gen} through $e^{2q}$, together with the mixture-mismatch term, both of which grow with $q$ irrespective of heterogeneity. The optimal $q$ therefore increases with skew, which predicts that FedWAvg($10$) should outperform FedWAvg($1$) primarily under highly skewed partitioning while exhibiting no consistent advantage at moderate skew. This is the pattern observed in \cref{tab:niid4,tab:niid2}.
	\end{remark}
	
	\begin{remark}[Precedent]\label{rem:precedent}
		Five independent studies support this family of aggregation weights.   \cite{rehman@2023@ldawa}  by Rehman \emph{et al.} uses the angular divergence between client and global weights as a layer-wise aggregation weight, demonstrating that it reduces client bias similarly to momentum. \emph{Cao et al.} \cite{cao@2021@fltrust} establish a bound on the distance between a cosine-similarity-weighted global model and the attack-free optimal solution. Ye \emph{et al.} \cite{ye@2023@feddisco} show that aggregation weights based on local-global discrepancy provide a strictly tighter bound on optimization error than those based on dataset size-aligning with the principle that FedWAvg employs an angular discrepancy on the classification head. Li \emph{et al.} \cite{li@2023@fedlaw} arrive at a similar conclusion from the opposite angle, indicating that dataset-size weights are sub-optimal and that client coherence should guide weighting decisions. Specifically, within the FAT context, Zhu \emph{et al.} \cite{zhu@2023@combating} also reweight clients during aggregation to mitigate adversarially amplified heterogeneity.
	\end{remark}
	
	\begin{remark}[Scope]\label{rem:scope}
		Theorem~\ref{thm:contract} applies independently of dataset and architecture, as it pertains to the aggregation rule itself. The only element that depends on the dataset is the \emph{magnitude} of the benefit, which, according to Eq.~\eqref{eq:gap}, increases with the dispersion of client head drift. Sec.~\ref{sec:scope} explores the consequences for larger label spaces and higher values of $K$. Moreover, FedWAvg functions as a heterogeneity-aware aggregation rule rather than a Byzantine-robust one: Lemma~\ref{lem:bounded} ensures that no client ever receives a weight less than $e^{-2q}/K$, allowing a sufficiently adversarial participant to maintain influence.
	\end{remark}
	
	\vspace{-5mm}
	\subsubsection{What a convergence theorem would additionally require}
	\label{sec:openproblems}
	\vspace{-1mm}
	We make it clear that certain points are not established in this case, since the aggregation weights of Eq.~\eqref{eq:weights_cos} are adaptive and a convergence analysis for FedWAvg cannot therefore be derived from an argument based on fixed weights. Three obstacles must be addressed.
	
	First of all, the weights are \emph{time-varying}: the value $w^{(t)}$ is determined by the global and local heads in round $t$, so the weighted objective $F_{w^{(t)}}$ varies from round to round and as a result the telescoping sum used in a standard descent argument does not close. To address this, a bound on the round-to-round change in the weights, namely on $\sum_t \|w^{(t+1)}-w^{(t)}\|_1$, is needed, which we do not derive.
	
	Second, the weights are \emph{parameter-dependent}. Should the fixed objective be defined as $\widetilde F(\theta) = \sum_k w_k(\theta)F_k(\theta)$, then $\nabla \widetilde F = \sum_k w_k \nabla F_k + \sum_k F_k \nabla w_k$, whereas Algorithm~\ref{alg:algo_fedwavg} only implements the first of these sums. FedWAvg is thus not gradient descent with respect to $\widetilde F$, and the term that is omitted must be bounded.
	
	Third, the min-max structure has to be dealt with. Assumption~\ref{as:inner} introduces a gradient-level inexactness $\zeta$; in order to obtain such a bound from a value-level PGD guarantee, additional structure is needed on the inner problem, usually in the form of strong concavity in $\varepsilon$ or a Polyak-{\L}ojasiewicz condition.
	
	One cannot simply refer to existing analyses in order to overcome these difficulties. The convergence result for federated adversarial learning given in \cite{li@2023@federated} applies only to \emph{uniform} FedAvg aggregation in the over-parameterized two-layer ReLU setting, on the assumption of data separability and with a particular random initialization, and provides a bound on the robust training loss rather than on a gradient norm; the authors themselves regard the extension of their results to other aggregation rules as something for future research. We therefore quote this work as showing that a convergence analysis of FAT is in principle feasible, not as indicating that such an analysis applies to FedWAvg. The two main theoretical issues that we leave open are to establish a convergence theorem for adaptive similarity-based aggregation under the min-max objective and to connect the head-drift contraction stated in Theorem~\ref{thm:contract} to robust accuracy.
	
	\color{black}
	
	\vspace{-2.5mm}
	\subsection{Vision Transformer Models}
	\label{sec:vitmodels}
	\vspace{-1mm}
	Considering the model architecture is one approach to optimizing the \ac{fl} process. To the best of the authors’ knowledge, this is the first paper to investigate the robustness and performance of \ac{vit} models and their variants, using different patch embeddings and classification head techniques, in \acf{fat} settings. Further details on vision transformers are provided in \cref{app:vits_method}.

	\begin{figure*}[!t]
		\vspace{-4mm}
		\centering
		\caption{The calculated aggregation weights using cosine similarity, \cref{eq:weights_cos}, during the federated training process using \ac{tnt}-S model and \ac{niid} partitioning (each client has 4 classes). a) for the natural training, and b) for the adversarial training. More examples are in \cref{app:more_cosine_sim_w}.}
		\resizebox{0.80\textwidth}{!}{%
			\setlength\tabcolsep{1.5pt}
			\begin{tabular}{c}
				\begin{tabular}{l}\subfloat[]{\includegraphics[width=\textwidth]{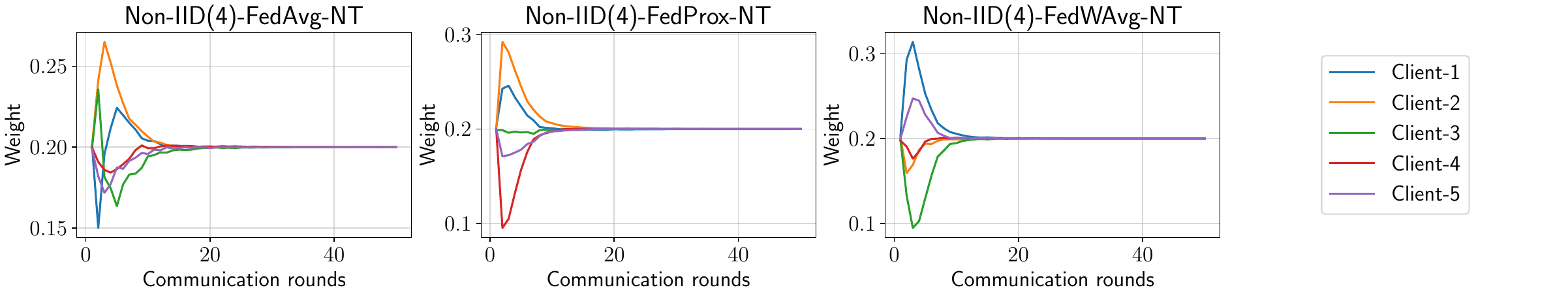}\label{fig:wieghts_tnts_cls_b_nt}}\end{tabular}  \\
				\begin{tabular}{l}\subfloat[]{\includegraphics[width=\textwidth]{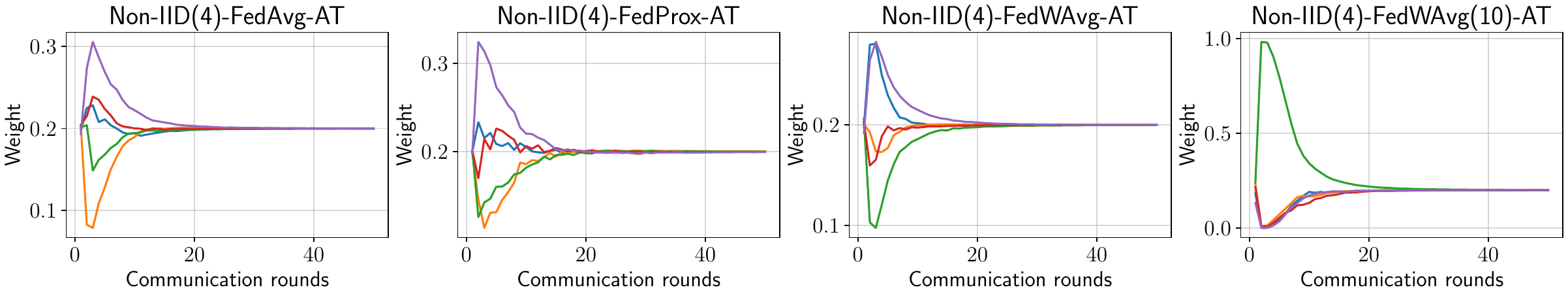}\label{fig:wieghts_tnts_cls_b_at}} \end{tabular} \\
			\end{tabular}
		}%
		\vspace{-4mm}
		\label{fig:wieghts_tnts_cls}
	\end{figure*}
	
	\vspace{-2mm}
	\section{Experiments}
	\vspace{-2mm}
	\subsection{Experiment Setup} 
	\label{sec:setup}
	\vspace{-1mm}
	We developed our code on top of the FedTorch \cite{haddadpour@2021@compression} library. It is an open-source Python package for distributed and federated training of machine learning models. 
	
	\textbf{Dataset.} In our experiments, we use CIFAR-10 dataset\footnote{Available at \url{https://www.cs.toronto.edu/~kriz/cifar.html}} to investigate different visual transformer models under the \ac{fat} environment. CIFAR-10 is a collection of images that is usually used in computer vision tasks. It comprises $32\times 32$ RGB images across 10 classes: airplanes, cars, birds, cats, deer, dogs, frogs, horses, ships, and trucks. It contains 60,000 images, 50,000 for training and 10,000 for testing. The training images are processed by resizing to 224$\times$224, applying random cropping with a 28-pixel padding, and random horizontal flipping. The 10,000-image test dataset is used as the global test set.
	Since each image is resized to $224\times224$ before patch embedding, every model works at its native token resolution, comprising 196 patches of $16\times16$ plus a class token for ViT-S/16 and ViT-B/16, and the equivalent tokens for T2T and TNT embeddings. All backbones are initialized from ImageNet-pretrained weights. Consequently, the self-attention is performed over the same token sequence length as in ImageNet training, allowing the transformer's global-context pathway to be fully utilized. The limitations posed by CIFAR-10 lie in the information content of the upsampled image, not in the sequence length or receptive field. Sec.~\ref{sec:scope} explores how this can be extended to datasets with larger label spaces.
	
	\textbf{Models.} As mentioned earlier in \cref{sec:related_work}, we investigate the vision transformer models with three different embedding methods and three different classification head methods. \cref{fig:fat_models} in \cref{app:vits_method} illustrates the models' architectures. For \ac{vit} embedding, we use \ac{vit}-S-16 and \ac{vit}-B-16 models. For \ac{t2t} embedding, we use \ac{t2t}-14 and finally, for \ac{tnt} embedding, we use \ac{tnt}-S model. It was found in \cite{aldahdooh@2025@beyond} that some small vision transformer architectures, such as \ac{vit}-S and \ac{tnt}-S, exhibit greater robustness than larger architectures. For these four models, three classification heads are used: the first uses only the [CLS] token, the second uses only the visual [VIS] tokens, and the third uses both the [CLS] and [VIS] tokens. In total, we tested 12 vision transformer models. 
	
	\textbf{\Acf{fat} setting.} Following the setup of \cite{qu@2022@rethinking}, we assume that we have a server and $K=5$ available clients, and each client will use Madry's \ac{at} procedure with adversarial ratio of 0.5, i.e., for each batch in each training epoch 50\% of the clean samples are replaced with adversarial samples. For the \ac{pgd} attack, we set the perturbation budget to $\epsilon=8/255$, the step size to $\alpha=2/255$, and the steps to $s=7$. We investigate three data partitioning methods. The first setting is the \ac{iid} setting, in which the training data is evenly distributed across the clients. The second setting is the \ac{niid}(4) setting, in which each client is assigned data from 4 classes only. While the third partitioning method is \ac{niid}(2), in which each client is assigned data from 2 classes only.
	
	\textbf{Regime.} Our focus is on the \emph{cross-silo} setting, where a limited number of consistent institutional clients participate in each round. This contrasts with the cross-device regime, which involves over $10^3$ intermittently available, resource-limited clients \cite{kairouz@2021@advances}. This choice is driven by scope rather than convenience: as shown in Sec.~\ref{sec:cost}, training a FAT of a 22-86 million-parameter vision transformer increases local computation by roughly $4.5$ times compared to natural training, and demands 82-330~MiB of uplink per client per round, characteristics typical of cross-silo workloads. We use a fixed $K=5$ with full participation, aligning with \cite{qu@2022@rethinking}, which studied vision transformers in Non-IID FL. By maintaining their client count, partitioning, learning rates, and round budget, our results directly compare to theirs, attributing differences to adversarial training rather than setup changes. Note that Algorithm~\ref{alg:algo_fedwavg} samples $m=\max(1,CK)$ clients each round, and the softmax in FedWAvg is normalized over the sampled set $S_t$, so the method remains unchanged under partial participation. Sec.~\ref{sec:scope} discusses extending to larger values of $K$.
	
	\textbf{\Ac{fl} aggregation methods.} As mentioned earlier in Sec.~\ref{sec:related_work}, we investigate the server-based \ac{fl} paradigms that produce a global model. Hence, we compare the proposed FedWAvg algorithm with the other four \stateart algorithms: FedAvg, FedProx, FedGate, and SCAFFOLD. For FedProx, the proximal control parameter $\mu$ is set to 0.1 as tuned in \cite{qu@2022@rethinking}.

	\begin{figure*}[!t]
		\vspace{-4mm}
		\centering
		\caption{The accuracy and robust accuracy of \ac{tnt}-CLS model in the \ac{fat} process with different aggregation methods. The accuracies against the communication rounds under the \ac{nt} process using \ac{iid}, \ac{niid}(4), and \ac{niid}(2) are shown in a), b), and c) respectively. The accuracies (top) and the robust accuracies (bottom) against the communication rounds under the \ac{at} process using \ac{iid}, \ac{niid}(4), and \ac{niid}(2) are shown in d), e), and f) respectively. More figures can be found in \cref{app:more_convergence}.}
		\resizebox{0.9\textwidth}{!}{%
			\setlength\tabcolsep{1.5pt}
			\begin{tabular}{ccc}
				\begin{tabular}{l}\subfloat[]{\includegraphics[width=0.33\textwidth]{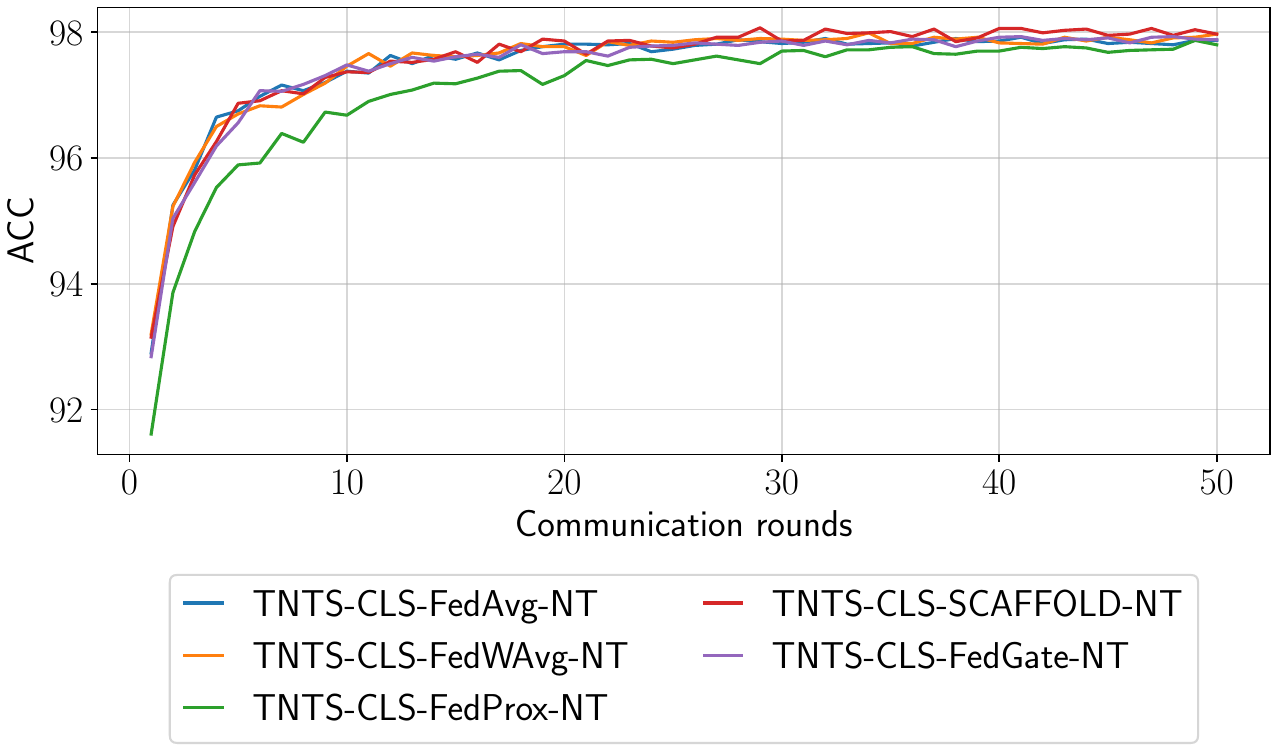}\label{fig:conv_tnts_cls_iid_nt}}\end{tabular} &
				\begin{tabular}{l}\subfloat[]{\includegraphics[width=0.33\textwidth]{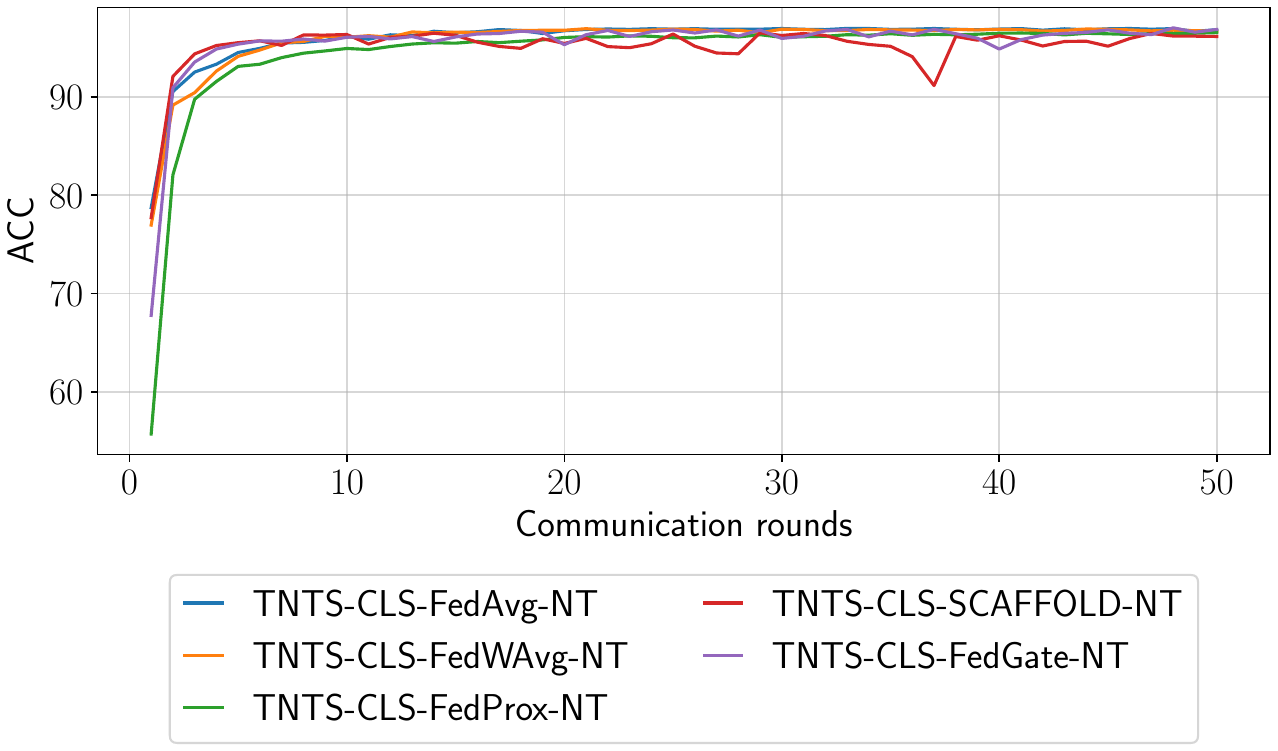}\label{fig:conv_tnts_cls_niid4_nt}} \end{tabular} &
				\begin{tabular}{l}\subfloat[]{\includegraphics[width=0.33\textwidth]{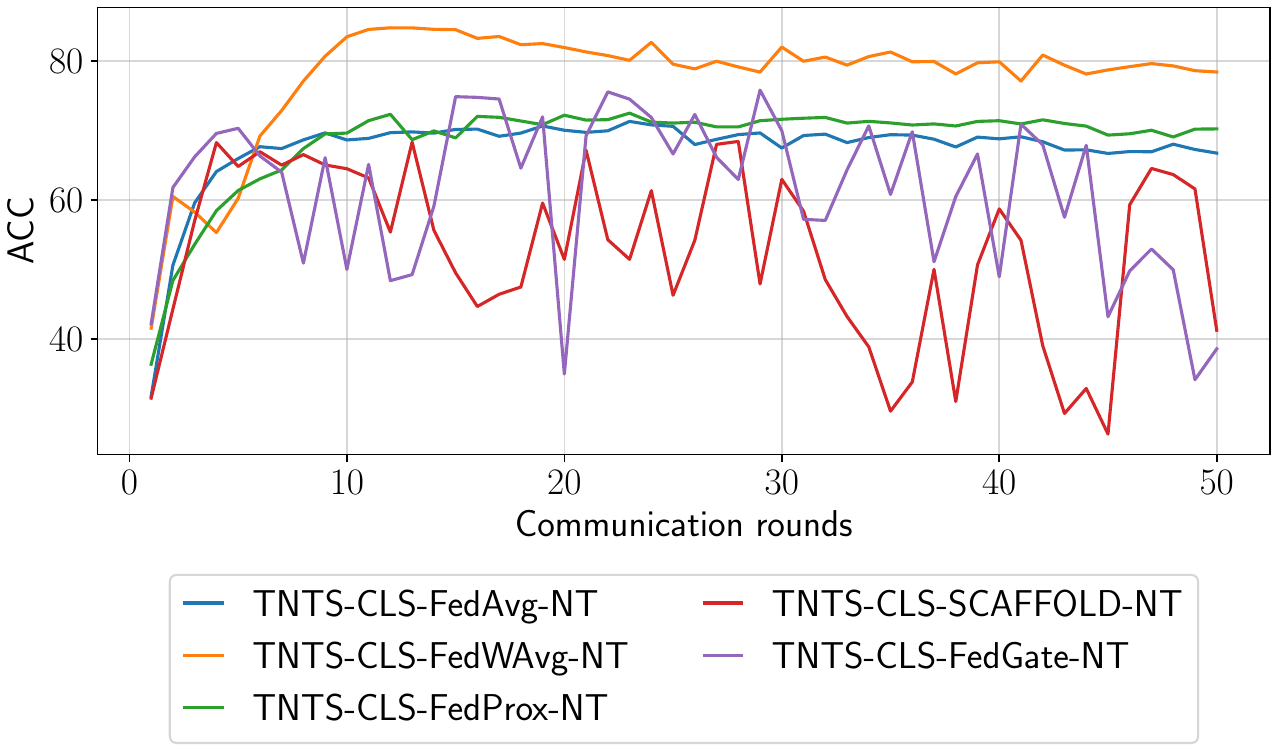}\label{fig:conv_tnts_cls_niid2_nt}} \end{tabular} \\
				
				\begin{tabular}{l}\subfloat[]{\includegraphics[width=0.33\textwidth]{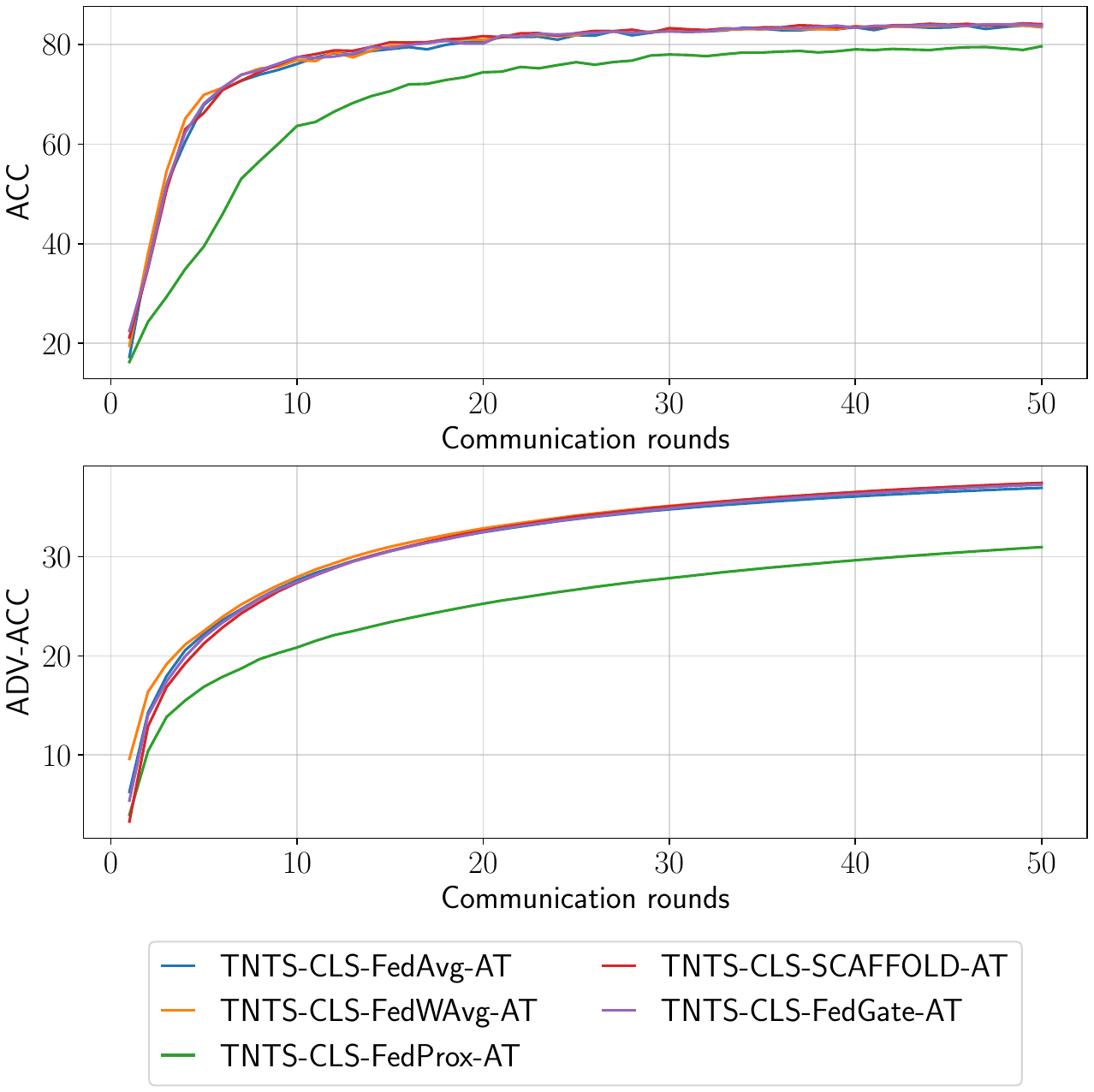}\label{fig:conv_tnts_cls_iid_at}}\end{tabular} &
				\begin{tabular}{l}\subfloat[]{\includegraphics[width=0.33\textwidth]{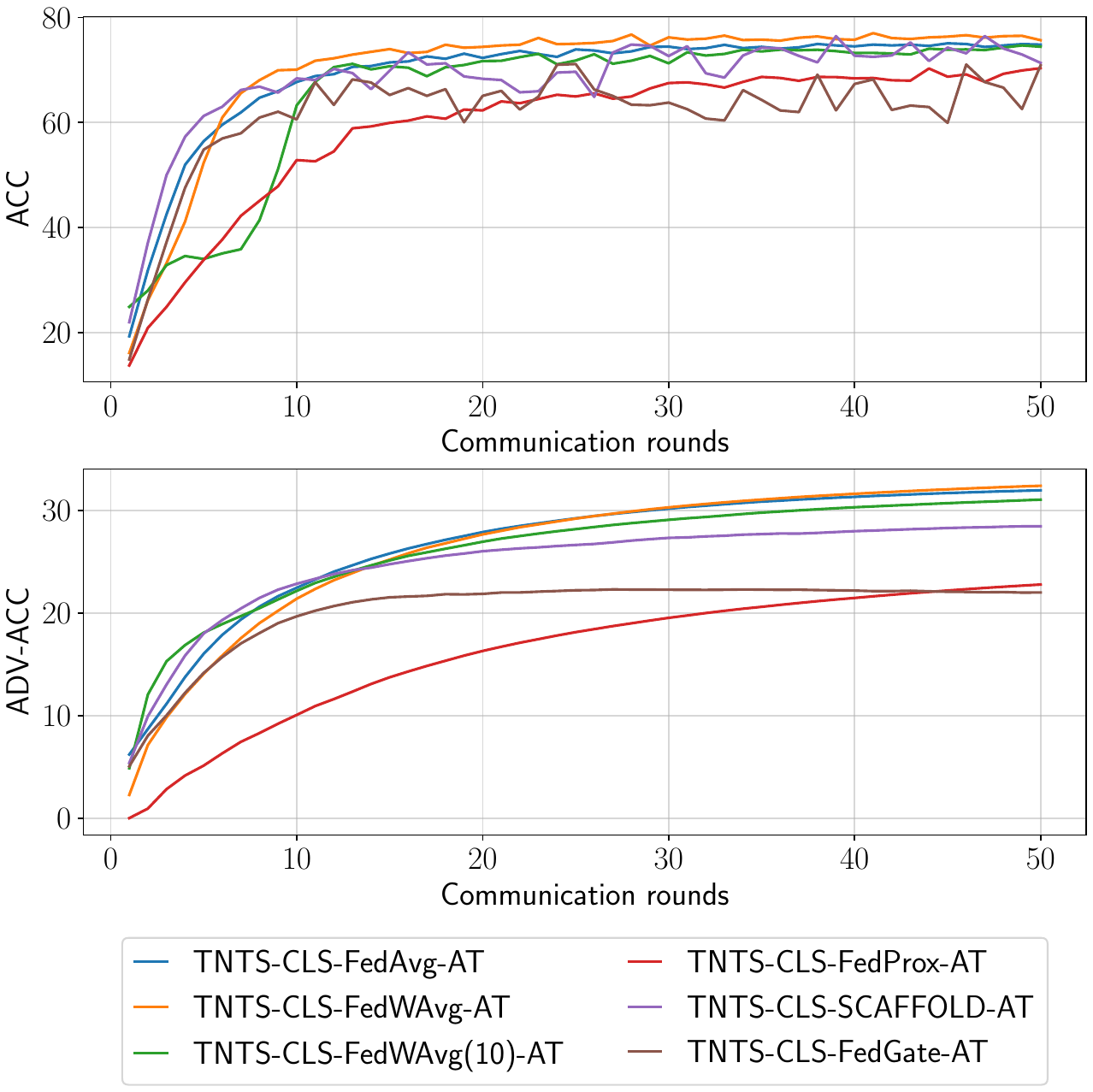}\label{fig:conv_tnts_cls_niid4_at}} \end{tabular} &
				\begin{tabular}{l}\subfloat[]{\includegraphics[width=0.33\textwidth]{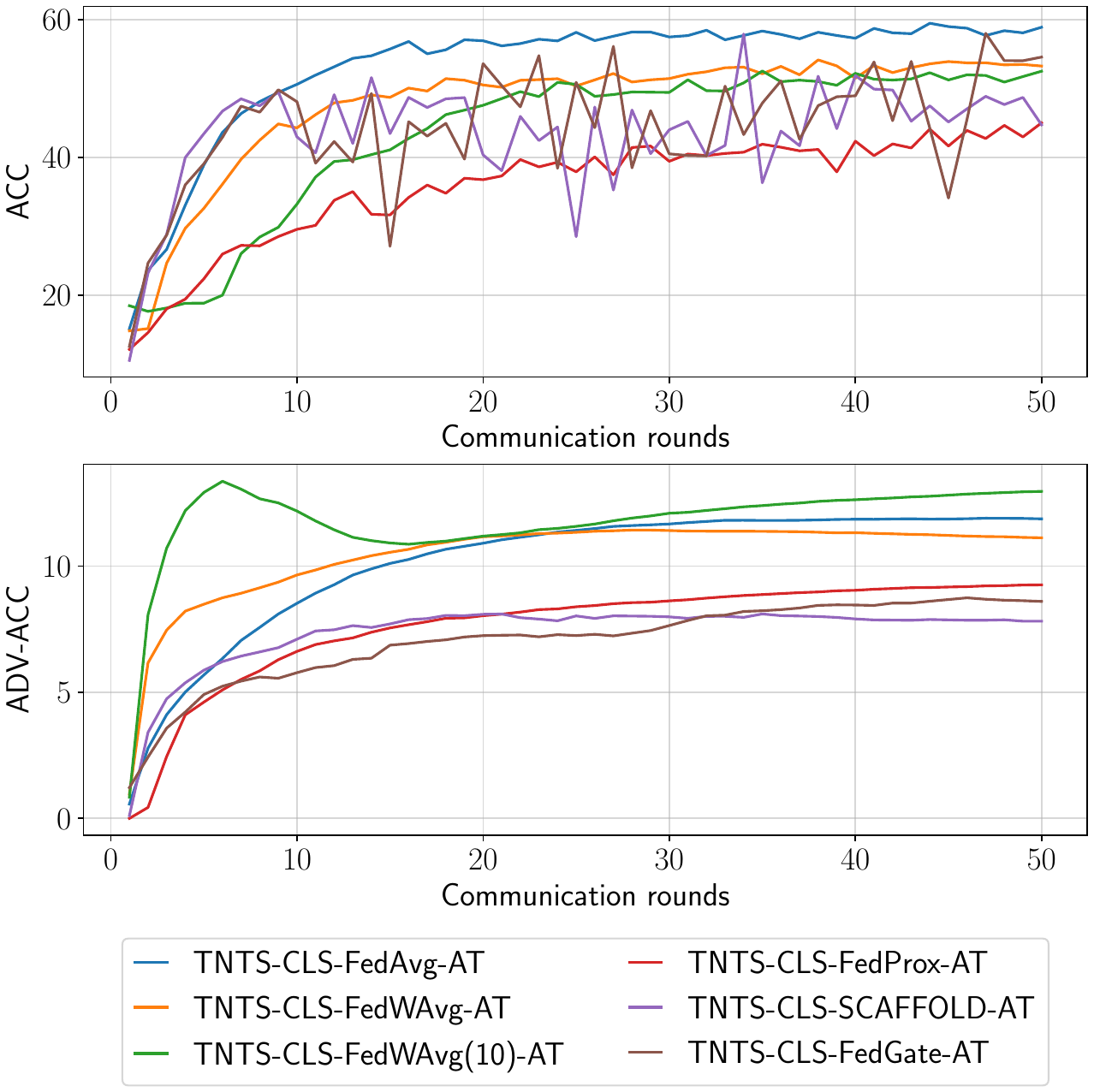}\label{fig:conv_tnts_cls_niid2_at}} \end{tabular} \\
				
			\end{tabular}
		}%
		\vspace{-4mm}
		\label{fig:conv_tnts_cls}
	\end{figure*}
	
	\textbf{Training configuration.} The server first initializes the model with the weights from the pre-trained model that is trained using the ImageNet dataset\footnote{The weights are available \href{https://github.com/rwightman/pytorch-image-models}{here}, for all models except for \acp{t2t} which are available \href{https://github.com/yitu-opensource/T2T-ViT}{here}.}. Due to the GPU limited memory, we set the batch size to 24. The \ac{sgd} algorithm is used as optimizer with the momentum set to 0.9. In \ac{nt}, the learning rate is set to 0.03 as tuned in \cite{qu@2022@rethinking}, while in \acf{at}, the learning rate is set to 0.1. The learning rate is decreased in every epoch by 3.5\% in the \ac{nt} and by 5\% in the \ac{at}. It was shown that \ac{vit} model converges faster than ResNet-50, hence, we set the number of communication rounds $T$ to 50 \cite{qu@2022@rethinking}. Moreover, for highly heterogeneous data partitions \cite{qu@2022@rethinking}, it is recommended to set the number of local epochs $E$ to a small value ($\leq$5); in our experiments, we set $E$ to 1.
	
	\begin{figure*}[!t]
		\vspace{-4mm}
		\centering
		\caption{The \acs{svcca} for the first and the ninth layer of server, client 1, and client 4 models against communication rounds under \ac{fat} process using \ac{tnt} model with \ac{niid}(4). a) using clean test samples and b) using adversarial test samples. The top row shows the \acs{svcca} between client 1 and client 2, the middle row shows the \acs{svcca} between the server and client 1, and the bottom row shows the \acs{svcca} between the server and client 4. More figures can be found in \cref{app:more_model_drift}.}
		\resizebox{0.9\textwidth}{!}{%
			\setlength\tabcolsep{1.5pt}
			\begin{tabular}{cc}
				\begin{tabular}{l}\subfloat[With clean samples]{\includegraphics[width=0.5\textwidth]{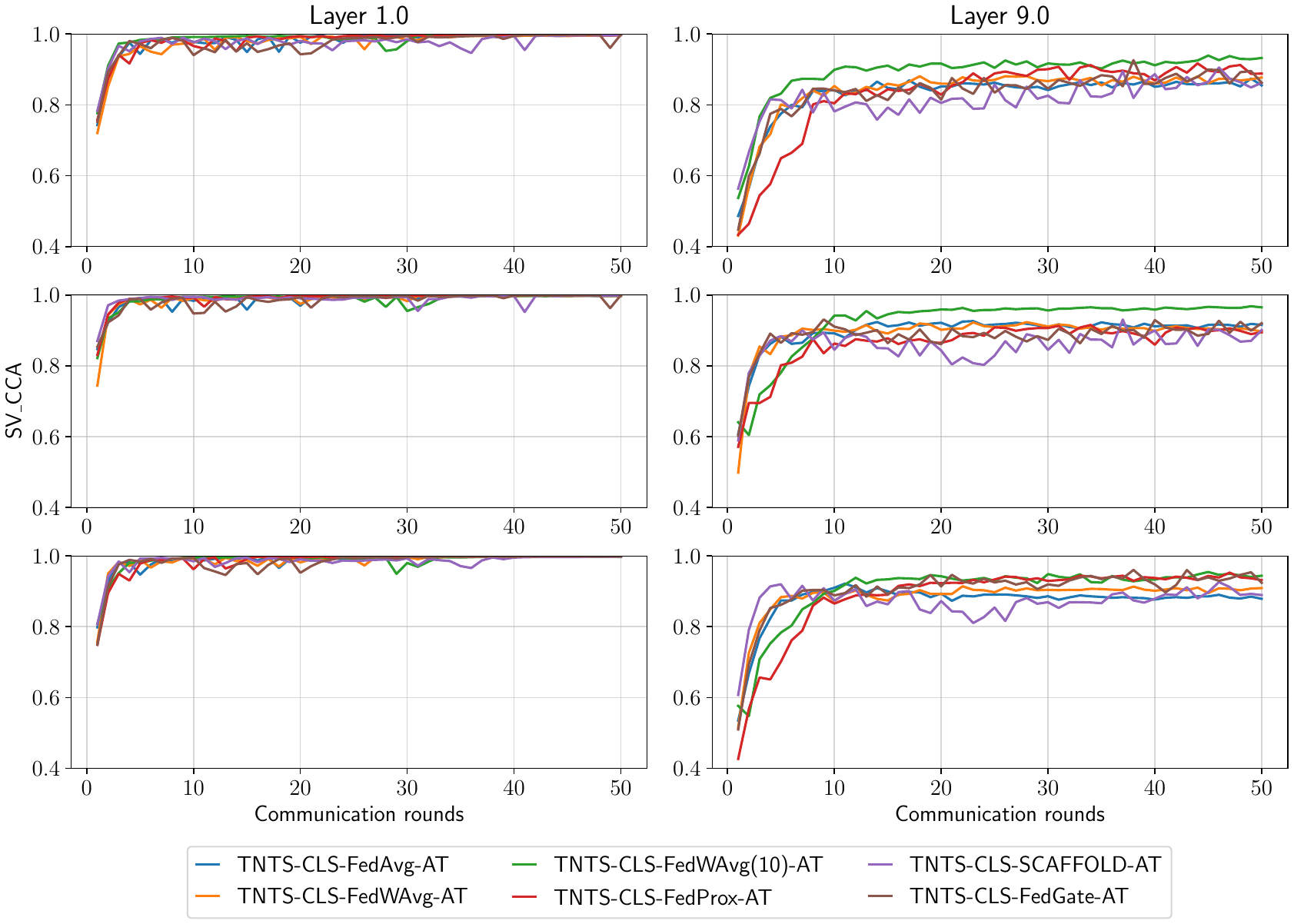}\label{fig:drft_tnts_cls_iid4_clean_at}}\end{tabular} &
				\begin{tabular}{l}\subfloat[With adversarial samples]{\includegraphics[width=0.5\textwidth]{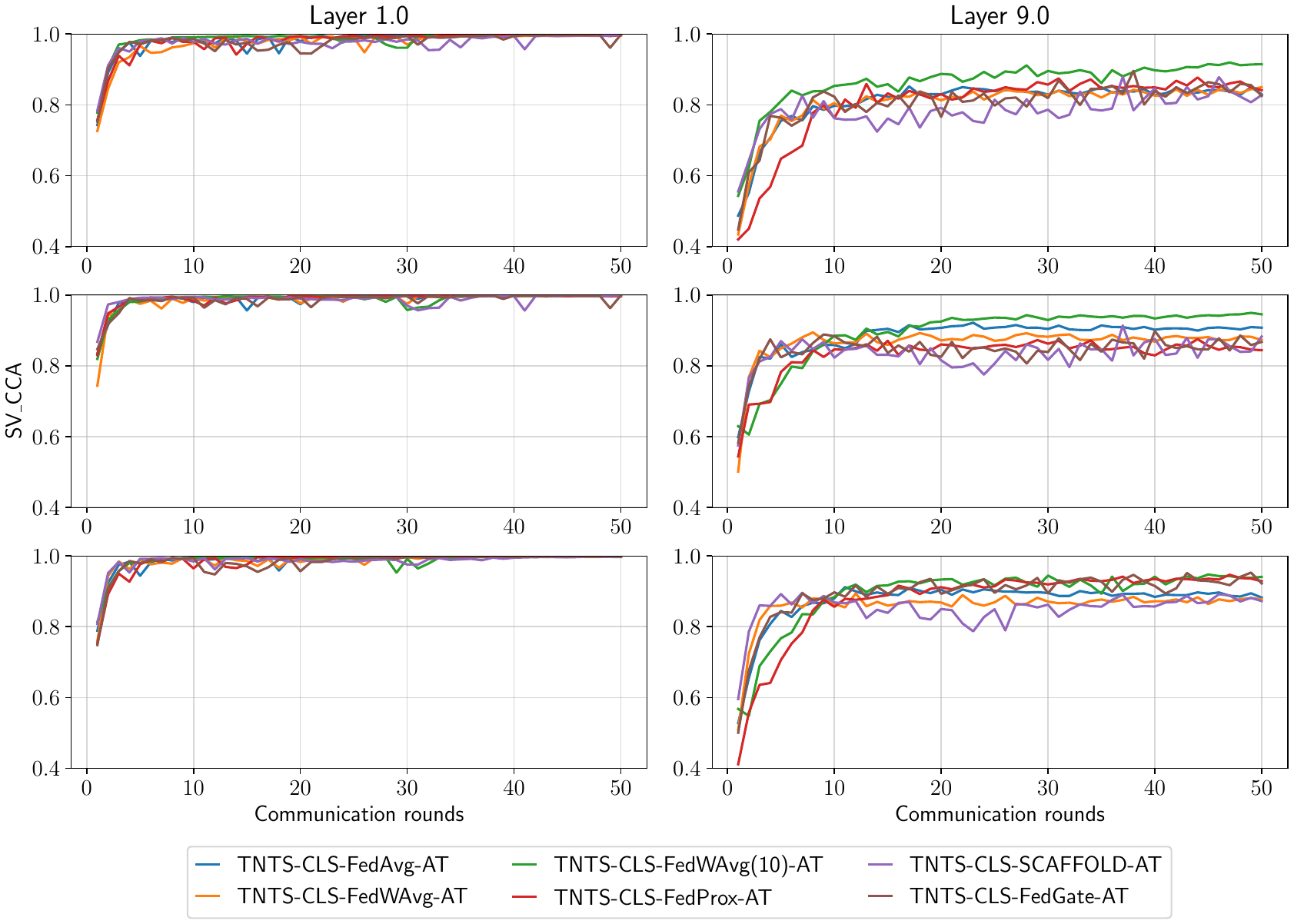}\label{fig:drift_tnts_cls_niid4_adv_at}} \end{tabular} \\
			\end{tabular}
		}%
		\vspace{-3mm}
		\label{fig:drift_tnts_cls_iid4_at}
	\end{figure*}
	
	\vspace{-3.5mm}
	\subsection{Results and Discussions}
	\label{sec:results}

	\begin{table*}[!htb]
		\centering
		\caption{Models accuracy under \ac{nt} and models (accuracy, robust accuracy) under \ac{at} in \ac{fl} with \ac{iid}. The \textbf{first} and the \underline{second} best robust accuracy are marked.}
		\label{tab:iid}
		\resizebox{\textwidth}{!}{%
			\begin{tabular}{|c|c|c||c|c|c|c|c|c|c|c|c|c|}
				\hline
				\multicolumn{3}{|c||}{Architecture}                                                                                      & \multicolumn{2}{c|}{FedAvg}                  & \multicolumn{2}{c|}{FedProx}                 & \multicolumn{2}{c|}{FedGate}                  & \multicolumn{2}{c|}{SCAFFOLD}                & \multicolumn{2}{c|}{FedWAvg(1)}               \\ \hline
				
				Token                        & Model                  & Class. Head &NT    & AT              & NT    & AT              & NT    & AT               & NT    & AT              & NT    & AT               \\ \hline
				
				\multirow{6}{*}{\begin{tabular}[c]{@{}l@{}}Image\\ patch\end{tabular}} & \multirow{3}{*}{VIT-S}  & CLS                 & 97.35   & (82.73, 35.57)   & 97.14    & (77.96, 28.82)   & 97.35    & (82.91, \textbf{35.76})   & 97.46    & (83.01, \underline{35.71})    & 97.33     & (82.91, 35.67)     \\ \cline{3-13} 
				&                         & VIS                 & 97.31   & (81.04, 34.9)    & 97.2     & (75.70, 28.06)   & 97.31    & (81.55, \underline{35.13})   & 97.35    & (81.31, \textbf{35.16})    & 97.26     & (81.26, 34.79)                  \\ \cline{3-13} 
				&                         & CLS+VIS             & 97.25   & (82.9, 35.82)    & 97.12    & (78.05, 29.03)   & 97.36    & (83.11, \textbf{36.05})   & 97.41    & (82.92, 35.71)    & 97.27     & (82.49, \underline{35.83})     \\ \cline{2-13} 
				& \multirow{3}{*}{VIT-B}  & CLS                 & 98.71   & (96.87, \underline{28.01})   & 98.67    & (97.41, 21.3)    & 98.82    & (97.30, 26.84)   & 98.77    & (97.27, \textbf{28.3})     & 98.76     & (96.98, 25.77)     \\ \cline{3-13} 
				&                         & VIS                 & 98.7    & (97.03, 22.79)   & 98.66    & (98.16, 8.90)                 & 98.74    & (97.12, \textbf{25.63})   & 98.74    & (97.01, \underline{23.23})    & 98.74     &   (96.90, 20.98)                 \\ \cline{3-13} 
				&                         & CLS+VIS             & 98.73   & (97.14, \textbf{29.79})   & 98.71    & (97.22, 20.89)    & 98.81    & (97.18, 27.82)   & 98.81    & (97.27, \underline{29.07})    & 98.81     & (97.04, 28.02)     \\ \hline
				\multirow{3}{*}{T2T}                                                   & \multirow{3}{*}{T2T-14} & CLS                 & 98.14   & (97.79, 11.07)   & 97.89    & (96.87, \textbf{14.2})    & 98.09    & (97.69, 11.79)   & 98.09    & (97.87, 11.04)    & 98.11     & (97.670, \underline{12.41})    \\ \cline{3-13} 
				&                         & VIS                 & 97.99   & (98.05, \underline{9.0})     & 97.94    & (97.72, 7.81)    & 98.07    & (98.14, 8.9)     & 98.08    & (98.15, \textbf{9.37})     & 97.93     & (98.03, 8.67)      \\ \cline{3-13} 
				&                         & CLS+VIS             & 98.07   & (96.97, 14.52)   & 97.97    & (96.97, 10.73)    & 98.09    & (97.37, \underline{15.85})   & 98.07    & (97.13, \textbf{16.07})    & 98.12     & (97.04, 15.49)     \\ \hline
				\multirow{3}{*}{TNT}                                                   & \multirow{3}{*}{TNT-S}  & CLS                 & 97.92   & (83.50, 36.93)   & 97.87    & (79.64, 30.96)   & 97.93    & (83.82, 37.27)   & 98.07    & (84.07, \textbf{37.43})    & 97.97     & (83.57, \underline{37.41})     \\ \cline{3-13} 
				&                         & VIS                 & 97.82   & (81.57, 35.88)   & 97.67    & (74.19, 29.11)   & 97.84    & (82.19, \underline{35.89})   & 97.81    & (82.49, \textbf{36.01})    & 97.79     & (79.95, 34.58)     \\ \cline{3-13} 
				&                         & CLS+VIS             & 97.95   & (80.44, 35.17)   & 97.81    & (75.02, 28.80)   & 98.07    & (82.24, \textbf{36.28})   & 97.99    & (82.27, \underline{36.22})    & 98.02     & (79.27, 34.09)     \\ \hline
			\end{tabular}%
		}
		\vspace{-2mm}
	\end{table*}

	\begin{table*}[!htb]
		\centering
		\caption{Models accuracy under \ac{nt} and models (accuracy, robust accuracy) under \ac{at} in \ac{fl} with \ac{niid}(4). The \textbf{first} and the \underline{second} best robust accuracy are marked.}
		\label{tab:niid4}
		\resizebox{\textwidth}{!}{%
			\begin{tabular}{|c|c|c|c|c|c|c|c|c|c|c|c|c|c|c|}
				\hline
				\multicolumn{3}{|c|}{Architecture}                                             & \multicolumn{2}{c|}{FedAvg} & \multicolumn{2}{c|}{FedProx} & \multicolumn{2}{c|}{FedGate} & \multicolumn{2}{c|}{SCAFFOLD} & \multicolumn{2}{c|}{FedWAvg(1)} & \multicolumn{2}{c|}{FedWAvg(10)} \\ \hline
				
				Token                        & Model                   & Class. Head & NT      & AT               & NT       & AT               & NT       & AT               & NT       & AT                & NT        & AT                 & NT      & AT                    \\ \hline
				
				\multirow{6}{*}{\begin{tabular}[c]{@{}l@{}}Image\\ patch\end{tabular}} & \multirow{3}{*}{VIT-S}  & CLS                 & 96.05   & (75.18, \textbf{29.62})   & 95.62    & (66.85, 22.01)   & 95.16    & (71.04, 21.04)   & 95.41    & (68.34, 20.25)    & 96        & (71.94, 29.21)     & -        & (73.32, \underline{29.37})       \\ \cline{3-15} 
				&                         & VIS                 & 95.98   & (76.17, \textbf{31.09})   & 95.62    & (67.86, 24.09)   & 95.45    & (68.21, 20.89)   & 94.68    & (71.67, 23.83)    & 96.02     & (74.96, \underline{30.43})     & -        & (76.43, 29.4)        \\ \cline{3-15} 
				&                         & CLS+VIS             & 96.07   & (75.91, \textbf{29.9})    & 95.66    & (68.39, 22.27)   & 94.99    & (69.91, 21.38)   & 95.46    & (64.20, 19.69)    & 95.95     & (75.18, \underline{29.57})     & -        & (74.59, 29.2)        \\ \cline{2-15} 
				& \multirow{3}{*}{VIT-B}  & CLS                 & 98.45   & (95.55, 14.41)   & 98.27    & (96.57, 7.47)    & 98.49    & (96.46, 3.97)    & 98.41    & (97.37, 14.50)    & 98.42     & (95.93, \underline{17.6})      & -        & (96.15, \textbf{21.11})       \\ \cline{3-15} 
				&                         & VIS                 & 98.4    & (96.06, 14.09)   & 98.33    &     (96.33, 11.11)             & 98.52    & (96.75, 10.16)   & 98.32    & (97.08, 10.39)    & 98.42     &  (96.09, \textbf{18.63})                  & -        &   (95.870, \underline{17.93})                   \\ \cline{3-15} 
				&                         & CLS+VIS             & 98.33   & (95.87, \underline{19.82})   & 98.25    & (96.35, 11.11)   & 98.32    & (96.92, 11.39)   & 98.51    & (96.91, 7.61)     & 98.47     & (96.13, \textbf{22.36})     & -        & (95.43, 18.92)       \\ \hline
				\multirow{3}{*}{T2T}                                                   & \multirow{3}{*}{T2T-14} & CLS                 & 97.29   & (96.36, \textbf{16.89})   & 97.1     & (96.79, 7.12)    & 97.25    & (97.06, 7.52)    & 97.03    & (93.87, 8.55)     & 97.41     & (96.77, 14.40)     & -        & (96.70, \underline{15.14})       \\ \cline{3-15} 
				&                         & VIS                 & 97.26   & (97.51, \underline{9.12})    & 97.13    & (96.72, 8.66)    & 96.95    & (97.41, 6.81)    & 97.12    & (95.64, 6.8)      & 97.07     & (97.49, \textbf{10.22})     & -        & (97.54, 8.97)        \\ \cline{3-15} 
				&                         & CLS+VIS             & 97.22   & (96.31, \textbf{15.65})   & 97.11    & (96.30, 8.04)    & 97.14    & (96.11, 8.75)    & 97       & (95.67, 9.33)     & 97.46     & (96.38, 10.94)     & -        & (96.13, \underline{13.99})       \\ \hline
				\multirow{3}{*}{TNT}                                                   & \multirow{3}{*}{TNT-S}  & CLS                 & 96.96   & (74.84, \underline{31.94})   & 96.53    & (70.34, 22.76)   & 97.02    & (70.99, 22.13)   & 96.51    & (71.35, 28.44)    & 96.93     & (75.68, \textbf{32.39})     & -        & (74.43, 31.03)       \\ \cline{3-15} 
				&                         & VIS                 & 96.7    & (75.71, \textbf{32.6})    & 96.41    & (61.21, 19.12)   & 96.59    & (61.32, 19.39)   & 96.52    & (66.23, 22.19)    & 96.65     & (74.26, 30.85)     & -        & (75.91, \underline{31.83})       \\ \cline{3-15} 
				&                         & CLS+VIS             & 96.99   & (73.04, 28.42)   & 96.91    & (68.29, 22.87)   & 96.81    & (66.44, 20.66)   & 97.1     & (65.460, 21.5)    & 96.77     & (76.36, \textbf{32.34})     & -        & (75.70. \underline{32.16})       \\ \hline
			\end{tabular}%
		}
		\vspace{-3mm}
	\end{table*}

	\begin{table*}[!htb]
		\vspace{-5mm}
		\centering
		\caption{Models accuracy under \ac{nt} and models (accuracy, robust accuracy) under \ac{at} in \ac{fl} with \ac{niid}(2). The \textbf{first} and the \underline{second} best robust accuracy are marked.}
		\label{tab:niid2}
		\resizebox{\textwidth}{!}{%
			\begin{tabular}{|c|c|c|c|c|c|c|c|c|c|c|c|c|c|c|}
				\hline
				\multicolumn{3}{|c|}{Architecture}                                             & \multicolumn{2}{c|}{FedAvg} & \multicolumn{2}{c|}{FedProx} & \multicolumn{2}{c|}{FedGate} & \multicolumn{2}{c|}{SCAFFOLD} & \multicolumn{2}{c|}{FedWAvg(1)} & \multicolumn{2}{c|}{FedWAvg(10)} \\ \hline
				
				Token                        & Model                   & Class. Head & NT      & AT               & NT       & AT               & NT       & AT               & NT       & AT                & NT        & AT                 & NT      & AT                    \\ \hline
				
				\multirow{6}{*}{\begin{tabular}[c]{@{}l@{}}Image\\ patch\end{tabular}} & \multirow{3}{*}{VIT-S}  & CLS                 & 83.52   & (58.51, \underline{10.83})   & 79.86    & (46.70, 6.89)    & 82.92    & (55.41, 7.62)    & 77.88     & (53.28, 7.78)    & 84.62     & (58.44, 10.52)     & -        & (58.20, \textbf{14.01})       \\ \cline{3-15} 
				&                         & VIS                 & 82.56   & (61.31, \textbf{13.36})   & 73.46    & (51.02, 7.51)    & 75.61    & (48.71, 7.34)    & 73.44     & (50.12, 7.65)    & 80.49     &  (57.63, 11.45)   & -        & (58.28, \underline{13.3})        \\ \cline{3-15} 
				&                         & CLS+VIS             & 85.03   & (58.63, \underline{10.19})   & 80.89    & (46.11, 8.76)    & 80.64    & (47.17, 9.33)    & 78.91     & (51.4, 6.10)     & 86.49     & (59.36, 9.76)      & -        & (54.75, \textbf{11.1})        \\ \cline{2-15} 
				& \multirow{3}{*}{VIT-B}  & CLS                 & 96.43   & (87.81, 4.96)    & 94.88    & (82.04, 1.2)     & 94.51    & (90.64, 3.15)    & 94.89     & (90.17, 6.68)    & 96.7      & (86.56, \underline{8.10})      & -        & (83.62, \textbf{13.79})       \\ \cline{3-15} 
				&                         & VIS                 & 96.72   & (83.76, 5.91)    & 96.24    &  (80.69, 3.04)                & 95.23    & (89.18, 2.45)    & 94.86     & (88.99, 3.85)    & 97.15     &  (90.49, \underline{8.95})                  & -        &  (90.61, \textbf{12.22})                    \\ \cline{3-15} 
				&                         & CLS+VIS             & 96.42   & (83.95, 7.26)    & 95.7     & (82.62, 2.48)    & 95.38    & (91.39, 4.81)    & 93.23     & (88.15, 5.69)    & 96.48     & (86.90, \underline{7.61})      & -        & (84.68, \textbf{14.09})       \\ \hline
				\multirow{3}{*}{T2T}                                                   & \multirow{3}{*}{T2T-14} & CLS                 & 92.1    & (85.6, 1.88)     & 89.45    & (83.10, 4.83)    & 87       & (86.65, 5.93)    & 86.87     & (83.02, 3.90)    & 92.12     & (83.66, \underline{7.08})      & -        & (80.77, \textbf{8.75})        \\ \cline{3-15} 
				&                         & VIS                 & 81.8    & (84.75, 3.59)    & 81.45    & (80.84, 2.84)    & 72.1     & (80.26, 5.04)    & 79.2      & (79.19, 2.99)    & 84.84     & (82.31, \underline{6.87})      & -        & (87.66, \textbf{9.33})        \\ \cline{3-15} 
				&                         & CLS+VIS             & 90.81   & (83.47, 5.28)    & 83.77    & (79.43, 1.40)    & 88.5     & (81.79, 4.52)    & 88.9      & (80.42, 6.07)    & 85.47     & (78.32, \underline{6.28})      & -        & (79.02, \textbf{7.32})        \\ \hline
				\multirow{3}{*}{TNT}                                                   & \multirow{3}{*}{TNT-S}  & CLS                 & 71.31   & (57.69, \underline{11.89})   & 72.5     & (45.01, 9.25)    & 75.81    & (58.01, 8.68)    & 68.43     & (51.78, 8.00)    & 84.78     & (50.94, 11.43)     & -        & (52.51, \textbf{12.96})       \\ \cline{3-15} 
				&                         & VIS                 & 64.11   & (59.23, \textbf{12.08})   & 66.04    & (44.73, 5.8)     & 63.39    & (50.64, 6.3)     & 59.42     & (47.75, 5.9)     & 60.45     & (54.88, \underline{9.97})      & -        & (51.45, \underline{9.97})        \\ \cline{3-15} 
				&                         & CLS+VIS             & 80.3    & (50.09, 8.56)    & 68.81    & (42.13, 5.96)    & 68.19    & (51.77, 9.59)    & 64.67     & (46.09, 8.07)    & 78.77     & (50.98, \textbf{9.94})      & -        & (53.66, \underline{9.31})        \\ \hline               
			\end{tabular}%
		}
		\vspace{-3mm}
	\end{table*}

	\begin{table*}[t]
		\centering
		\caption{Per-round communication and client memory cost of the tested models
			under FAT at fp32 precision. Client memory counts parameters, gradients and SGD
			momentum, excluding activations. Totals assume $K=5$ clients and $T=50$ rounds,
			bidirectional.}
		\label{tab:commcost}
		\begin{tabular}{lcccc}
			\hline
			Model & Parameters & Uplink per client per round & Client memory & Total run \\
			\hline
			ViT-S/16    & $\approx$22.0\,M & 84\,MiB  & 252\,MiB & 41\,GiB  \\
			T2T-ViT-14  & 21.5\,M          & 82\,MiB  & 246\,MiB & 40\,GiB  \\
			TNT-S       & 23.8\,M          & 91\,MiB  & 272\,MiB & 44\,GiB  \\
			ViT-B/16    & 86.5\,M          & 330\,MiB & 990\,MiB & 161\,GiB \\
			\hline
		\end{tabular}
		\vspace{-3mm}
	\end{table*}
	\textbf{FedWAvg convergence.} \cref{fig:conv_tnts_cls} shows the accuracy and robust accuracy of \ac{tnt}-CLS model in the \ac{fl} and \ac{fat} processes, respectively, with different aggregation methods. Similar figures with loss values for other models are in \cref{app:more_convergence}. In federated natural training with \ac{iid} and \ac{niid}(4) data distributions as shown in \cref{fig:conv_tnts_cls_iid_nt} and \cref{fig:conv_tnts_cls_niid4_nt}, we notice that all the aggregation methods yield a fast convergence roughly from round 2. Unlike other methods, FedProx requires more rounds to achieve comparable accuracy. FedWAvg exhibits convergence behavior comparable to that of other methods. While with \ac{niid}(2) data distribution, as shown in \cref{fig:conv_tnts_cls_niid2_nt}, FedWAvg shows comparable convergence, better for some models, than FedAvg. FedProx exhibits slow convergence, whereas FedGate and SCAFFOLD exhibit unstable convergence. 
	
	In \ac{fat} with \ac{iid} data distribution as shown in \cref{fig:conv_tnts_cls_iid_at}, FedWAvg shows comparable convergence to other methods while FedProx shows slow convergence. With \ac{niid}(4) and \ac{niid}(2) data distributions, as shown in \cref{fig:conv_tnts_cls_niid4_at} and \cref{fig:conv_tnts_cls_niid2_at}, FedGate and SCAFFOLD show unstable convergence for the accuracy and show slow convergence for the robust accuracy. Moreover, FedProx exhibits slow convergence in robust accuracy, whereas FedWAvg shows comparable or better convergence. 
	
	These findings are consistent with Lemma~\ref{lem:dev} and Lemma~\ref{lem:graddev}. The FedWAvg weights approach $1/K$ after about ten rounds in each partition, see Figs.~\ref{fig:wieghts_tnts_cls}, and \ref{fig:wieghts_vits_cls_iid}-\ref{fig:wieghts_tnts_cls_vis_niid_4}. As a result, the cosine spread $\Delta_c$ approaches zero, so by Lemma~\ref{lem:dev} the deviation between the FedWAvg and FedAvg aggregates vanishes, and by Lemma~\ref{lem:graddev} so does the gap between the corresponding gradients. The two methods therefore differ during the initial transient phase, where client drift is at its peak, which is what the observed curves show. We note that this is a statement about the aggregates converging to one another, not a convergence guarantee for either method; see Sec.~\ref{sec:openproblems}.
	
	Moreover, we calculate the model drift using \ac{svcca} \cite{raghu@2017@svcca}, as shown in \cref{fig:drift_tnts_cls_iid4_at}, between client 1 and client 4 (top), between client 1 and the server (middle), and between client 4 and the server (bottom) for layer 1 and layer 9 of \ac{tnt}-CLS model. In \cref{fig:drft_tnts_cls_iid4_clean_at}, the \ac{svcca} is calculated using clean samples, while in \cref{fig:drift_tnts_cls_niid4_adv_at} the \ac{svcca} is calculated using the adversarial samples. Similar figures for other models are given in \cref{app:more_model_drift}. We found that at layer 1, drift between clients and the server decreases during training, and all methods exhibit comparable behavior. At layer 9, the drift behavior of FedGate and SCAFFOLD is unstable, unlike that of FedAvg and FedWAvg. 
	
	The localization of the persistent drift in the later layers aligns with our expectations from Sec.~\ref{sec:whylast}: at layer~1, representations converge across all methods, but it is in the classifier-adjacent layers that label skew and adversarial training jointly cause distortions in the local models. Furthermore, Theorem~\ref{thm:contract} predicts that FedWAvg will reduce the drift in these layers.
	
	\textbf{Accuracy and robust accuracy with FedWAvg.} For \acp{cnn}, the aggregation methods for \ac{fl} are mainly developed and tested in the natural federated learning process. In this work, we test the \stateart aggregation methods in the \ac{fat} process for transformers. \cref{tab:iid}, \cref{tab:niid4}, and \cref{tab:niid2} show the best accuracy for the natural federated learning and the best robust accuracy and the corresponding accuracy for the \ac{fat} with \ac{iid}, \ac{niid}(4), and \ac{niid}(2) data distributions. 
	
	In federated natural training with \ac{iid} and \ac{niid}(4) data, all aggregation methods, including FedWAvg, yield comparable model accuracy, while with \ac{niid}(2), FedProx, FedGate, and SCAFFOLD achieve lower model accuracy compared to FedAvg. On the other hand, FedWAvg achieves better accuracy than FedAvg for models like \ac{t2t}-VIS and \ac{tnt}-CLS, and achieves comparable accuracy for models like \ac{t2t}-CLS and \ac{vit}-B, and achieves less accuracy for models like \ac{t2t}-CLS+VIS. 
	
	For the \ac{fat} process, we evaluate the performance of FedWAvg using the \ac{niid} data partitioning. With \ac{niid}(4), we notice that FedWAvg always yields better robust accuracy than FedProx, FedGate, and SCAFFOLD. Compared to FedAvg, we notice that FedWAvg yields better robust accuracy, like in \ac{tnt}-CLS-VIS and \ac{vit}-B, comparable robust accuracy, like in \ac{vit}-CLS, and lower robust accuracy than FedAvg, like in \ac{t2t}-CLS. While with \ac{niid}(2), FedWAvg yields better robust accuracy than other aggregation methods with all models except with FedAvg in \ac{tnt}-VIS model. 
	
	The behavior of FedWAvg($1$) versus FedWAvg($10$) illustrates the trade-off described in Remark~\ref{rem:q}: increasing the scale factor benefits more when client drift is highly dispersed, but the cost grows exponentially with $e^{4q}$, regardless of the partition. Consequently, FedWAvg($10$) is likely most effective in highly skewed partitions and does not provide a consistent advantage at moderate skew levels, as demonstrated in \cref{tab:niid4,tab:niid2}

	\textbf{Robustness and tokenization.}
	Tokenization played an important role in the \ac{fl} and the \ac{fat} processes. As shown in \cref{tab:iid}, \cref{tab:niid4}, and \cref{tab:niid2}, the tokens-to-token method has a positive role in enhancing the models' accuracies in \ac{nt} and \ac{at} processes when compared with size-comparable vision transformer \ac{vit}-S and \ac{tnt}-S, while it has a negative role in enhancing models' robust accuracies. The tokens-to-token method demonstrated its ability to represent training samples with local and global features, thereby enhancing model accuracy. The main reason for the low robust accuracy, as mentioned in \cite{aldahdooh@2025@beyond}, is that the energy spectrum of the perturbation that is generated using \ac{pgd} for the \ac{t2t} model is not spread across all frequencies, which makes \ac{t2t} not robust against the \ac{pgd} attack. On the other hand, image patches in \ac{vit}-S and the mapping of local pixel dependencies in \ac{tnt} models don't help with \ac{niid} data partitioning. With \ac{niid}(4), \ac{vit}-S and \ac{tnt} show comparable performance that decreases the model's accuracy and increases the model's robust accuracy. While with \ac{niid}(2), \ac{vit} shows better model accuracy in \ac{nt} and in \ac{at} process. In conclusion, you can select the tokenization method based on the relative priority you assign to accuracy and robustness. 
	
	\textbf{Robustness and classification head.}
	The role of the classification head type appears during the \ac{nt} process with \ac{niid}(2) data partitioning. As shown in \cref{tab:niid2}, using VIS tokens only for the classification head notably decreases the model's accuracy, whereas using the CLS token only for the classification head significantly improves it. Compared with using the CLS token, combining both configurations (CLS+VIS) may yield higher model accuracy, as in \ac{tnt} with FedAvg, or lower model accuracy, as in \ac{t2t} with FedWAvg.
	
	Moreover, the role of the classification head type becomes apparent during the \ac{at} process, particularly with \ac{iid} and \ac{niid} data partitioning. With the \ac{iid} data distribution, as shown in \cref{tab:iid}, using CLS token for the classification head enhances the model's accuracy and the model's robust accuracy in \ac{tnt} models, while using VIS tokens for the classification head decreases the model's robust accuracy in \ac{t2t} models. With the \ac{niid}, the preference of one classification head type is not clear. Finally, we conclude that using VIS tokens alone for the classification head is not recommended.
	
	\textbf{\ac{vit}-S and \ac{vit}-B}. We notice that the \ac{vit}-B significantly enhances the model accuracy and fails to improve the robust accuracy with \ac{iid} and \ac{niid}(4). While with \ac{niid}, \ac{vit}-B achieves comparable robust accuracy to \ac{vit}-S. Hence, we recommend, for high heterogeneous data partitioning, to use \ac{vit} model with large number of attention blocks.
	
	\begin{table*}[t]
		\centering
		\vspace{-6mm}
		\caption{Cost of the aggregation methods per communication round. $P$ is the
			model size, $m$ the number of participating clients and $d_L$ the last-layer
			dimension ($d_L \ll P$; $d_L=3{,}850$ for a ViT-S CLS head).}
		\label{tab:aggoverhead}
		\begin{tabular}{lccccc}
			\hline
			Method & Uplink & Downlink & Client state & Extra client compute & Server compute \\
			\hline
			FedAvg   & $P$   & $P$   & stateless & ---                & $O(mP)$ \\
			FedProx  & $P$   & $P$   & $+P$      & proximal term      & $O(mP)$ \\
			FedGate  & $P$   & $P$   & $+P$      & gradient tracking  & $O(mP)$ \\
			SCAFFOLD & $2P$  & $2P$  & $+P$      & drift correction   & $O(mP)$ \\
			\textbf{FedWAvg} & $P$ & $P$ & \textbf{stateless} & \textbf{---} & $O(mP)+O(m\,d_L)$ \\
			\hline
		\end{tabular}
		\vspace{-4mm}
	\end{table*}
	
	\vspace{-1.5mm}
	\subsection{Computational and Communication Cost Analysis}
	\label{sec:cost}
	\vspace{-1mm}
	Adversarial training requires significant computation, and vision transformers have large parameter counts. Therefore, evaluating the practicality of the FAT pipeline warrants a dedicated discussion. This section identifies three key costs: communication, client computation, and client memory, and compares the expense of FedWAvg to other options.
	
	\textbf{Communication.} Each client per round transmits $4P$ bytes, with $P$ parameters sent at fp32 precision, regardless of whether training is natural or adversarial. Table~
	\ref{tab:commcost} shows the communication budget for the four backbones. The total data transferred over a full run amounts to $2KT\cdot 4P$ bytes, since the global model is broadcast each round.
	
	\textbf{Client computation.} For each minibatch, the $s$-step PGD attack targets $n_{adv}=rB$ samples, adding $s$ extra forward-backward passes on a fraction $r$ of the batch. Afterward, a training step is performed on the entire batch. The approximate per-batch computational cost compared to natural training is therefore approximately \vspace{-1.5mm}
	
	\begin{equation}
		\vspace{-1.5mm}
		1 + r\,s \;=\; 1 + 0.5\times 7 \;=\; 4.5 ,
		\label{eq:atcost}
	\end{equation}
	for the adversarial ratio $r=0.5$ and $s=7$ PGD steps used here. Communication remains the same. The FAT bottleneck is thus client computation, not bandwidth, and this influences the choice of affordable aggregation methods.
	
	\textbf{Client memory.} When using SGD with momentum, each client's memory footprint before activations is roughly $3\times 4P$ bytes, covering parameters, gradients, and momentum. This amounts to 252-272~MiB for the three models with approximately 22 million parameters, and about 990~MiB for ViT-B/16. PGD does not require persistent activation memory because its attack iterations are sequential single-graph passes; however, it increases latency as per Eq.~\eqref{eq:atcost}. At a $224\times224$ input resolution, the activation footprint limits the batch size to 24, as mentioned in Sec.~\ref{sec:setup}. Consequently, the three smaller models can be trained on a mid-range accelerator using FAT, while ViT-B/16 requires a cross-silo setup.
	
	\textbf{Cost of the aggregation methods.} \Cref{tab:aggoverhead} shows a comparison of the five aggregation methods discussed in this paper. SCAFFOLD increases the communication per round in both directions, since its control variates match the size of the model. Both SCAFFOLD and FedGate require clients to hold state and maintain an additional model-sized buffer. FedProx does not add communication but requires clients to keep the global model and evaluate the proximal term at each local step. FedWAvg adds no communication, computation, or client state: the last-layer parameters examined by Eq.~\eqref{eq:weights_cos} are already part of the updates sent to the server. Its similarity calculation on the server is $O(m\,d_L)$, where $d_L$ is the last-layer dimension. For a ViT-S CLS head, $d_L = 384\times 10 + 10 = 3{,}850$, meaning that at $K=5$, FedWAvg's total cost is on the order of $10^4$ floating-point operations per round, which is roughly $10^{-9}$ of a client's local epoch. The cost increases linearly with the number of participating clients.
	
	This reframes the empirical result from Sec.~\ref{sec:results} as a cost-benefit analysis rather than just a negative outcome. In scenarios where local computation is increased by a factor of $4.5$ due to inner maximization, and clients already store an extra model-sized optimizer state, applying communication- and state-heavy drift correction is ineffective: SCAFFOLD and FedGate use twice as much bandwidth and an additional client buffer, yet achieve still worse robust accuracy than the standard FedAvg, while FedWAvg improves robustness at no significant additional cost.
	
	Three mitigation strategies can be implemented with FedWAvg without any modifications and are important to consider for deployment. Partial participation ($C<1$) linearly decreases the per-round cost and keeps Algorithm~\ref{alg:algo_fedwavg} unchanged. Using mixed-precision or quantized uplink halves the communication cost shown in Tab.~\ref{tab:commcost}; the FedTorch library \cite{haddadpour@2021@compression} already supports compressed updates. Lastly, parameter-efficient FAT, which trains and transmits only the classification head and normalization layers, can reduce $P$ significantly; FedWAvg is especially suitable here, as it already only reads the last layer.
	
	\vspace{-3mm}
	\section{Scope and Limitations}
	\label{sec:scope}
	\vspace{-1mm}
	This paper's results are based on federated adversarial training of vision transformers on CIFAR-10, using IID, Non-IID($4$), and Non-IID($2$) partitions, with 5 fully participating cross-silo clients and a PGD-7 adversary at $\varepsilon = 8/255$. We clarify what is supported and what isn't, along with our planned future extensions. Each configuration involves a 50-round federated run with adversarial training on all clients. The reported study includes 12 models from Sec.~\ref{sec:vitmodels} tested with five to six aggregation methods and three partitions in both natural and adversarial settings. A single adversarial run takes roughly 6 hours \emph{on a single} NVIDIA RTX~A6000 which influences our prioritization of future work.
	
	\textbf{Dataset coverage.} CIFAR-10 serves as the main benchmark in the FAT literature 
	\cite{shah@2021@adversarial}, \cite{chen@2021@certifiably}, \cite{hong@2023@propagation}, \cite{chen@2022@calfat}, \cite{zhu@2023@combating}, \cite{zhang@2023@delving}. This facilitates the interpretation of aggregation comparisons in Sec.~\ref{sec:results} relative to prior studies. The resizing to $224\times224$ ensures that transformers are assessed at their native token resolution, Sec.~\ref{sec:setup}. Although CIFAR-10 has a limited number of labels, the impact of label skew within a partition depends on the number of classes. Our first extension aims to include CIFAR-100 and Tiny-ImageNet, using consistent partitioning, aggregation methods, and attack budgets across the four backbones with the CLS head. Notably, \cite{chen@2022@calfat} observed similar qualitative degradation patterns on CIFAR-100 and an ImageNet subset, suggesting that the phenomenon persists at larger scales but does not replace the need for dedicated experiments. Remark~\ref{rem:scope}, Theorem~\ref{thm:contract} applies regardless of dataset, and the benefits of FedWAvg are expected to increase with the dispersion of client head drift, implying scale invariance even with a larger label space.
	
	\noindent\textbf{Client scale.} The experiments are conducted with $K=5$ and full participation, as detailed in \cite{qu@2022@rethinking}. In real-world federations, the number of clients can be much larger. While our cost analysis in Sec.~\ref{sec:cost} shows that FAT with transformers of this size is more of a cross-silo workload rather than a cross-device one, the cross-silo setup can include roughly 100 clients. We also plan to explore larger values of $K$, specifically $50$ and $100$, using Dirichlet($\beta$) partitioning with $C<1$ for a subset of four models and three aggregation methods. Although the theory is formulated for any value of $K$, Lemma~\ref{lem:bounded} ensures the weights stay within bounds at any scale, and Theorem~\ref{thm:contract} indicates that the improvement over FedAvg depends more on the dispersion of client drift than on the number of clients. Since reducing data per client increases this dispersion, we expect the advantage of FedWAvg to not only persist but possibly grow as $K$ increases. We present this as a prediction rather than a definitive result.

	\noindent\textbf{Convolutional baseline.} This paper does not feature a ResNet-50 baseline trained under the same FAT and partitioning conditions. The reason for exploring transformers in FAT is based on existing studies \cite{qu@2022@rethinking} for the federated benefit of ViTs in heterogeneous environments and \cite{bai@2021@recent} for the observation that transformers are not inherently more adversarially robust than CNNs at comparable capacity, rather than direct comparisons made in this work. We do not claim transformers outperform CNNs in the FAT context. The third extension we plan is to develop a matched ResNet-50 FAT baseline under Non-IID($4$) and Non-IID($2$) conditions.
	
	\noindent\textbf{Threat model.} 
	The main challenge here is a PGD-7 $\ell_\infty$ attacker working within a single budget, tested on the overall test set. Exploring robustness against stronger or adaptive attacks like AutoAttack, across different budgets, or against attacks tailored for transformer architectures is beyond the scope of this work. FedWAvg also tackles heterogeneity rather than Byzantine behavior: as shown in Lemma~\ref{lem:bounded}, it never excludes a client, so it shouldn't be considered a reliable defense against poisoning, see Remark~\ref{rem:scope}.
	
	\noindent\textbf{Theory.} Proposition~\ref{prop:gibbs}, Theorem~\ref{thm:contract} and Lemmas~\ref{lem:bounded}-\ref{lem:graddev} are exact and self-contained. Proposition~\ref{prop:transfer} is conditional in that it transfers a stationarity guarantee for the round-weighted objective to the uniform objective, without itself establishing that guarantee. Theorem~\ref{thm:gen} evaluates the weighting in terms of any admissible uniform-convergence bound for the adversarial loss class. We do not claim a convergence theorem for FedWAvg with respect to the min-max objective since the aggregation weights are adaptive and time-varying and therefore a fixed-weight argument cannot be applied to them; Section~\ref{sec:openproblems} clearly specifies what additional requirements such a theorem would need. Moreover, Theorem~\ref{thm:contract} provides a bound on the directional drift of the classification head, not on the robust accuracy directly. It is still the main open theoretical question to establish a direct link between the contraction of head drift and gains in robust accuracy.

	\vspace{-2mm}
	\section{Conclusion}
	\label{sec:conclusion}
	
	This study examined whether adversarial training is feasible within a federated learning framework for vision transformers. We tested 12 models across three tokenization methods and three classification-head configurations on CIFAR-10, in both IID and two label-skewed partitions. Our findings are threefold. First, aggregation methods designed to reduce client drift in Non-IID scenarios tend to decrease the model's robust accuracy compared to FedAvg when using a transformer backbone. The cost analysis in Sec.~\ref{sec:cost} shows these methods require up to twice the communication per round and add an extra model-sized client state, making FedAvg the more effective and cost-efficient baseline here. Second, the optimal tokenization approach depends on whether the goal is high clean or robust accuracy, tokens-to-token embedding favors the former but is vulnerable to PGD attacks. In addition, classification heads that rely solely on visual tokens should be avoided in FAT pipelines. Third, we introduced FedWAvg, an extension of FedAvg that calculates weights based on the cosine similarity between the global model's final layer and each client's last layer. This method enhances robust accuracy on highly heterogeneous data and demonstrates convergence and drift behaviors similar to FedAvg.
	
	The analysis in Sec.~\ref{sec:theory} clarifies why these two properties coexist. FedWAvg is essentially a Gibbs reweighting of clients based on the squared distance between their direction-normalized last layers. As a result, it never increases the overall directional drift of the classification head compared to FedAvg, with any improvement coming from the accumulated variance of client drifts. This variance diminishes under IID data and decreases as training progresses. Therefore, FedWAvg converges to FedAvg in regimes where FedAvg suffices and diverges in early, highly heterogeneous rounds where drift is largest. The same analysis also highlights the scale factor's role as a trade-off: balancing drift reduction against estimation variance. This explains why a larger scale factor performs better under extreme skew. Sec.~\ref{sec:scope} discusses the study's limitations and suggests future extensions, with a larger label space dataset being a priority.

	\vspace{-2mm}
	\section*{Funding Support}
	The project is funded by both R\'egion Bretagne (Brittany region), France, and direction g\'en\'erale de l'armement (DGA).
	\vspace{-2mm}
	\section*{Ethical Statement} This study does not contain any studies with human or animal subjects performed by any of the authors.
	\vspace{-2mm}
	\section*{Conflicts of Interest} The authors declare that they have no conflicts of interest to this work.
	\vspace{-2mm}
	\section*{Data Availability Statement} The data that support the findings of this study are available upon request.
	\vspace{-2mm}
	\section*{Author Contribution Statement}
	\textbf{Ahmed Aldahdooh:} Conceptualization, Methodology, Software, Validation, Formal analysis, Investigation, Resources, Data curation, Writing - original draft, Writing - review $\&$ editing, Visualization. \textbf{Wassim Hamidouche:} Conceptualization, Supervision, Project administration. \textbf{Olivier Déforges:} Supervision, Project administration, Funding acquisition.

	\vspace{-3mm}
	\appendix
	\section{Vision Transformers} 
	\label{app:vits}
	In this section, we briefly review related work on transformers and then present the models we consider in our experiments, as shown in \cref{fig:fat_models}. 
	\vspace{-3mm}
	\subsection{Related work}
	\label{app:vits_a}
	Transformer \cite{vaswani@2017@attention} was first introduced for \ac{nlp} tasks. It adopts the self-attention mechanism to learn the model. Transformers and their variants maintain \stateart performance across different \ac{nlp} tasks. Recently, transformers have proven effective for various computer vision tasks, including image processing \cite{chen@2021@pretrained}, video processing \cite{liang@2024@vrt}, and image classification \cite{dosovitskiy@2021@image}. Dosovitskiy \etal \cite{dosovitskiy@2021@image} was the first to build an image classification model, \acf{vit}, that uses the vanilla transformer encoder blocks. \ac{vit} and its variants establish the \stateart performance, especially if it is trained with significantly large-scale datasets, such as JFT-300M \cite{dosovitskiy@2021@image}. Using transfer learning, \ac{vit} models can be downgraded to smaller datasets, such as ImageNet-1k, and achieve performance comparable to or better than \stateart \acp{cnn} models. Recently, in \cite{qu@2022@rethinking}, Qu \etal investigated the feasibility of using \acp{vit} in the \ac{fl} settings and observed that \acp{vit} significantly accelerate convergence and reach a better global model, especially when dealing with heterogeneous data. 
	
	\ac{vit} has the capability to learn the global context of the input image, while it is not effectively capable to learn local features as \acp{cnn}, and that's why \ac{vit} requires a large-scale dataset for training to maintain the \stateart performance. Hence, many efforts have been made to mitigate this obstacle \cite{dosovitskiy@2021@image}, \cite{yuan@2021@tokens}, \cite{han@2021@transformer}, \cite{xie@2021@sovit}. One approach is the embedding-based approach, in which the method for generating embedded patches to be passed to the transformer encoder is modified. In \cite{dosovitskiy@2021@image}, \ac{vit}-Res is introduced, which replaces the input image patches with the flattened ResNet-50 feature maps to generate the embedded patches. While in \cite{yuan@2021@tokens}, \acf{t2t} replaces input image patches with a tokens-to-token (T2T) transformation. The T2T module progressively structures the image into tokens by recursively aggregating neighboring tokens into one token. The features that are learned by the T2T module are then passed to the transformer encoder. In \cite{han@2021@transformer}, Han \etal. suggested modeling both patch-level and pixel-level representations and proposed the \acf{tnt} architecture. It stacks multiple \ac{tnt} blocks that each have an inner transformer and an outer transformer. The inner transformer block further divides the image patch into sub-patches to extract local features from pixel embeddings. The output of the inner transformer block is concatenated with the patch embeddings to form the input to the outer transformer block.  Other architectures exist in the literature, like \ac{cvt} \cite{wu@2021@cvt} and \ac{cpvt} \cite{chu@2021@positional}. Another approach to mitigate the \ac{vit} obstacle is to customize the model's classification head. \ac{vit} only uses the class token for the classification while \ac{sovit} \cite{xie@2021@sovit} uses visual tokens along with the class token. \ac{sovit} proposes a second-order cross-covariance pooling of visual tokens, which is combined with the class token for final classification. Moreover, the work in \cite{mao@2021@towards} considered only the visual tokens.
	\vspace{-3mm}
	\subsection{Vision Transformer Models}
	\label{app:vits_method}
	\ac{vit}'s encoder receives as input a 1D sequence of token embeddings. The image $\mathbf{x}\in\mathbb{R}^{H\times W\times C}$ is reshaped into a sequence of flattened 2D image patches $\mathbf{x}_p \in \mathbb{R}^{N\times(P^2\times C)}$, where $H,W,C,P, \text{ and }N$ are image height, image width, number of image channels, patch width and height, and $N=\frac{HW}{P^2}$ is the number of patches, respectively. To prepare the patch embeddings, \cref{eq:vit1}, the flattened patches are mapped to $D$ dimensions via a trainable linear projection, since the transformer encoder uses a constant latent vector size $D$ across all layers. To maintain the positional information, \cref{eq:vit1}, position embeddings are added to the patch embeddings using the standard learnable 1D position embeddings.  The basic component of transformer-based \acp{nn} is the attention block. The standard transformer encoder in \ac{vit} stacks $L$ layers of attention blocks. The attention block consists of two sub-layers: the first is the \ac{msa}, \cref{eq:vit2}, and the second is a simple \ac{mlp} layer, \cref{eq:vit3}, also called a position-wise fully connected feed-forward network (FFN). Layernorm (LN) is applied before every sub-layer, and residual connections are applied after each sub-layer. \Ac{mlp} consists of two linear transformation layers and a nonlinear activation function, \ac{gelu} \cite{dosovitskiy@2021@image}, in between. For classification head, \ac{vit} adds [\texttt{CLS}] token to the sequence of embedded patches ($\mathbf{z}_0^0=\mathbf{x}_{class}$). The [\texttt{CLS}] token in the transformer's output $\mathbf{z_L^0}$ will be used for image representation, as shown in \cref{eq:vit4}.
	
	\begin{figure*}[!htb]
		\centering
		\vspace{-8mm}
		\caption{Tested Vision Transformers’ Architectures for Federated Adversarial Training. The left part shows the main blocks of the transformer. 1) The embedded patches block. 2) the Attention block, and 3) the classification head block. For the embedding patches, we consider three methods, in green blocks: a) image patches, tokens-to-token, and sub-patches for image patches.  For the Attention block, we consider the multi-head attention mechanism. For the classification head block, we consider using the CLS token, VIS tokens, and both. \vspace{-3mm}}
		
		\includegraphics[width=0.77\textwidth, keepaspectratio]{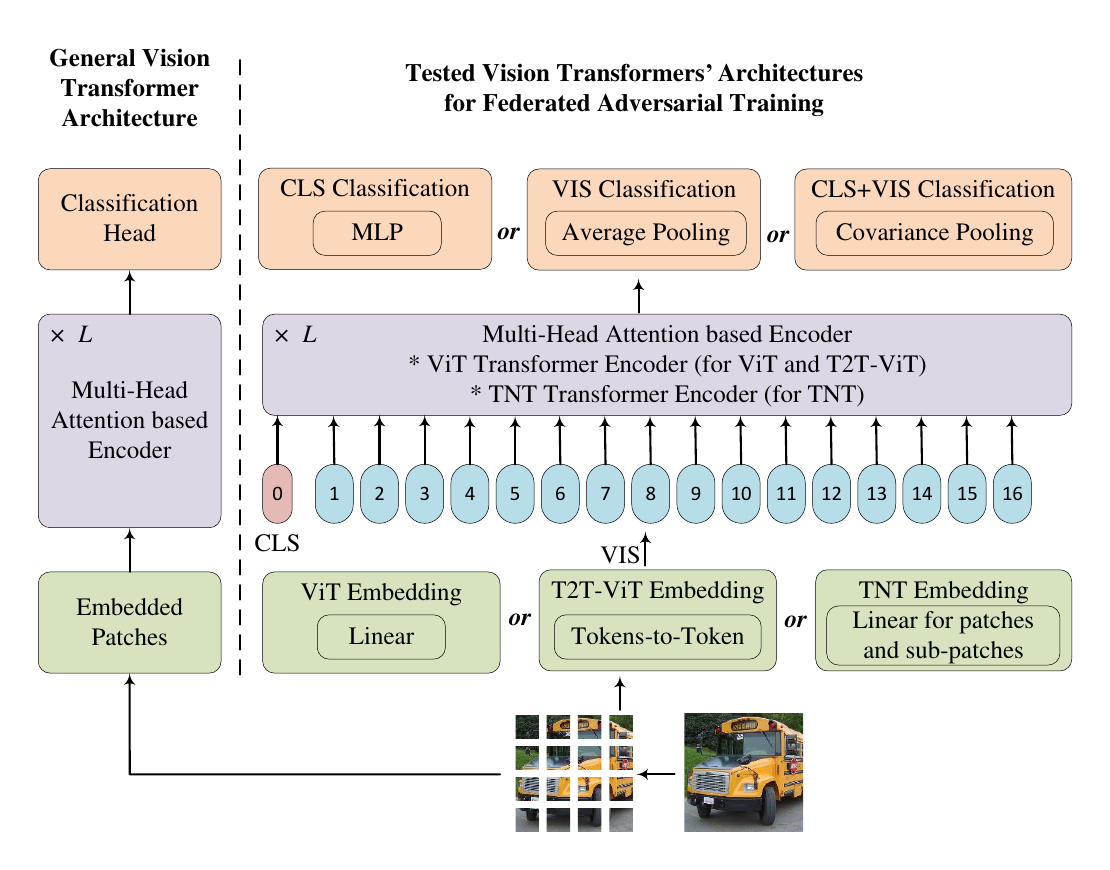}
		
		\vspace{-3mm}
		\label{fig:fat_models}
	\end{figure*}
	
	\begin{equation}
		\begin{gathered}
			\label{eq:vit1}
			\mathbf{z}_0 = [\mathbf{x}_{class};\mathbf{x}_p^1\mathbf{E};\mathbf{x}_p^2\mathbf{E};\dots;\mathbf{x}_p^N\mathbf{E}]+\mathbf{E}_{pos},\\
			\mathbf{E}\in\mathbb{R}^{(P^2\times C)\times D}, \text{  } \mathbf{E}_{pos}\in\mathbb{R}^{(N+1)\times D}
		\end{gathered}
	\end{equation}
	\begin{equation}
		\begin{gathered}
			\label{eq:vit2}
			\mathbf{z}_{l}^{\prime} = MSA(LN(\mathbf{z}_{l-1}))+\mathbf{z}_{l-1}, \text{  } l=1\dots L
		\end{gathered}
	\end{equation}
	\begin{equation}
		\begin{gathered}
			\label{eq:vit3}
			\mathbf{z}_{l} = MLP(LN(\mathbf{z}_{l}^{\prime}))+\mathbf{z}_{l}^{\prime}, \text{  } l=1\dots L
		\end{gathered}
	\end{equation}
	\begin{equation}
		\begin{gathered}
			\label{eq:vit4}
			\mathbf{y} = LN(\mathbf{z}_L^0)
		\end{gathered}
	\end{equation}

	\textbf{Other patch embeddings methods.} As discussed earlier in \cref{app:vits_a}, \ac{t2t} replaces input image patches with a tokens-to-token (T2T) transformation \cite{yuan@2021@tokens}. The T2T module progressively structures the image into tokens by recursively aggregating neighboring tokens into one token. The features that are learned by the T2T module are then passed to the transformer encoder. While \ac{tnt} \cite{han@2021@transformer} models both patch-level and pixel-level representations. It stacks multiple \ac{tnt} blocks, each with an inner transformer and an outer transformer. The inner transformer block further divides the image patch into sub-patches to extract local features from pixel embeddings. The output of the inner transformer block is concatenated with the patch embeddings to form the input to the outer transformer block.
	\begin{figure}[!htb]
		\centering
		\vspace{-5mm}
		\caption{The classification head that is proposed in \cite{xie@2021@sovit} that uses the second-order cross-covariance pooling of visual tokens. FC: fully connected layer. \vspace{-3mm}}
		\includegraphics[width=0.80\linewidth, keepaspectratio]{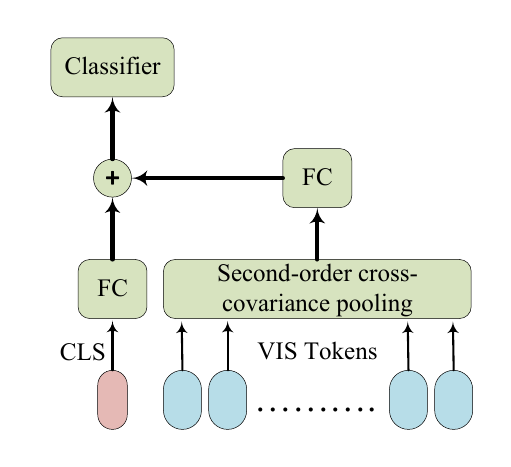}

		\vspace{-3mm}
		\label{fig:sovit_head}
	\end{figure}
	
	\textbf{Other classification head methods.} In order to improve vision transformer robustness, the work \cite{mao@2021@towards} suggested a classification head that depends on the visual tokens [$\mathbf{z}_L^{1\dots N}$], not on the [\texttt{CLS}] token $\mathbf{z}_L^0$. The proposed classification head performs average pooling of the visual tokens. The work in \cite{xie@2021@sovit} proposed a classification head that depends on class and visual tokens, as illustrated in \cref{fig:sovit_head}, to improve network accuracy. 
	
	In summary, we investigate vision transformer models using three embedding methods and three classification head methods. \cref{fig:fat_models} illustrates the models' architectures. For \ac{vit} embedding, we use \ac{vit}-S-16 and \ac{vit}-B-16 models. For \ac{t2t} embedding, we use \ac{t2t}-14, and finally, for \ac{tnt} embedding, we use the \ac{tnt}-S model. For these four models, three classification heads are used: the first uses only the [CLS] token, the second uses only the visual [VIS] tokens, and the third uses both the [CLS] and [VIS] tokens. In total, we tested 12 vision transformer models. 
	
	\vspace{-5mm}
	\section{More figures cosine similarity weights}
	\label{app:more_cosine_sim_w}
\begin{figure*}[!htbp]
		\centering
		\vspace{-8mm}
		\caption{The calculated weights using cosine similarity, \cref{eq:weights_cos} during the federated training process using \ac{vit}-S model and \ac{iid} partitioning. a) for the natural training, and b) for the adversarial training.\vspace{-3mm}}
		\resizebox{0.85\textwidth}{!}{%
			\setlength\tabcolsep{1.5pt}
			\begin{tabular}{c}
				\begin{tabular}{l}\subfloat[]{\includegraphics[width=\textwidth]{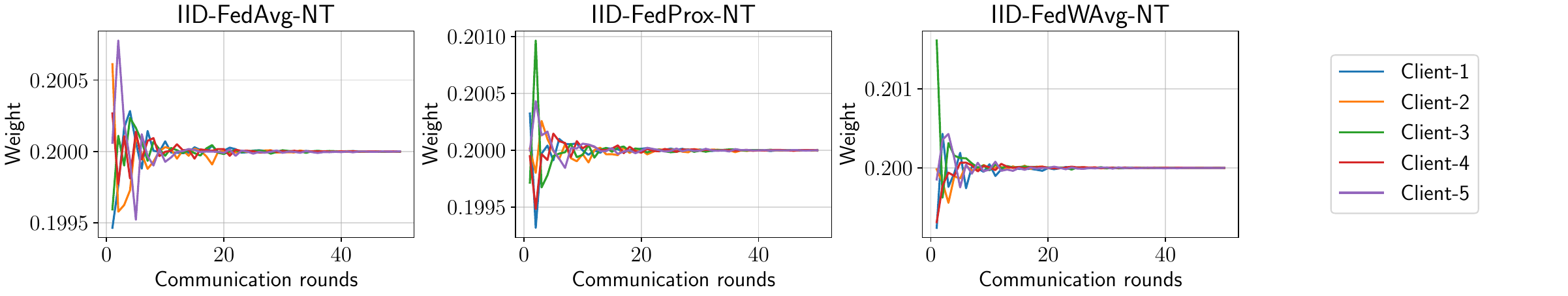}\label{fig:wieghts_vits_cls_iid_b_nt}}\end{tabular}  \\
				\begin{tabular}{l}\subfloat[]{\includegraphics[width=\textwidth]{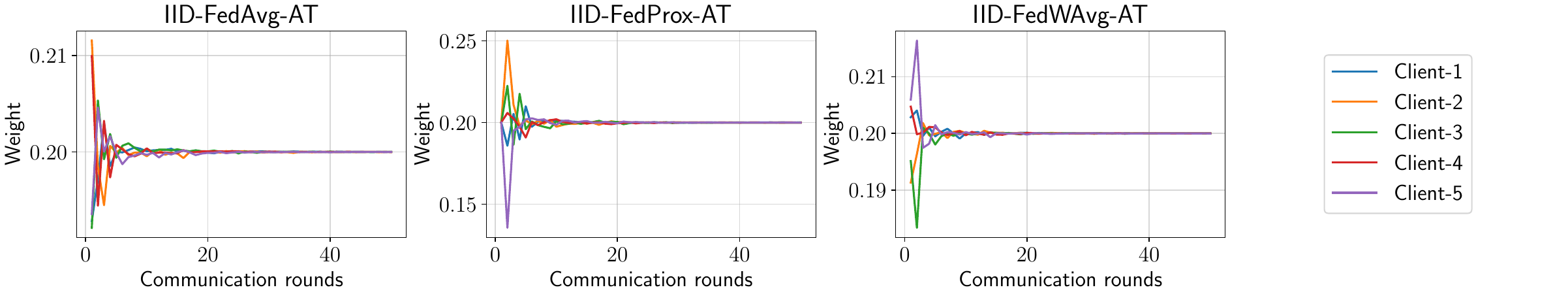}\label{fig:wieghts_vits_cls_iid_b_at}} \end{tabular} \\
			\end{tabular}
		}%

		\label{fig:wieghts_vits_cls_iid}
	\end{figure*}

\begin{figure*}[!htbp]
		\centering
		\vspace{-2mm}
		\caption{The calculated weights using cosine similarity, \cref{eq:weights_cos} during the federated training process using \ac{vit}-S model and \ac{niid} partitioning (each client has 4 classes). a) for the natural training, and b) for the adversarial training.\vspace{-3mm}}
		\resizebox{0.85\textwidth}{!}{%
			\setlength\tabcolsep{1.5pt}
			\begin{tabular}{c}
				\begin{tabular}{l}\subfloat[]{\includegraphics[width=\textwidth]{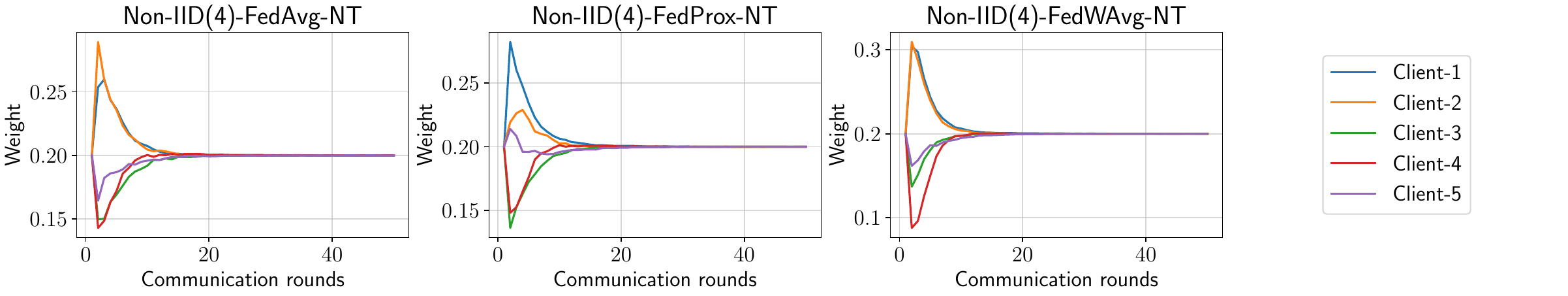}\label{fig:wieghts_vits_cls_niid4_b_nt}}\end{tabular}  \\
				\begin{tabular}{l}\subfloat[]{\includegraphics[width=\textwidth]{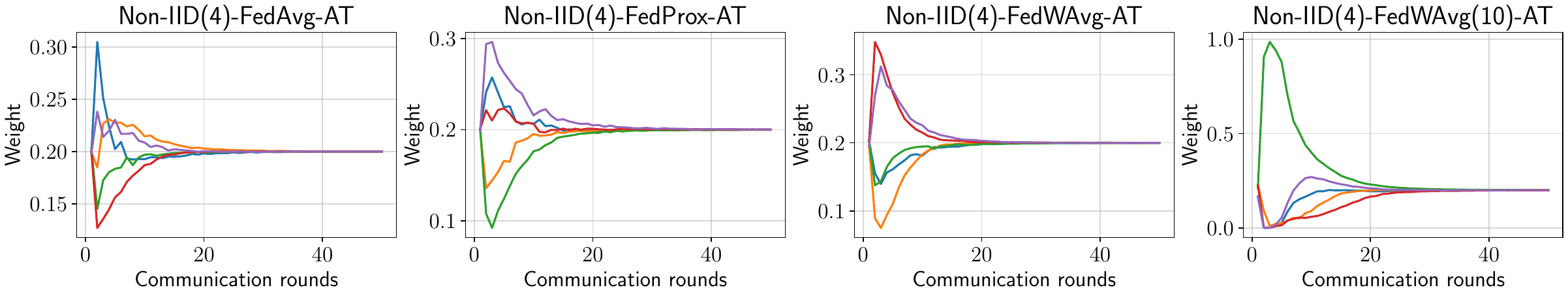}\label{fig:wieghts_vits_cls_niid4_b_at}} \end{tabular} \\
			\end{tabular}
		}%

		\label{fig:wieghts_vits_cls_niid4}
	\end{figure*}

\begin{figure*}[!htbp]
		\centering
		\vspace{-3mm}
		\caption{The calculated weights using cosine similarity, \cref{eq:weights_cos} during the federated training process using \ac{vit}-S model and \ac{niid} partitioning (each client has 2 classes). a) for the natural training, and b) for the adversarial training.\vspace{-3mm}}
		\resizebox{0.85\textwidth}{!}{%
			\setlength\tabcolsep{1.5pt}
			\begin{tabular}{c}
				\begin{tabular}{l}\subfloat[]{\includegraphics[width=\textwidth]{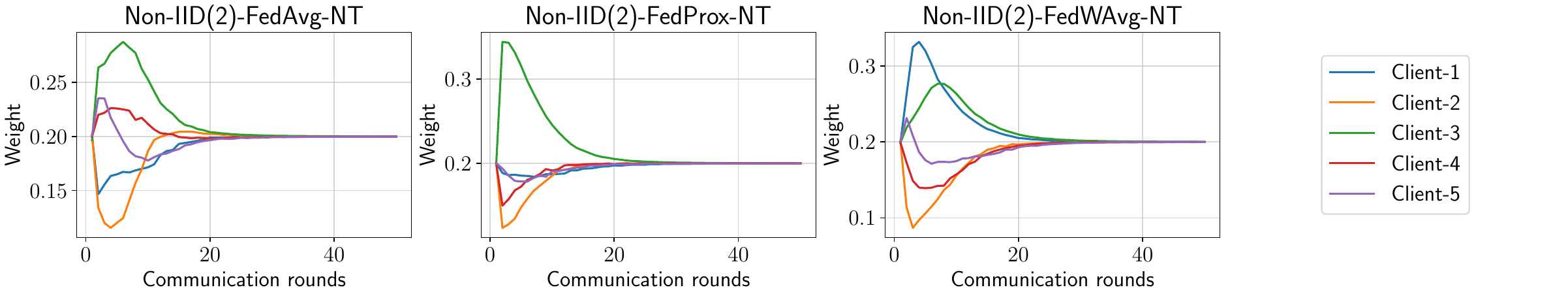}\label{fig:wieghts_vits_cls_niid2_b_nt}}\end{tabular}  \\
				\begin{tabular}{l}\subfloat[]{\includegraphics[width=\textwidth]{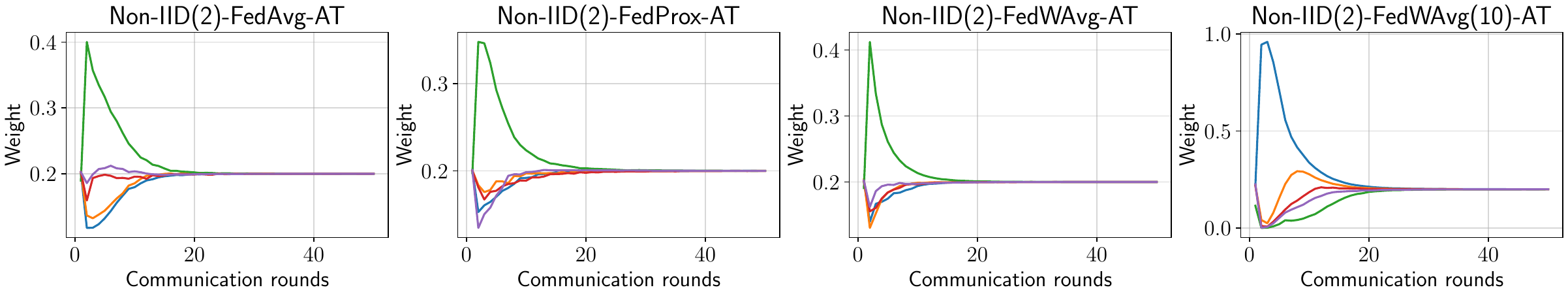}\label{fig:wieghts_vits_cls_niid2_b_at}} \end{tabular} \\
			\end{tabular}
		}%

		\label{fig:wieghts_vits_cls_niid2}
	\end{figure*}

\begin{figure*}[!htbp]
		\centering
		\caption{The calculated weights using cosine similarity, \cref{eq:weights_cos} during the federated training process using \ac{t2t}-14-VIS model and \ac{niid} partitioning (each client has 4 classes). a) for the natural training, and b) for the adversarial training.}
		\resizebox{0.9\textwidth}{!}{%
			\setlength\tabcolsep{1.5pt}
			\begin{tabular}{c}
				\begin{tabular}{l}\subfloat[]{\includegraphics[width=\textwidth]{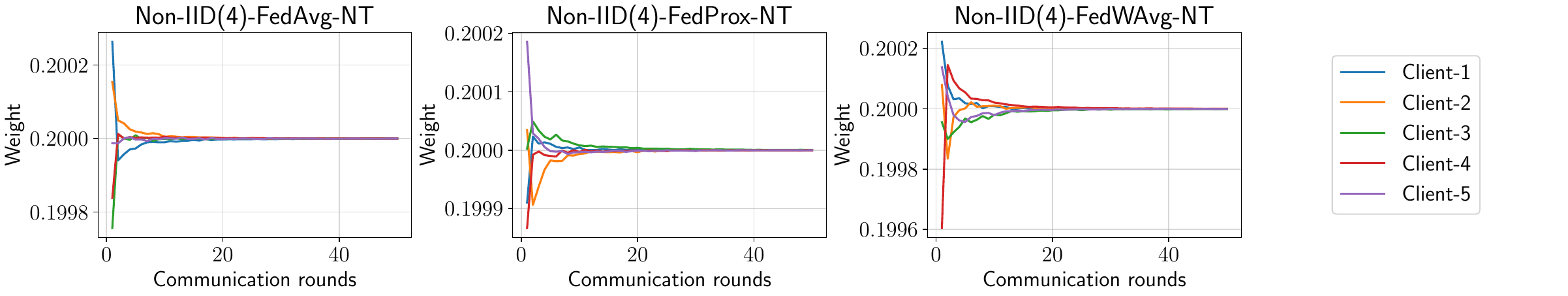}\label{fig:wieghts_t2t_vis_niid4_b_nt}}\end{tabular}  \\
				\begin{tabular}{l}\subfloat[]{\includegraphics[width=\textwidth]{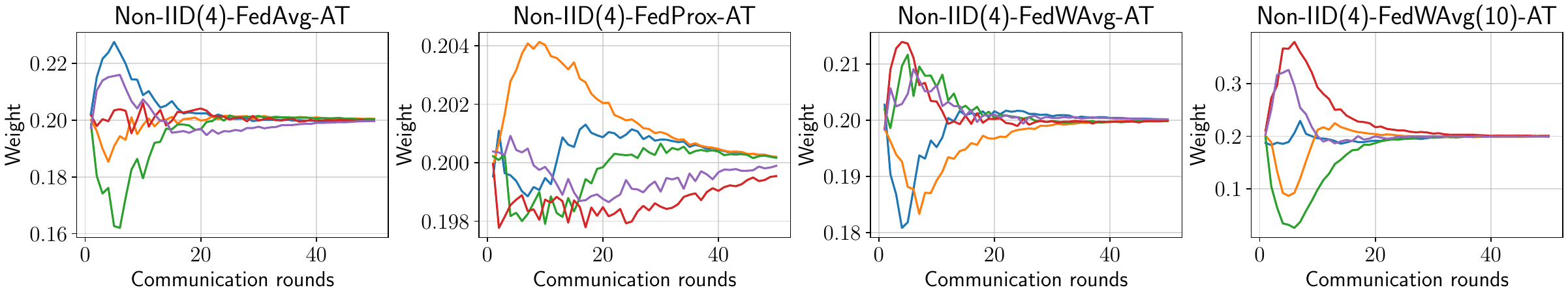}\label{fig:wieghts_t2t_vis_niid4_b_at}} \end{tabular} \\
			\end{tabular}
		}%

		\label{fig:wieghts_t2t_vis_niid4}
	\end{figure*}

\begin{figure*}[!htbp]
		\centering
		\caption{The calculated weights using cosine similarity, \cref{eq:weights_cos} during the federated training process using \ac{tnt}-S-CLS+VIS model and \ac{niid} partitioning (each client has 4 classes). a) for the natural training, and b) for the adversarial training.}
		\resizebox{0.9\textwidth}{!}{%
			\setlength\tabcolsep{1.5pt}
			\begin{tabular}{c}
				\begin{tabular}{l}\subfloat[]{\includegraphics[width=\textwidth]{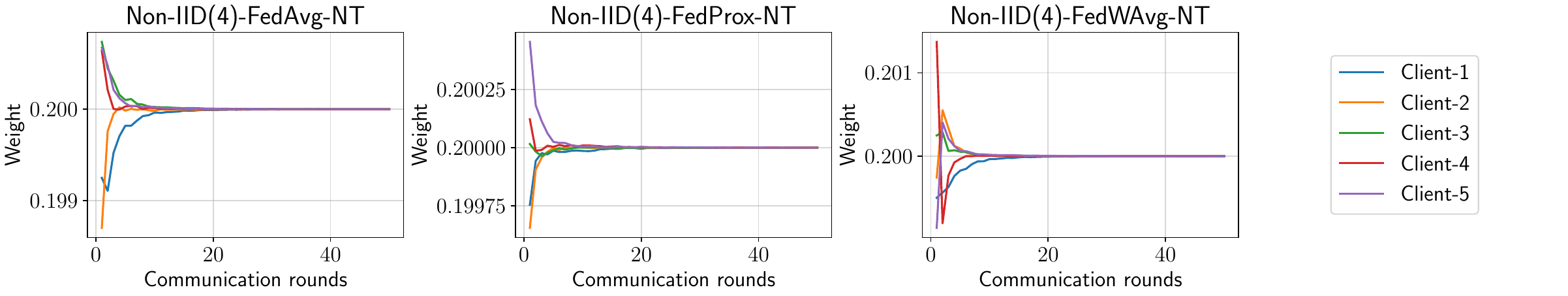}\label{fig:wieghts_tnts_cls_vis_niid_4_b_nt}}\end{tabular}  \\
				\begin{tabular}{l}\subfloat[]{\includegraphics[width=\textwidth]{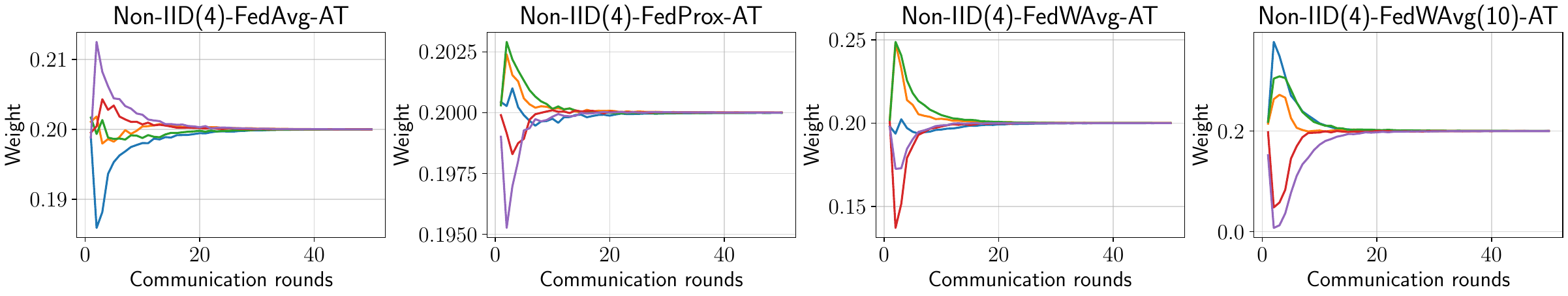}\label{fig:wieghts_tnts_cls_vis_niid_4_b_at}} \end{tabular} \\
			\end{tabular}
		}%

		\label{fig:wieghts_tnts_cls_vis_niid_4}
	\end{figure*}

	\vspace{-1mm}
	More figures to show the calculated cosine similarity weights for some of the tested models are shown in \cref{fig:wieghts_vits_cls_iid,fig:wieghts_vits_cls_niid4,fig:wieghts_vits_cls_niid2,fig:wieghts_t2t_vis_niid4,fig:wieghts_tnts_cls_vis_niid_4}
	
	\vspace{-5mm}
	\section{More figures for FedWAvg convergence}
	\label{app:more_convergence}
\begin{figure*}[!htbp]
		\centering
		\caption{The accuracy and robust accuracy of \ac{vit}-S-CLS model in the \ac{fat} process for different aggregation methods. The accuracies against the communication rounds under the \ac{nt} process using \ac{iid}, \ac{niid}(4), and \ac{niid}(2) are shown in a), b), and c), respectively. The accuracies (top) and the robust accuracies (bottom) against the communication rounds under the \ac{at} process using \ac{iid}, \ac{niid}(4), and \ac{niid}(2) are shown in d), e), and f), respectively.}
		\resizebox{0.9\textwidth}{!}{%
			\setlength\tabcolsep{1.5pt}
			\begin{tabular}{ccc}
				\begin{tabular}{l}\subfloat[]{\includegraphics[width=0.33\textwidth]{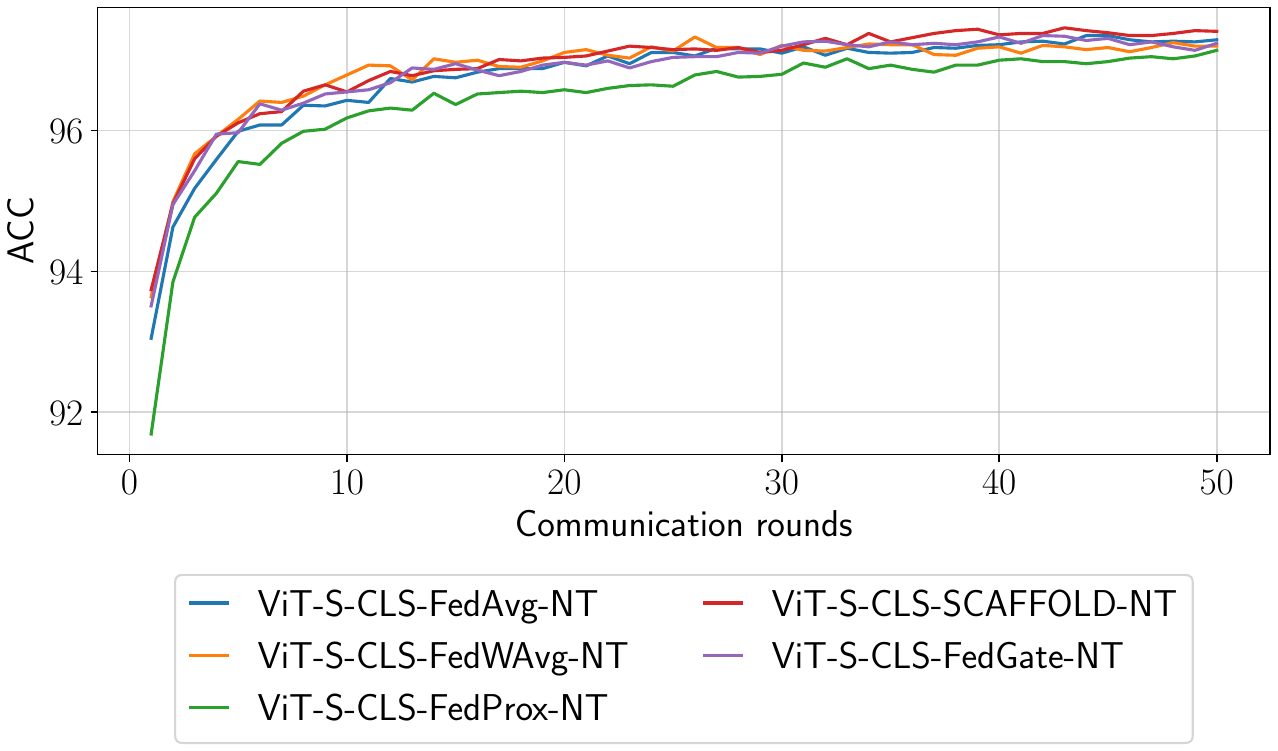}\label{fig:conv_vits_cls_iid_nt}}\end{tabular} &
				\begin{tabular}{l}\subfloat[]{\includegraphics[width=0.33\textwidth]{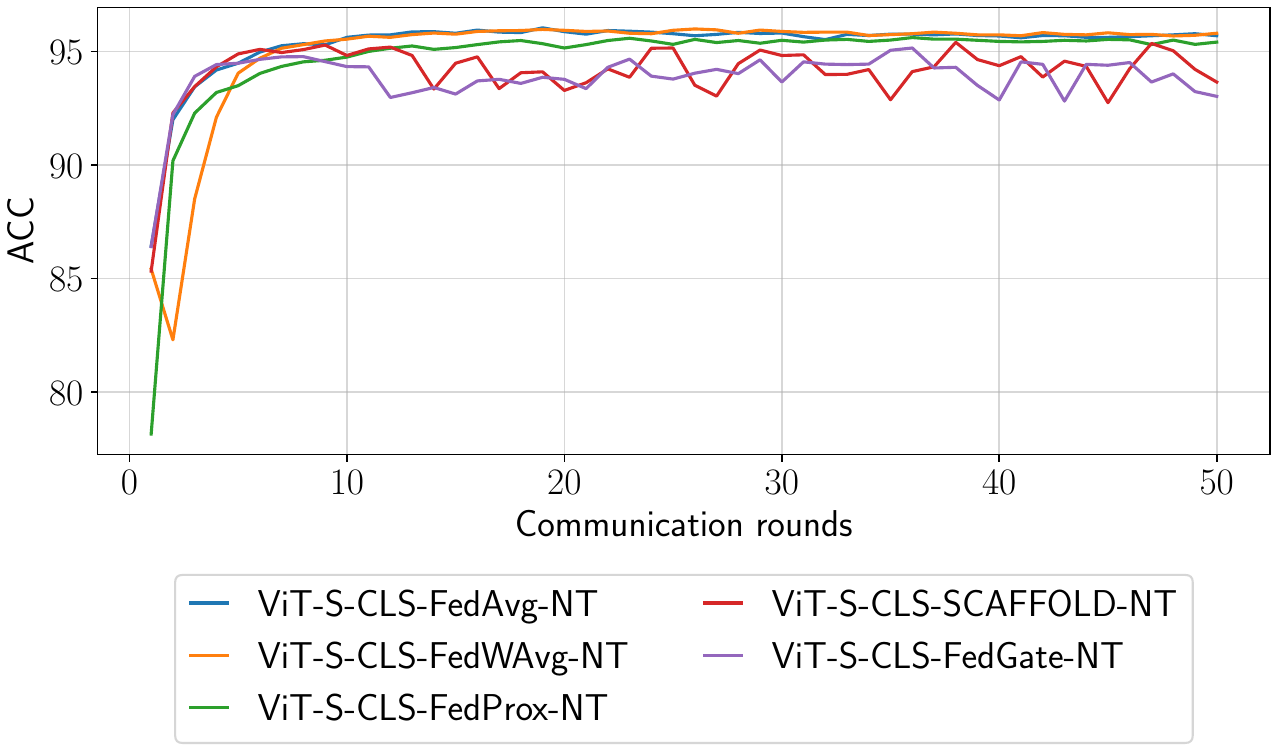}\label{fig:conv_vits_cls_niid4_nt}} \end{tabular} &
				\begin{tabular}{l}\subfloat[]{\includegraphics[width=0.33\textwidth]{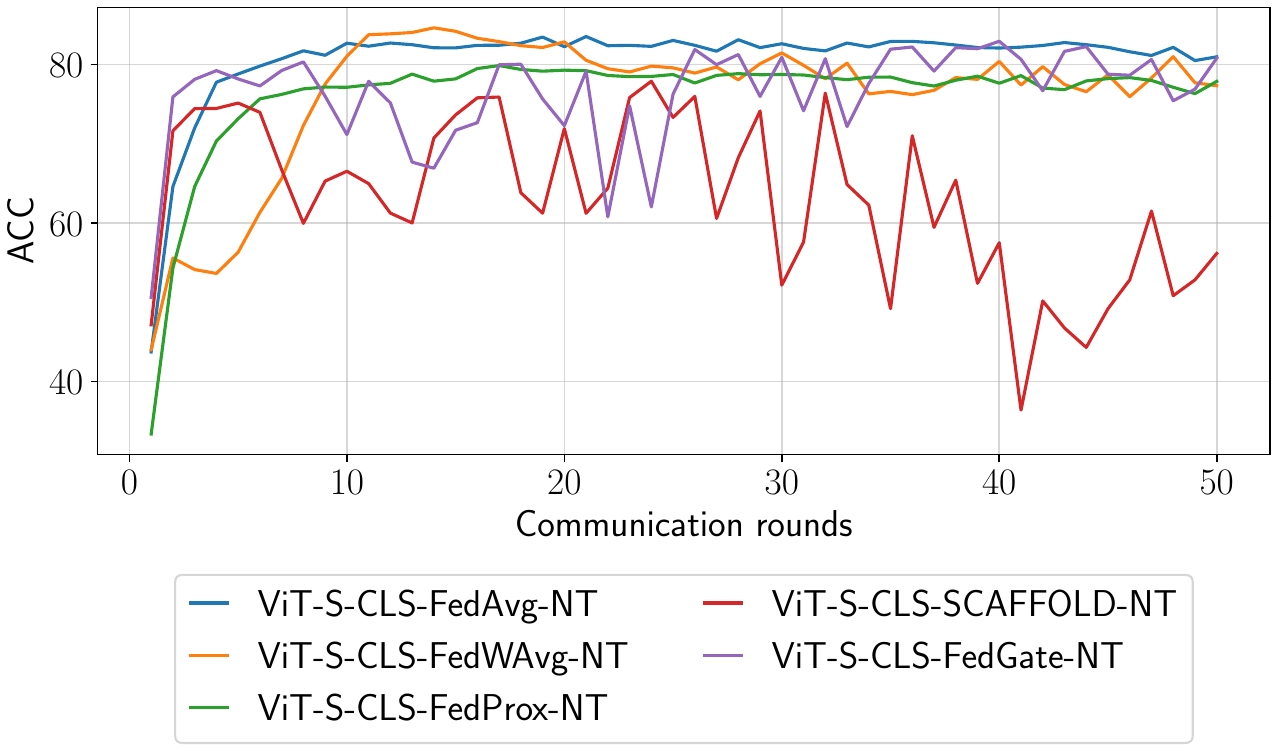}\label{fig:conv_vits_cls_niid2_nt}} \end{tabular} \\
				
				\begin{tabular}{l}\subfloat[]{\includegraphics[width=0.33\textwidth]{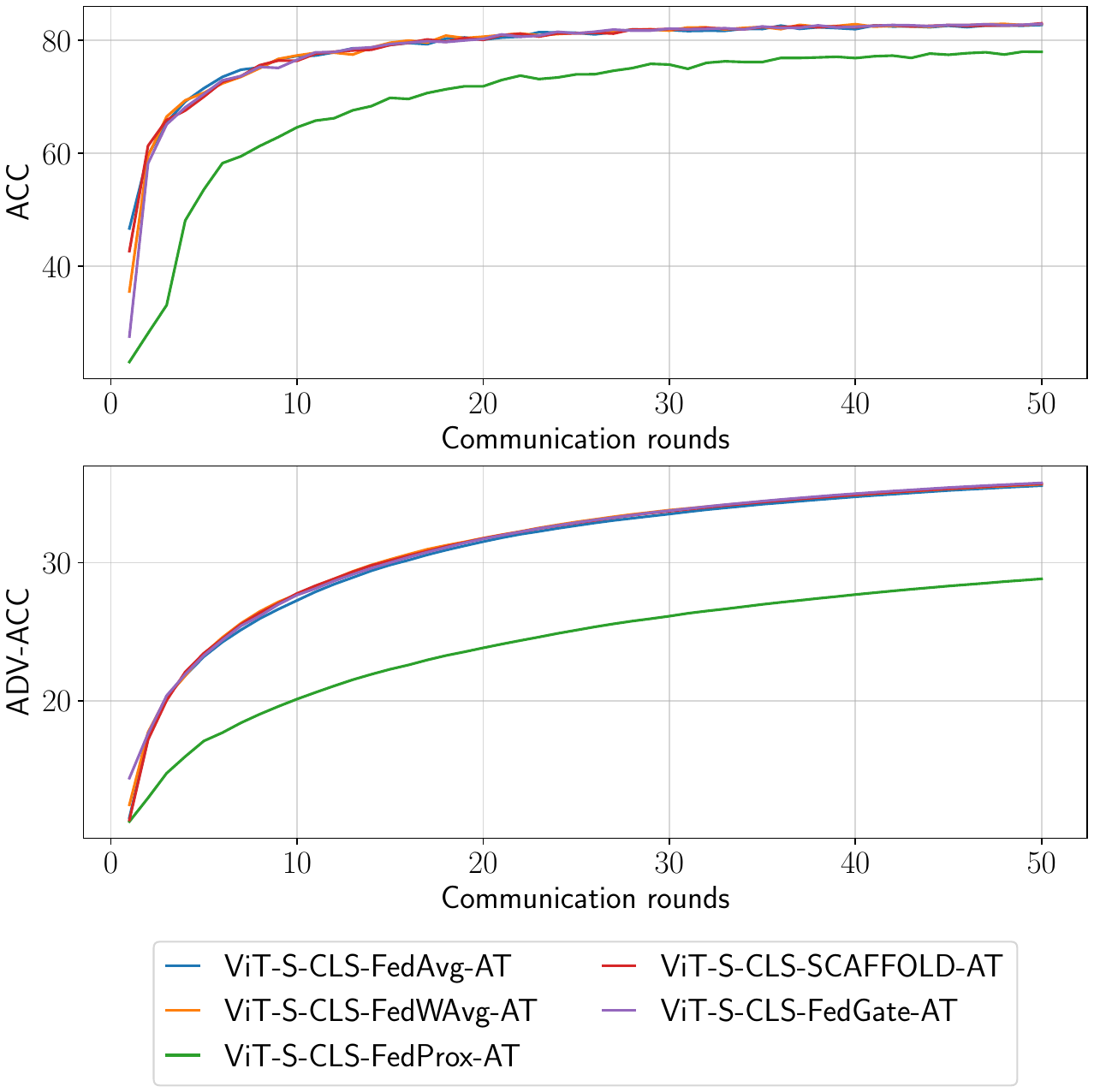}\label{fig:conv_vits_cls_iid_at}}\end{tabular} &
				\begin{tabular}{l}\subfloat[]{\includegraphics[width=0.33\textwidth]{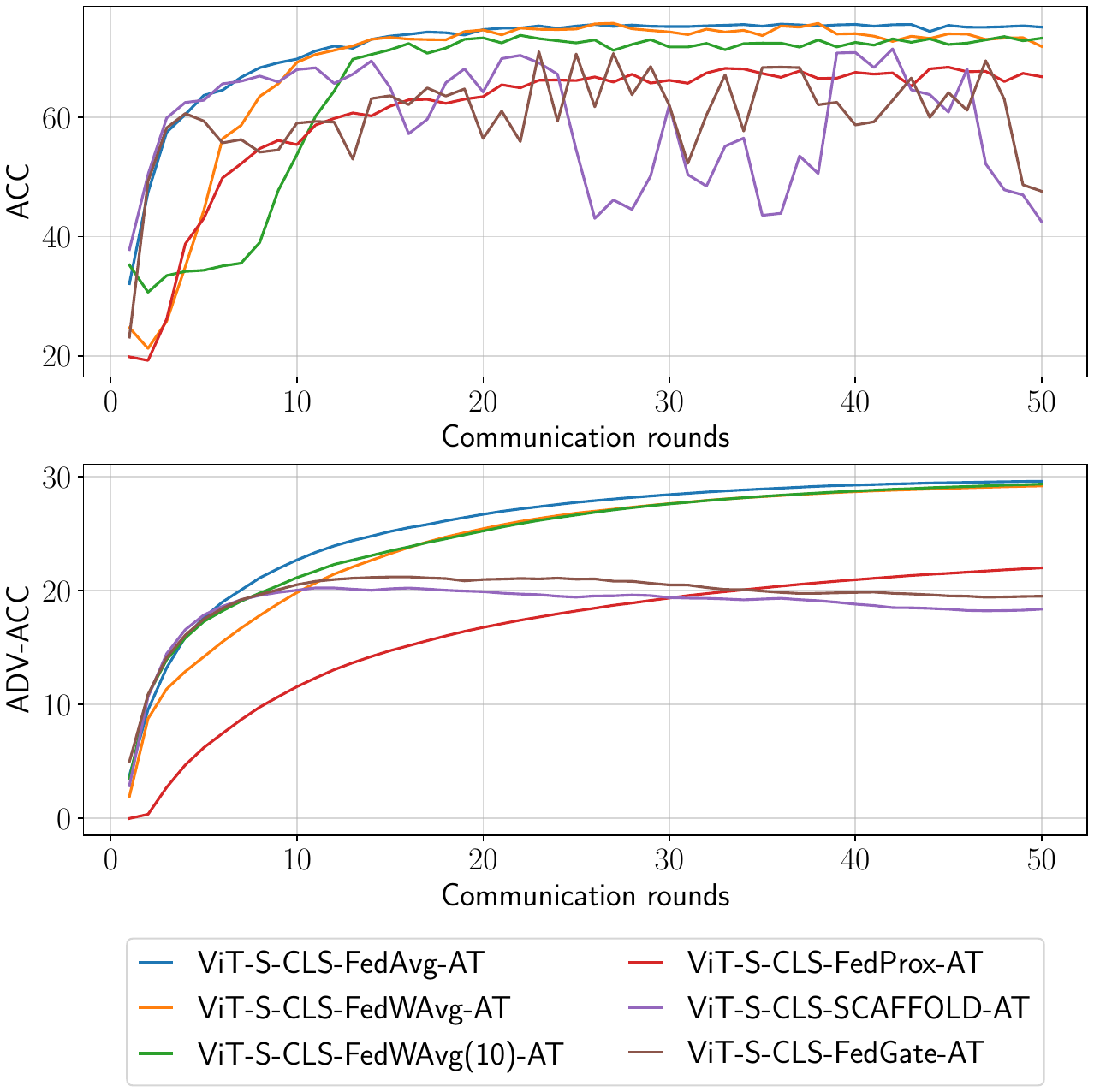}\label{fig:conv_vits_cls_niid4_at}} \end{tabular} &
				\begin{tabular}{l}\subfloat[]{\includegraphics[width=0.33\textwidth]{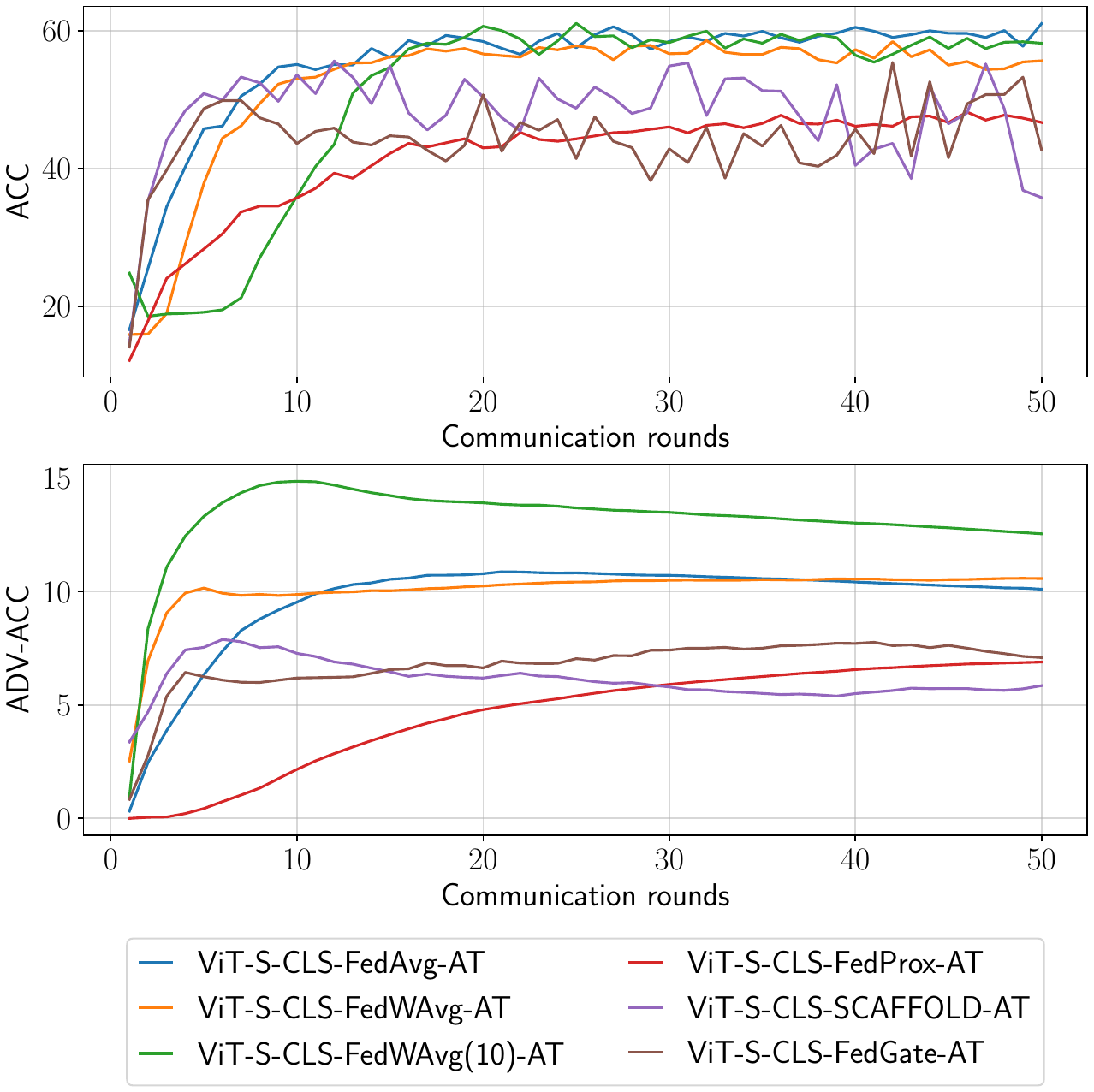}\label{fig:conv_vits_cls_niid2_at}} \end{tabular} \\
				
			\end{tabular}
		}%

		\label{fig:conv_vits_cls}
	\end{figure*}

\begin{figure*}[!htbp]
		\centering
		\caption{The accuracy and robust accuracy of \ac{t2t}-CLS model in the \ac{fat} process for different aggregation methods. The accuracies against the communication rounds under the \ac{nt} process using \ac{iid}, \ac{niid}(4), and \ac{niid}(2) are shown in a), b), and c), respectively. The accuracies (top) and the robust accuracies (bottom) against the communication rounds under the \ac{at} process using \ac{iid}, \ac{niid}(4), and \ac{niid}(2) are shown in d), e), and f), respectively.}
		\resizebox{0.9\textwidth}{!}{%
			\setlength\tabcolsep{1.5pt}
			\begin{tabular}{ccc}
				\begin{tabular}{l}\subfloat[]{\includegraphics[width=0.33\textwidth]{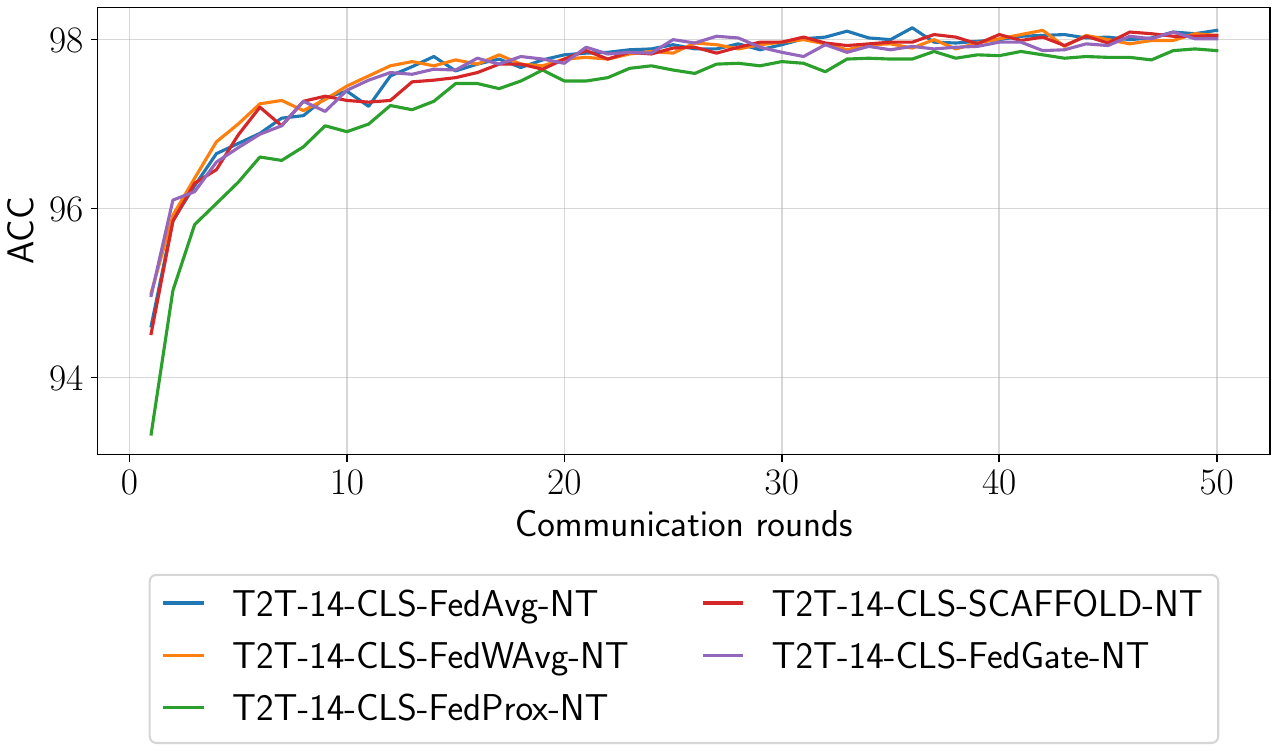}\label{fig:conv_t2t_cls_iid_nt}}\end{tabular} &
				\begin{tabular}{l}\subfloat[]{\includegraphics[width=0.33\textwidth]{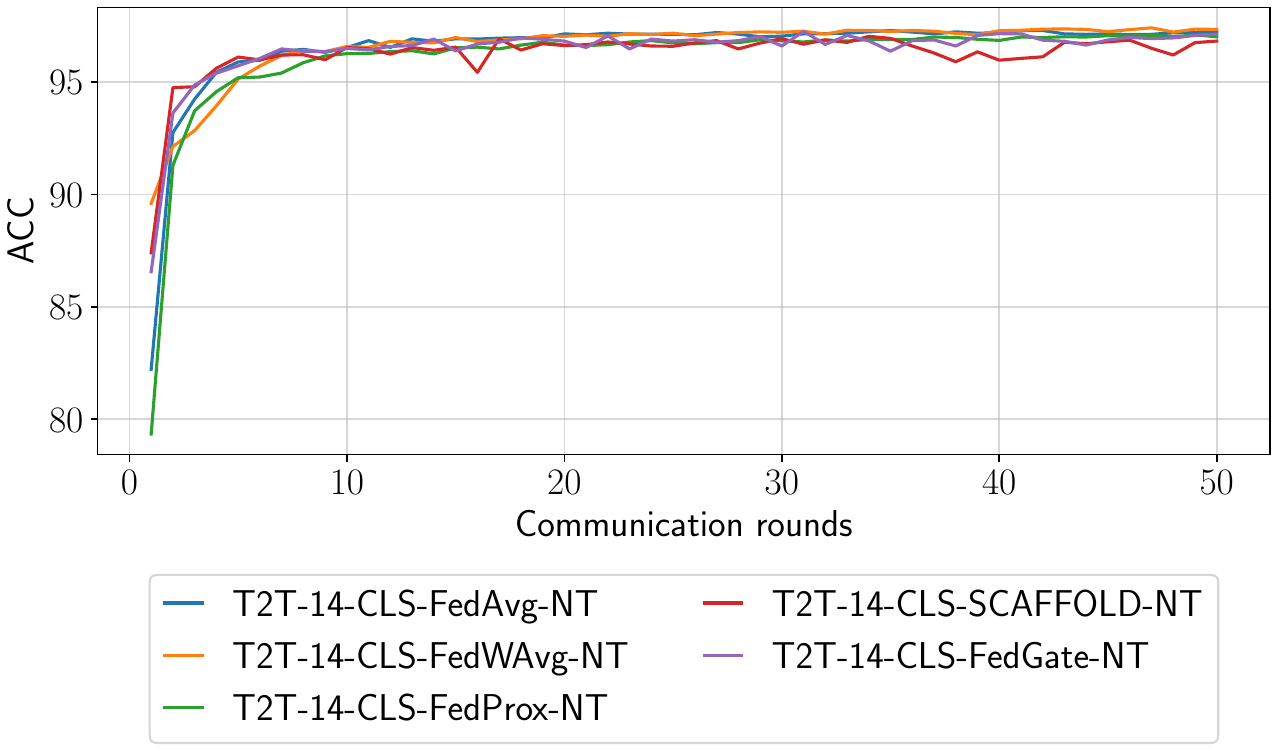}\label{fig:conv_t2t_cls_niid4_nt}} \end{tabular} &
				\begin{tabular}{l}\subfloat[]{\includegraphics[width=0.33\textwidth]{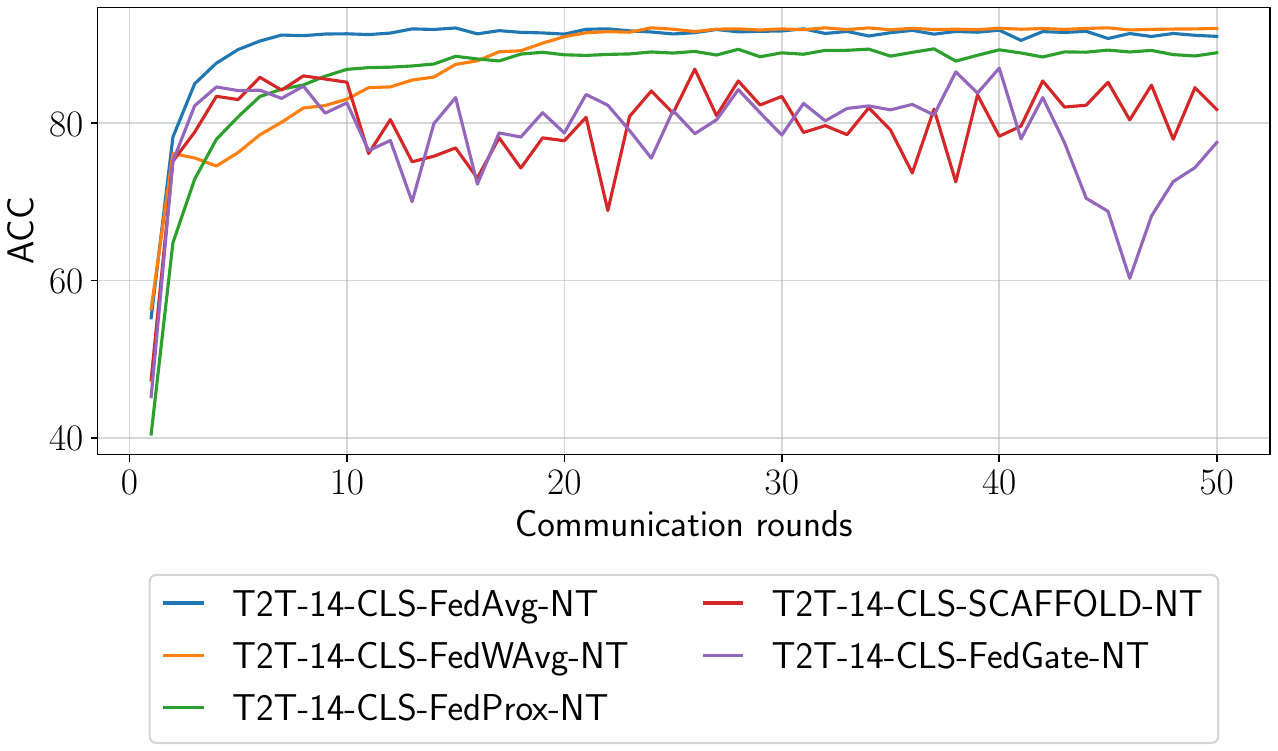}\label{fig:conv_t2t_cls_niid2_nt}} \end{tabular} \\
				
				\begin{tabular}{l}\subfloat[]{\includegraphics[width=0.33\textwidth]{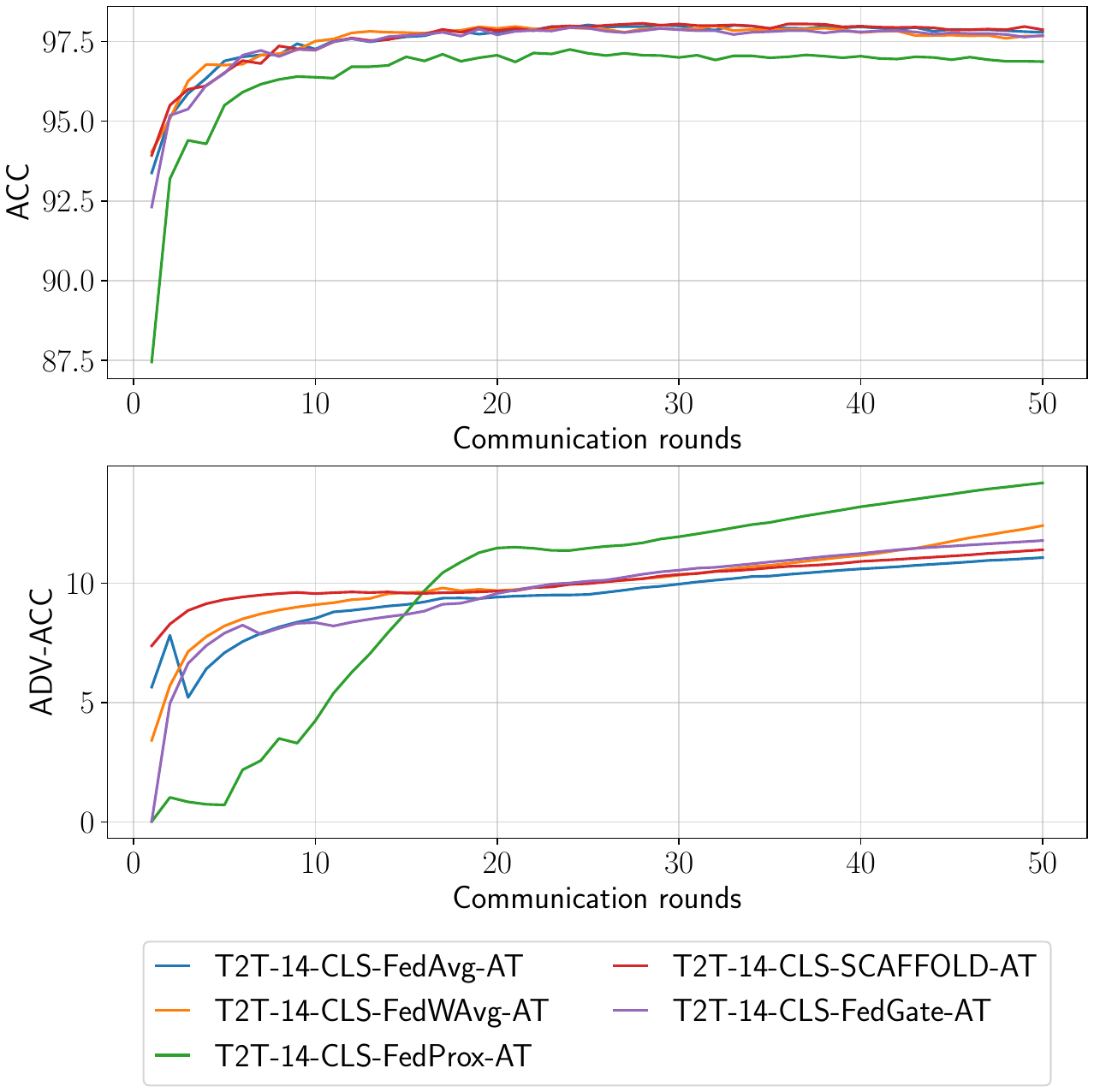}\label{fig:conv_t2t_cls_iid_at}}\end{tabular} &
				\begin{tabular}{l}\subfloat[]{\includegraphics[width=0.33\textwidth]{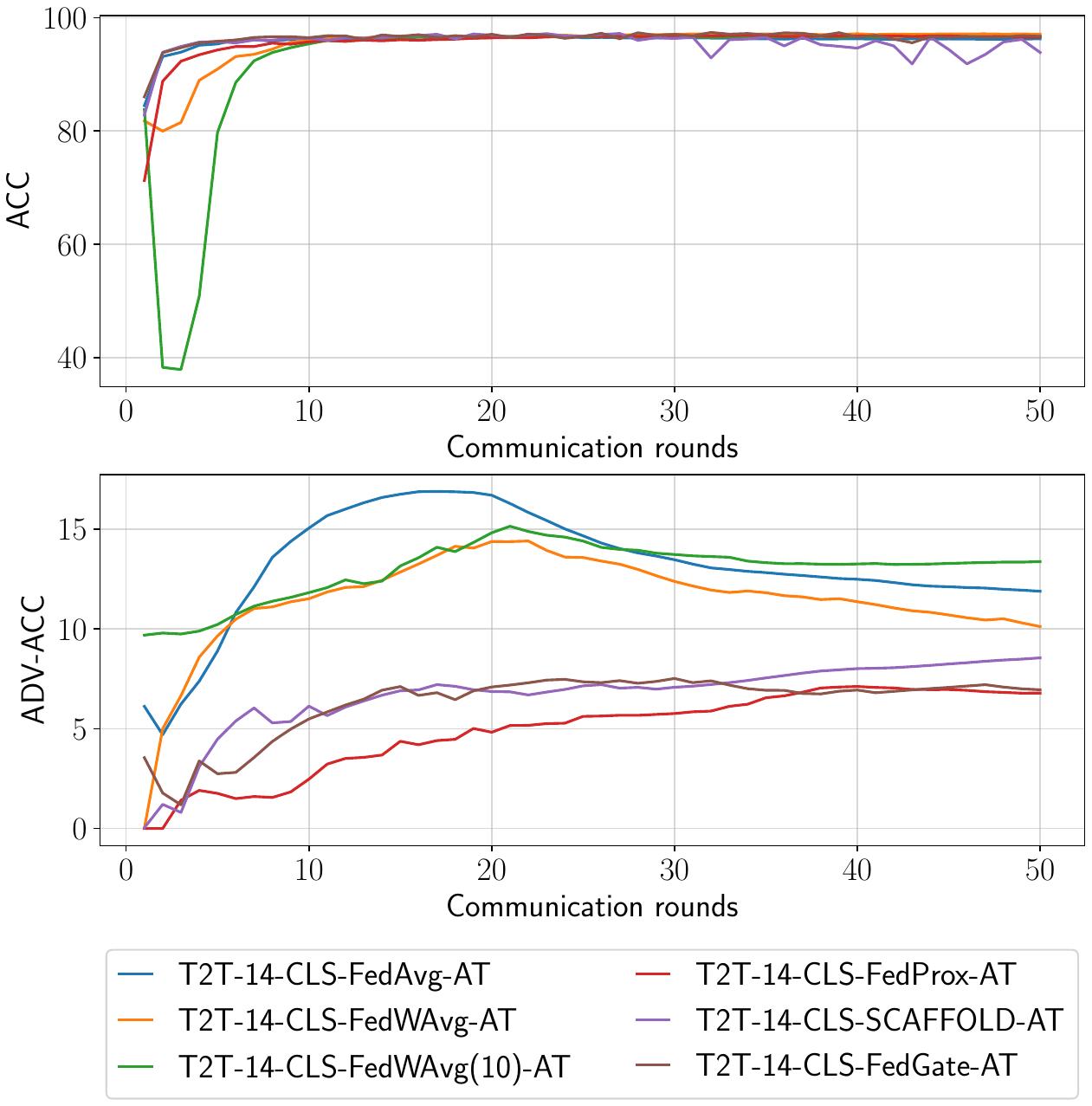}\label{fig:conv_t2t_cls_niid4_at}} \end{tabular} &
				\begin{tabular}{l}\subfloat[]{\includegraphics[width=0.33\textwidth]{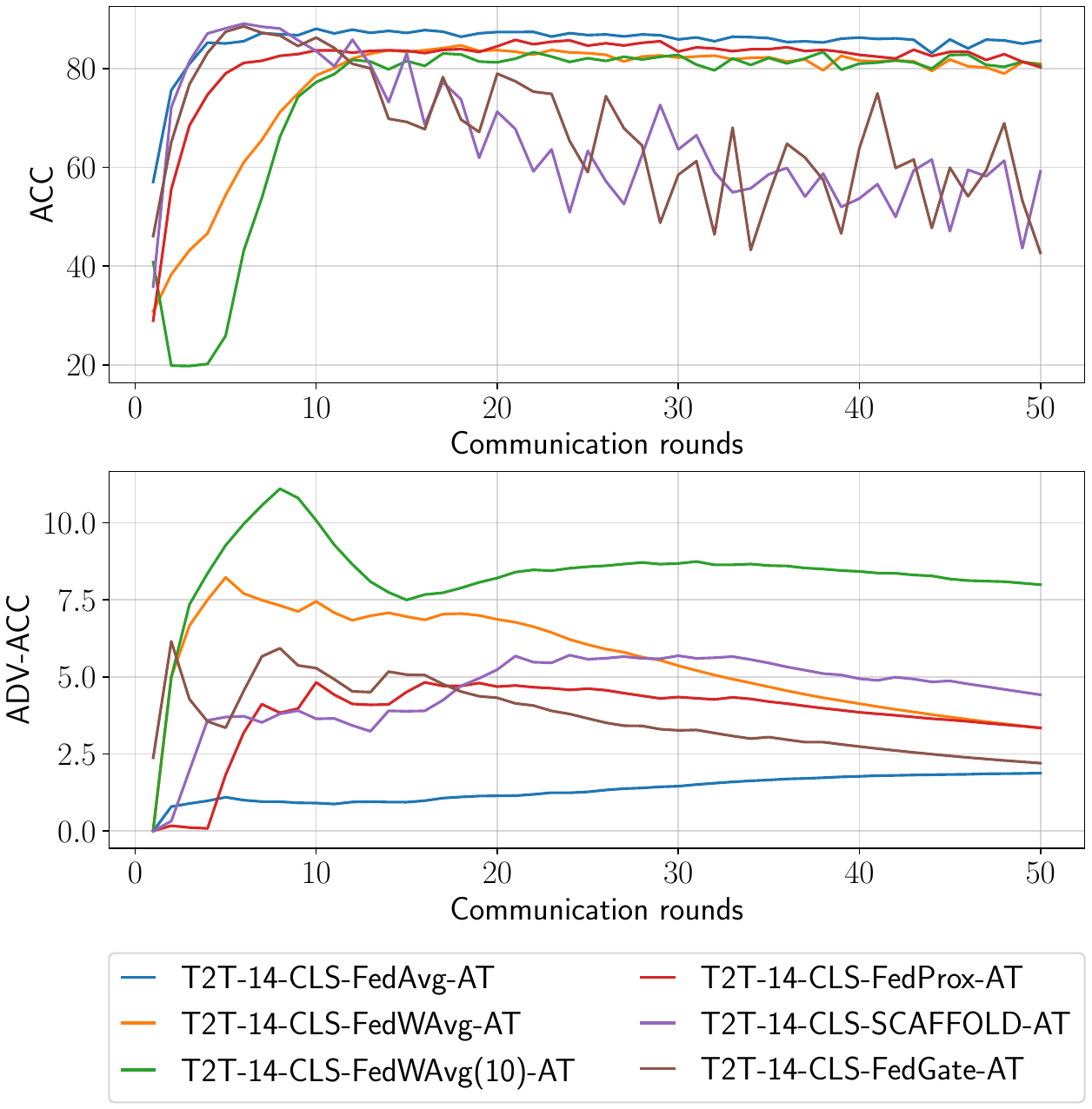}\label{fig:conv_t2t_cls_niid2_at}} \end{tabular} \\
				
			\end{tabular}
		}%

		\label{fig:conv_t2t_cls}
	\end{figure*}

\begin{figure*}[!htbp]
		\centering
		\caption{The accuracy and robust accuracy of \ac{tnt}-CLS+VIS model in the \ac{fat} process for different aggregation methods. The accuracies against the communication rounds under the \ac{nt} process using \ac{iid}, \ac{niid}(4), and \ac{niid}(2) are shown in a), b), and c), respectively. The accuracies (top) and the robust accuracies (bottom) against the communication rounds under the \ac{at} process using \ac{iid}, \ac{niid}(4), and \ac{niid}(2) are shown in d), e), and f), respectively.}
		\resizebox{0.9\textwidth}{!}{%
			\setlength\tabcolsep{1.5pt}
			\begin{tabular}{ccc}
				\begin{tabular}{l}\subfloat[]{\includegraphics[width=0.33\textwidth]{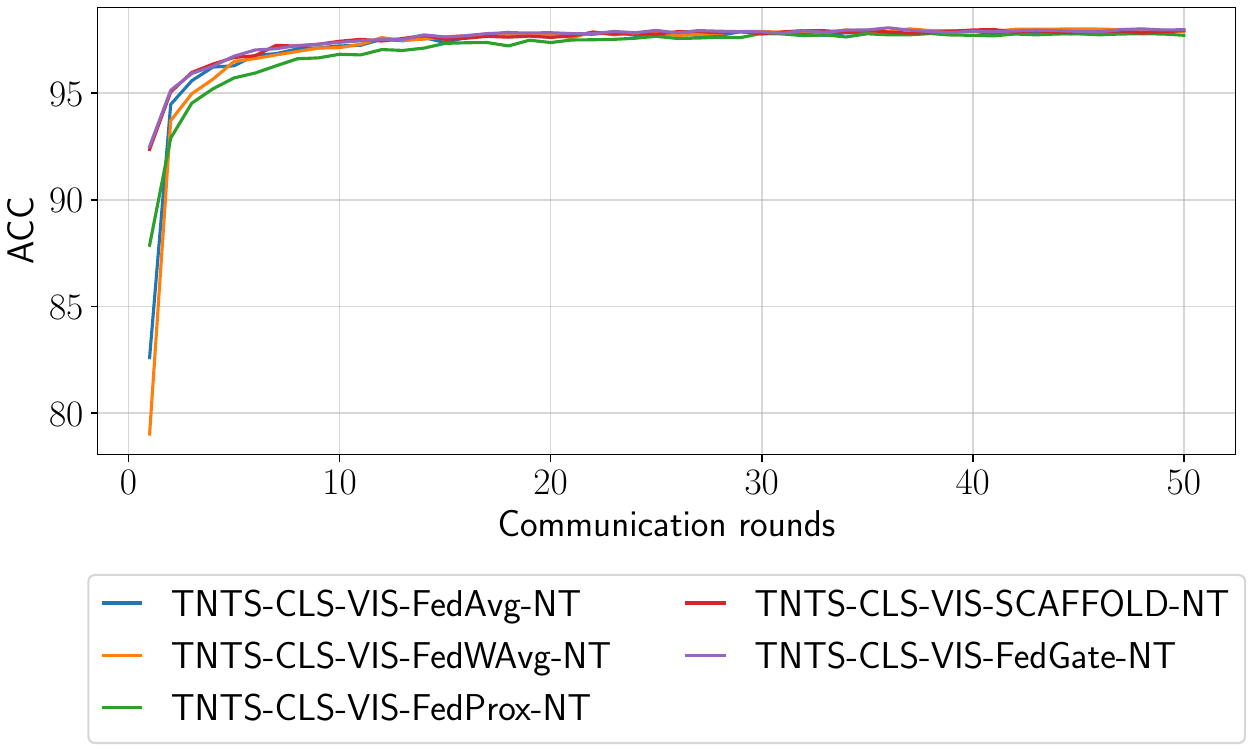}\label{fig:conv_tnt_cls_vis_iid_nt}}\end{tabular} &
				\begin{tabular}{l}\subfloat[]{\includegraphics[width=0.33\textwidth]{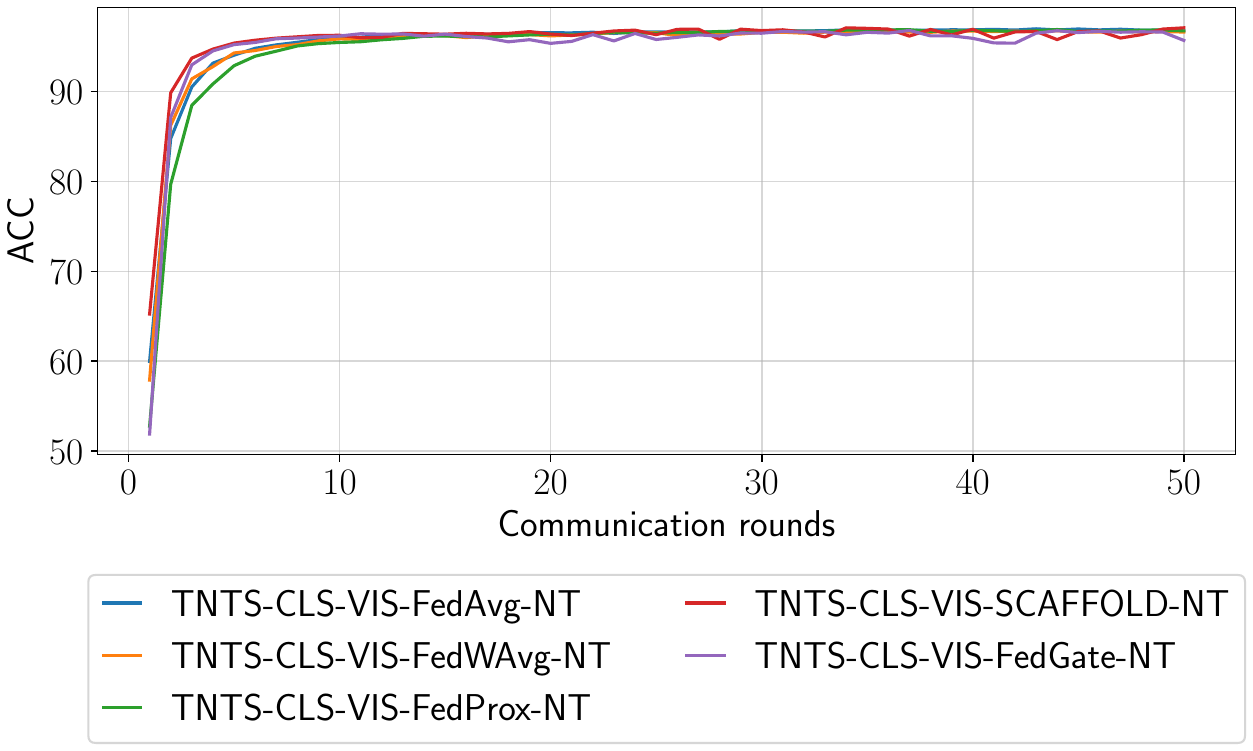}\label{fig:conv_tnt_cls_vis_niid4_nt}} \end{tabular} &
				\begin{tabular}{l}\subfloat[]{\includegraphics[width=0.33\textwidth]{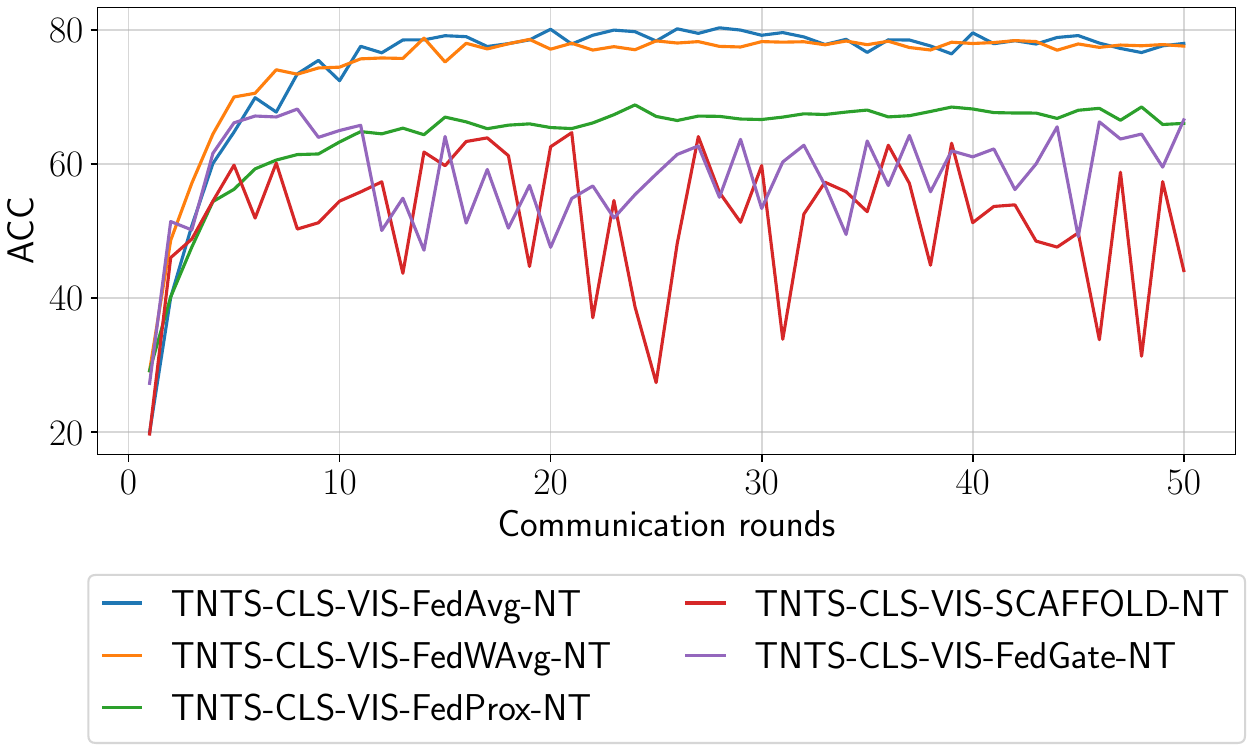}\label{fig:conv_tnt_cls_vis_niid2_nt}} \end{tabular} \\
				
				\begin{tabular}{l}\subfloat[]{\includegraphics[width=0.33\textwidth]{figs/Figure2d.pdf}\label{fig:conv_tnt_cls_vis_iid_at}}\end{tabular} &
				\begin{tabular}{l}\subfloat[]{\includegraphics[width=0.33\textwidth]{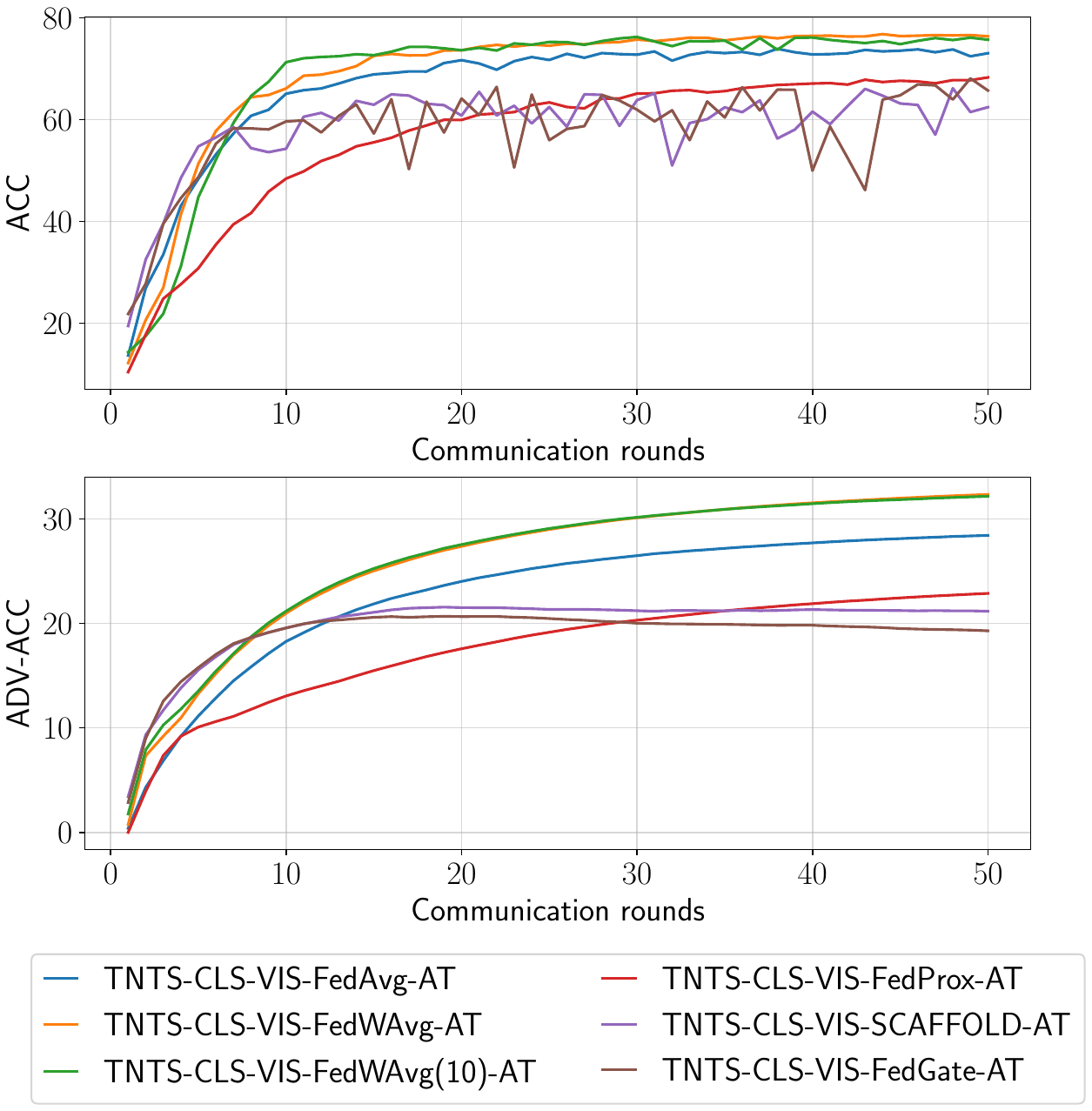}\label{fig:conv_tnt_cls_vis_niid4_at}} \end{tabular} &
				\begin{tabular}{l}\subfloat[]{\includegraphics[width=0.33\textwidth]{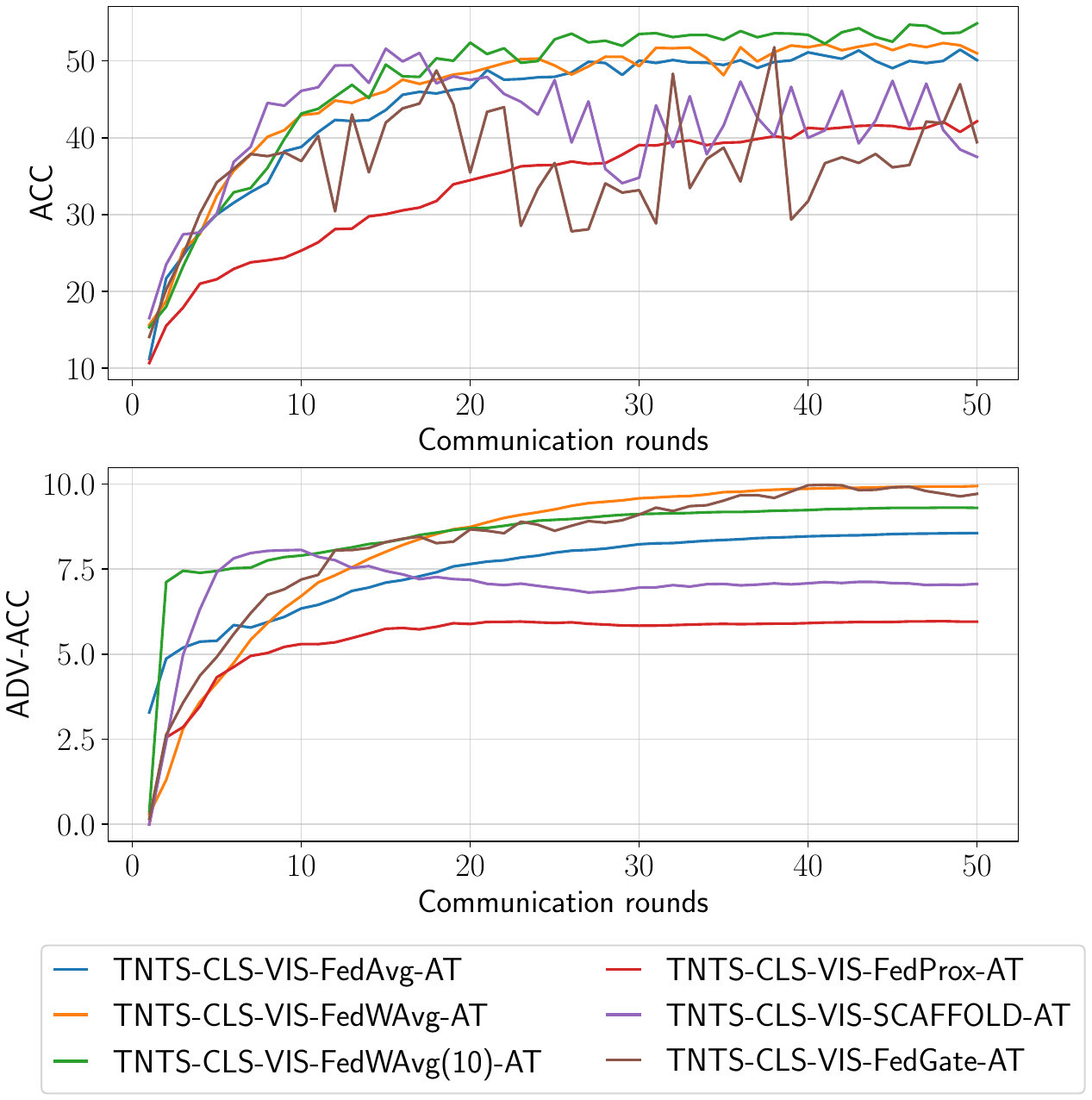}\label{fig:conv_tnt_cls_vis_niid2_at}} \end{tabular} \\
				
			\end{tabular}
		}%

		\label{fig:conv_tnt_cls_vis}
	\end{figure*}

\begin{figure*}[!htbp]
		\centering
		\caption{The accuracy and the robust accuracy of \ac{tnt}-CLS+VIS model with loss values in the \ac{fat} process for different aggregation methods. The accuracies (top) and the robust accuracies (bottom) against the communication rounds under the \ac{at} process using \ac{niid}(4), and \ac{niid}(2) are shown in a) and b), respectively.}
		\resizebox{\textwidth}{!}{%
			\setlength\tabcolsep{1.5pt}
			\begin{tabular}{ccc}
				\begin{tabular}{l}\subfloat[]{\includegraphics[width=0.5\textwidth]{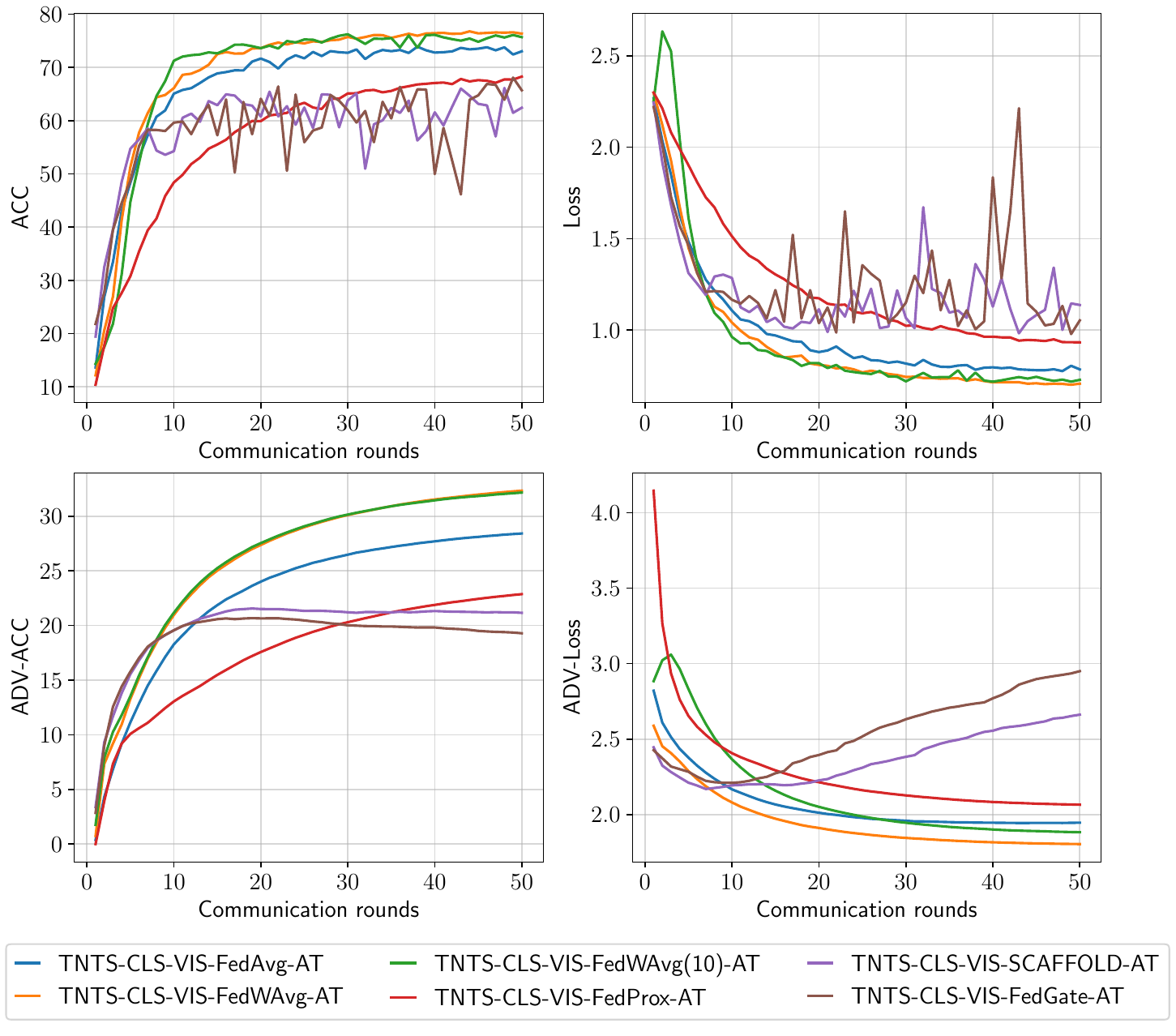}\label{fig:conv_loss_tnt_cls_vis_niid4_nt}} \end{tabular} &
				\begin{tabular}{l}\subfloat[]{\includegraphics[width=0.5\textwidth]{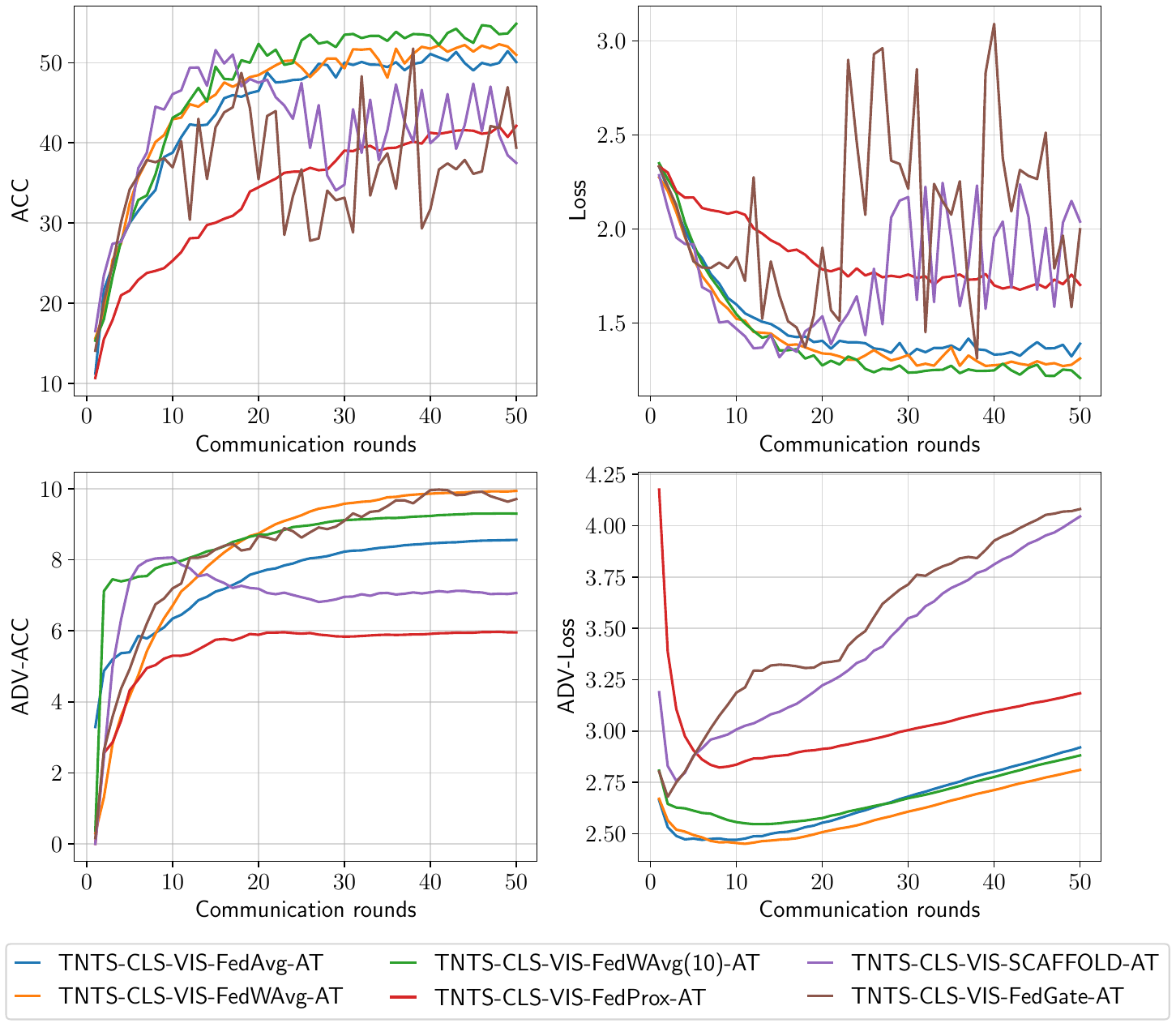}\label{fig:conv_loss_tnt_cls_vis_niid2_nt}} \end{tabular} \\
			\end{tabular}
		}%
		\label{fig:conv_loss_tnt_cls_vis}
	\end{figure*}

\begin{figure*}[!htbp]
		\centering
		\caption{The accuracy and the robust accuracy of \ac{vit}-CLS model with loss values in the \ac{fat} process for different aggregation methods. The accuracies (top) and the robust accuracies (bottom) against the communication rounds under the \ac{at} process using \ac{niid}(4), and \ac{niid}(2) are shown in a) and b), respectively.}
		\resizebox{\textwidth}{!}{%
			\setlength\tabcolsep{1.5pt}
			\begin{tabular}{ccc}
				\begin{tabular}{l}\subfloat[]{\includegraphics[width=0.5\textwidth]{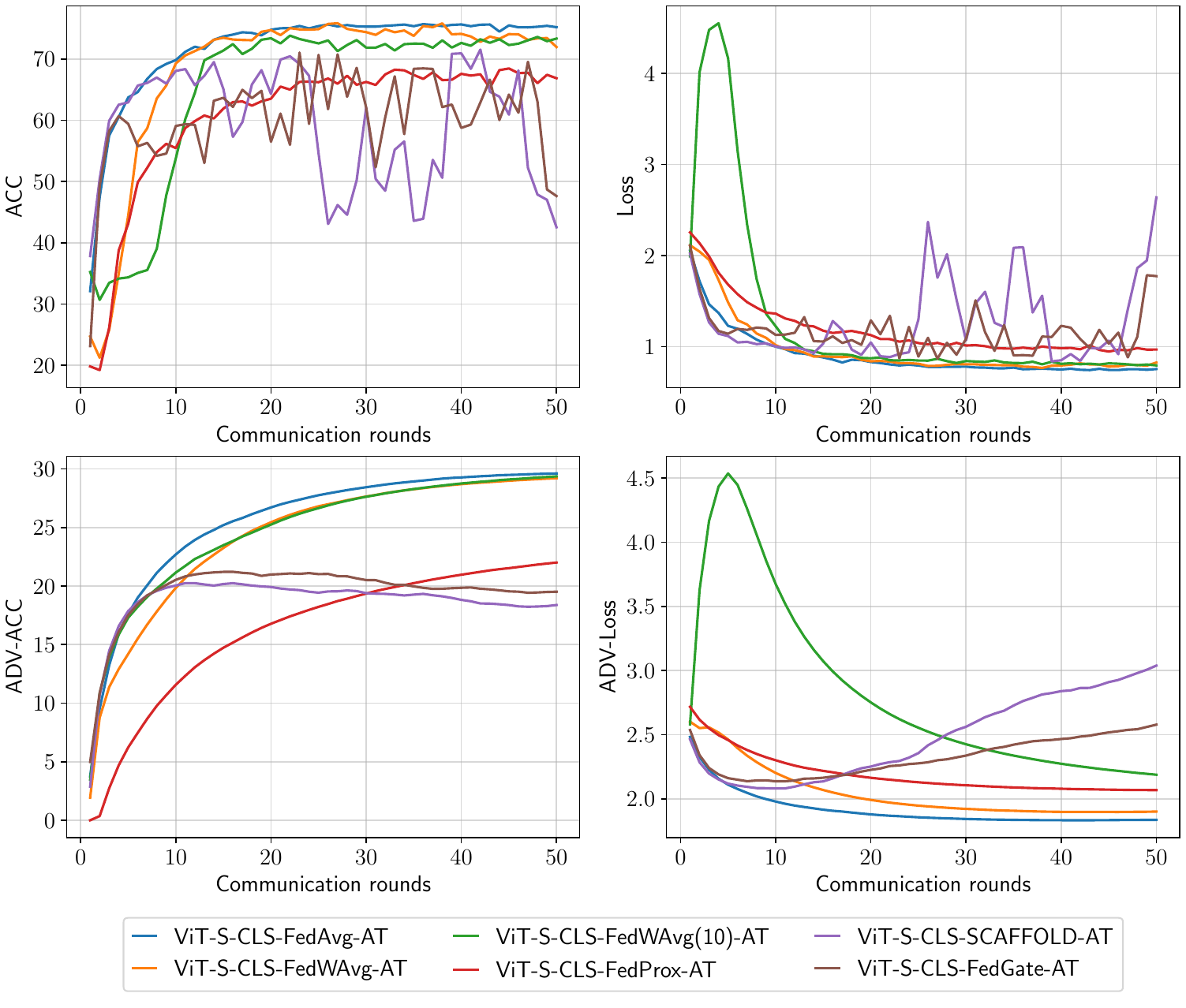}\label{fig:conv_loss_vits_cls_niid4_nt}} \end{tabular} &
				\begin{tabular}{l}\subfloat[]{\includegraphics[width=0.5\textwidth]{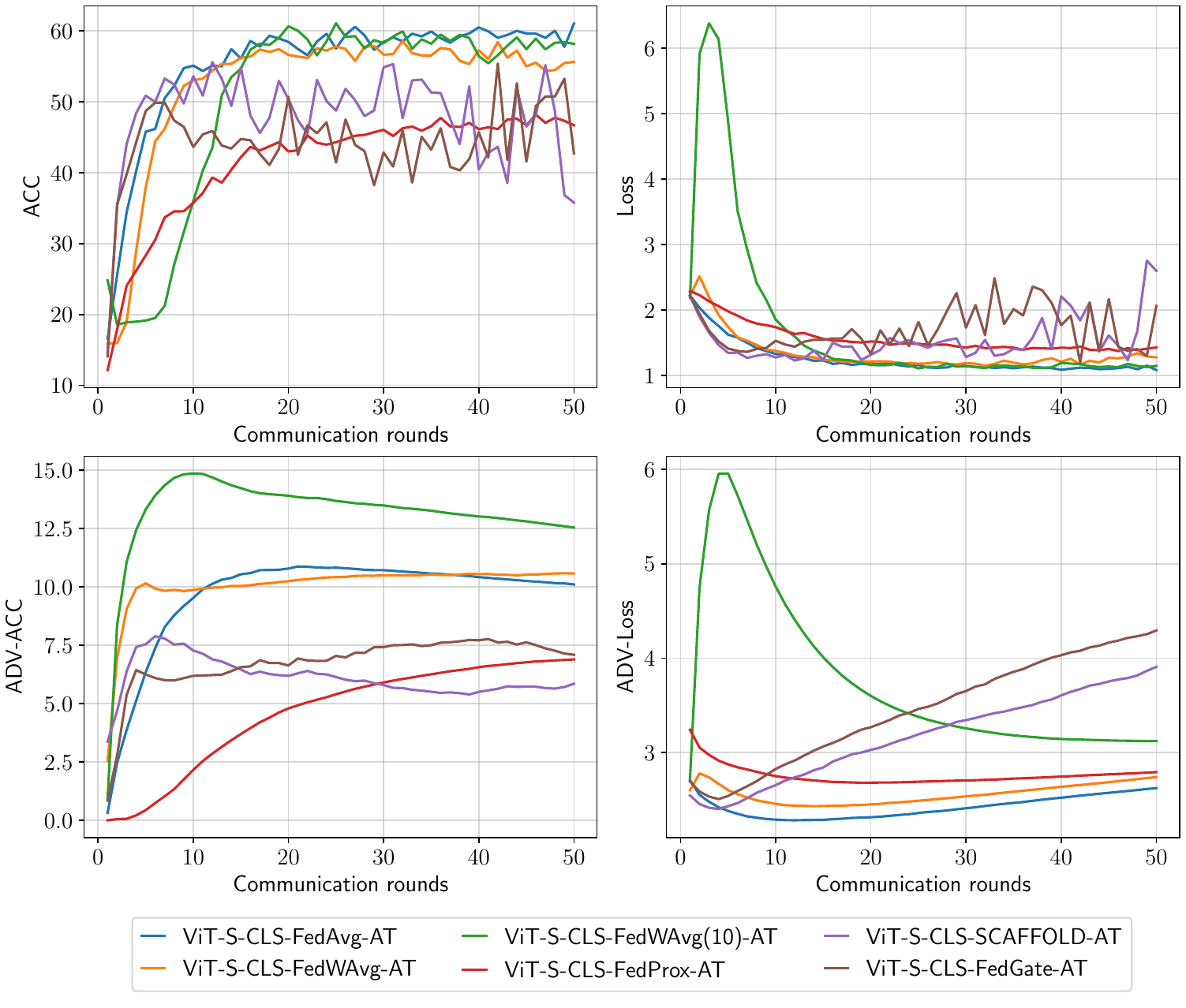}\label{fig:conv_loss_vits_cls_niid2_nt}} \end{tabular} \\
			\end{tabular}
		}%
		\label{fig:conv_loss_vits_cls}
	\end{figure*}

\begin{figure*}[!htbp]
		\centering
		\caption{The accuracy and the robust accuracy of \ac{t2t}-CLS model with loss values in the \ac{fat} process for different aggregation methods. The accuracies (top) and the robust accuracies (bottom) against the communication rounds under the \ac{at} process using \ac{niid}(4), and \ac{niid}(2) are shown in a) and b), respectively.}
		\resizebox{\textwidth}{!}{%
			\setlength\tabcolsep{1.5pt}
			\begin{tabular}{ccc}
				\begin{tabular}{l}\subfloat[]{\includegraphics[width=0.5\textwidth]{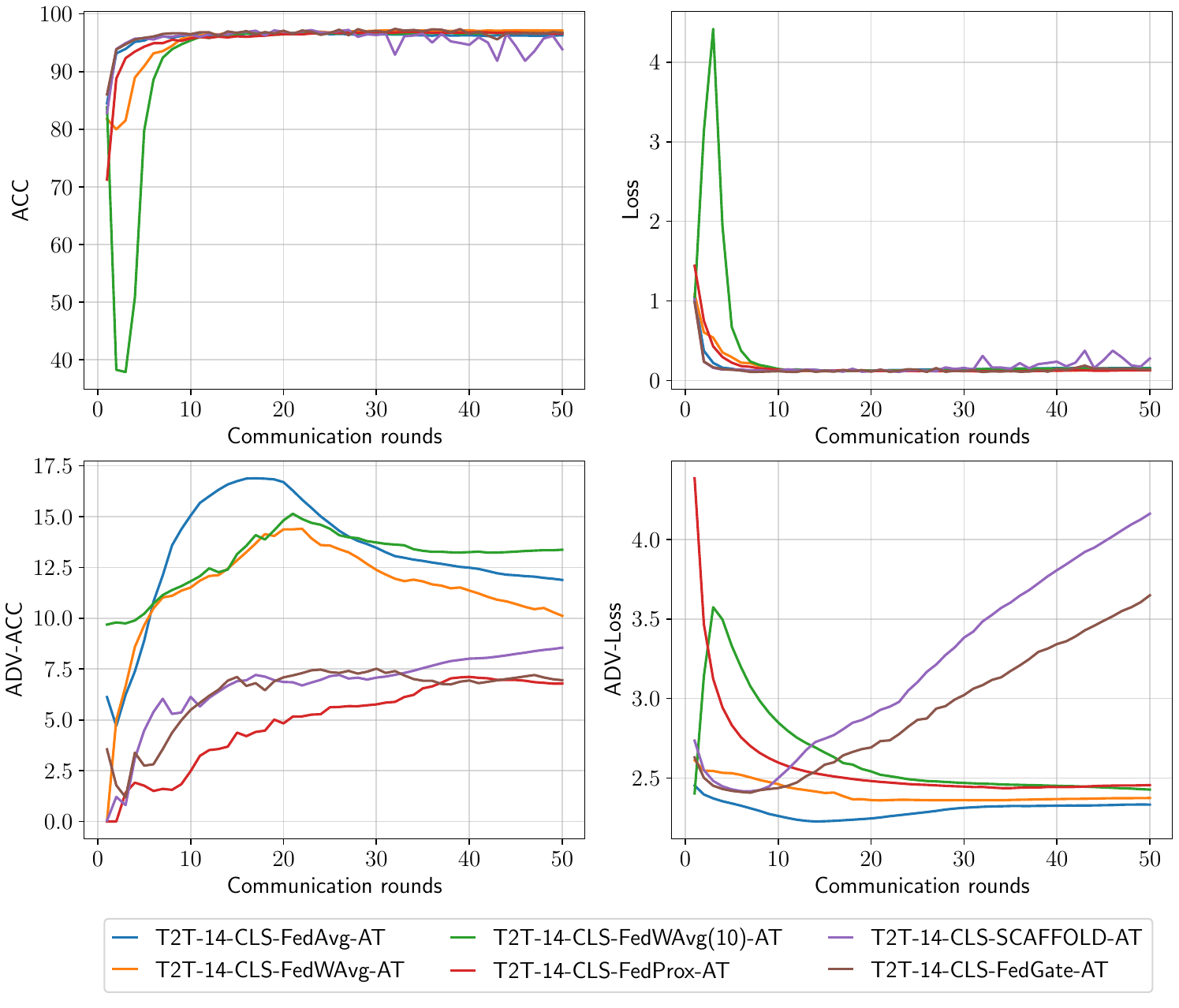}\label{fig:conv_loss_t2t_cls_niid4_nt}} \end{tabular} &
				\begin{tabular}{l}\subfloat[]{\includegraphics[width=0.5\textwidth]{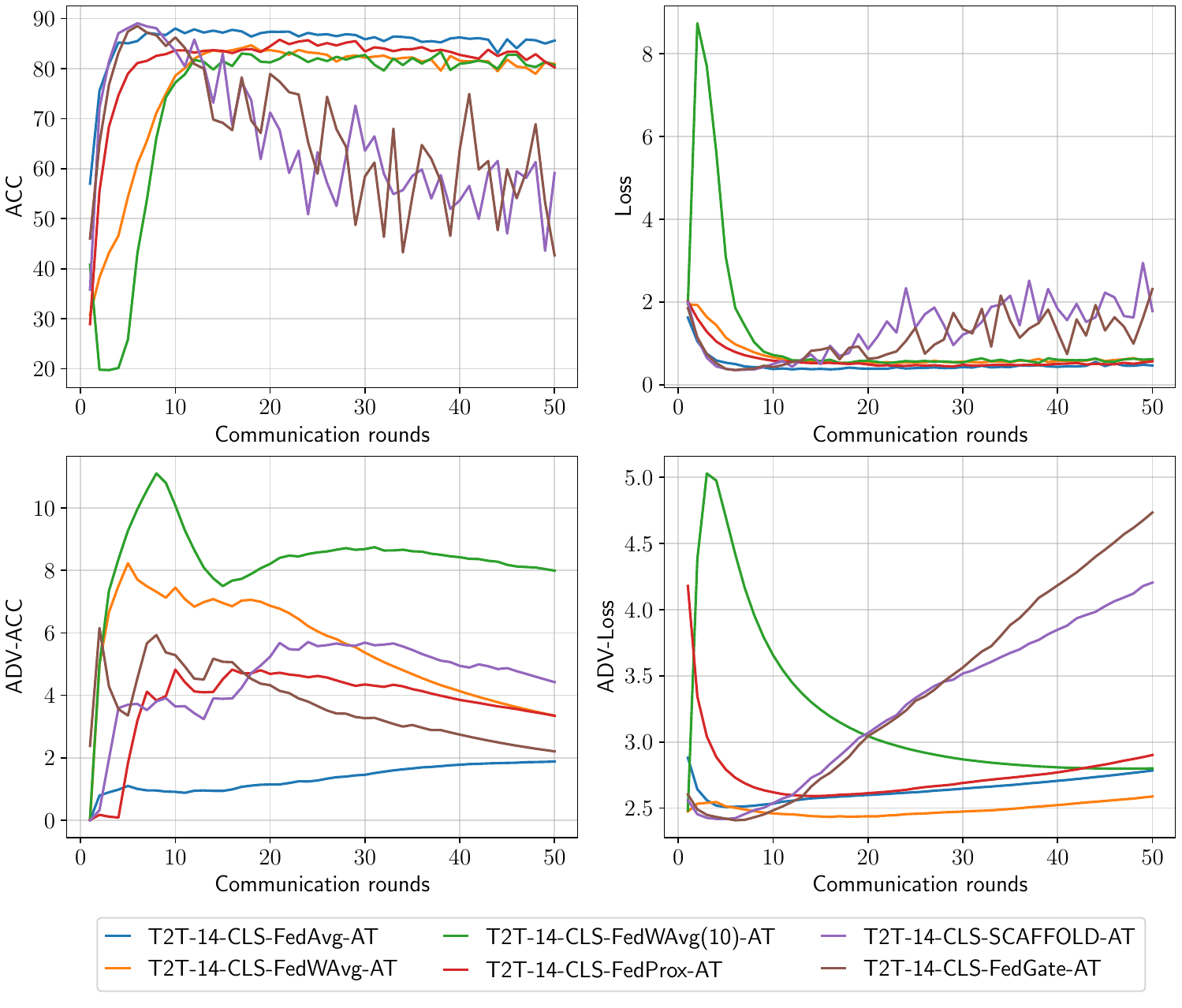}\label{fig:conv_loss_t2t_cls_niid2_nt}} \end{tabular} \\
			\end{tabular}
		}%
		\label{fig:conv_loss_t2t_cls}
	\end{figure*}

\begin{figure*}[!htbp]
		\centering
		\caption{The accuracy in training (dashed) and testing (solid) of \ac{vit}-CLS model with loss values in the \ac{fat} process for different aggregation methods. The accuracies against the communication rounds under the \ac{at} process using \ac{niid}(4), and \ac{niid}(2) are shown in a), and b) respectively. For better visualization, the dashed line is shifted one round forward.}
		\resizebox{\textwidth}{!}{%
			\setlength\tabcolsep{1.5pt}
			\begin{tabular}{ccc}
				\begin{tabular}{l}\subfloat[]{\includegraphics[width=0.5\textwidth]{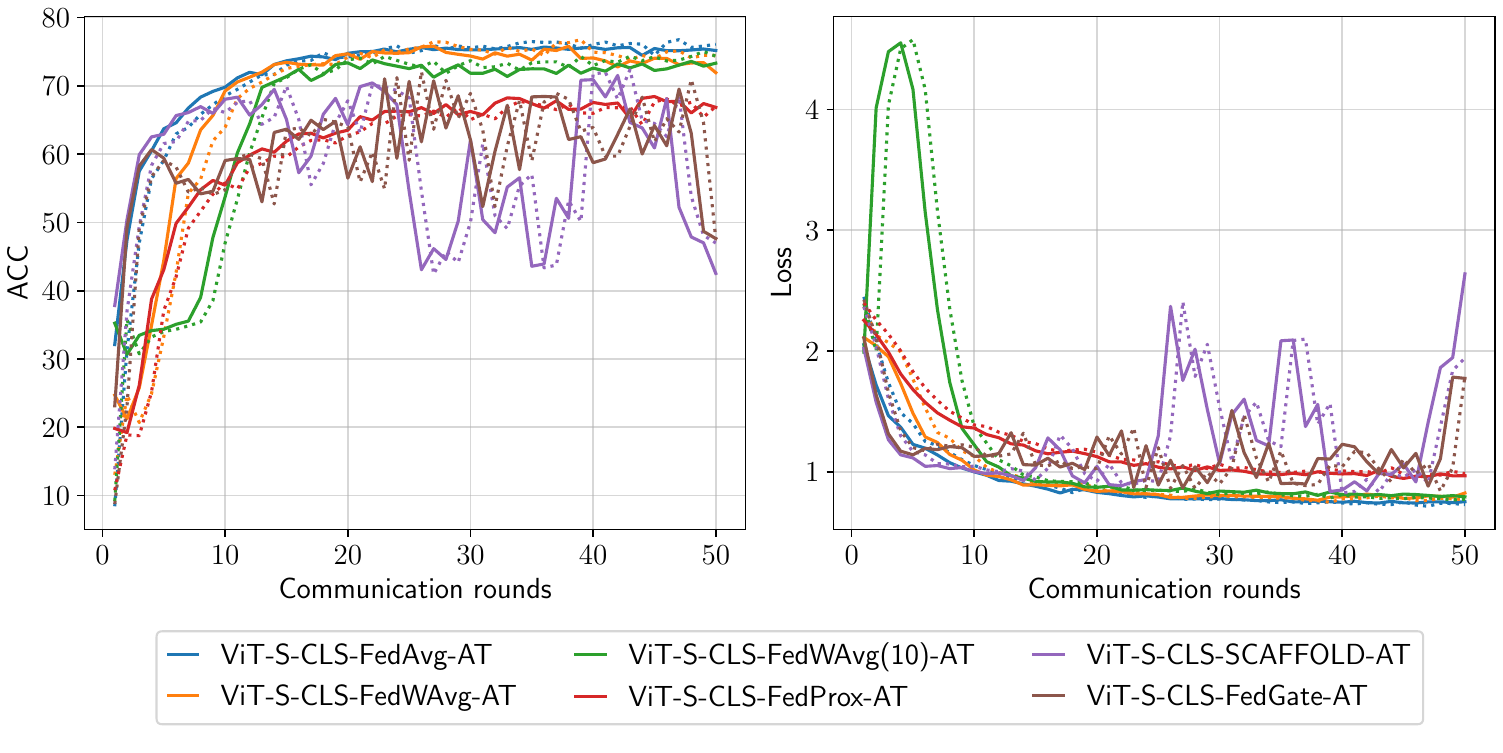}\label{fig:train_test_loss_vit_cls_niid4_nt}} \end{tabular} &
				\begin{tabular}{l}\subfloat[]{\includegraphics[width=0.5\textwidth]{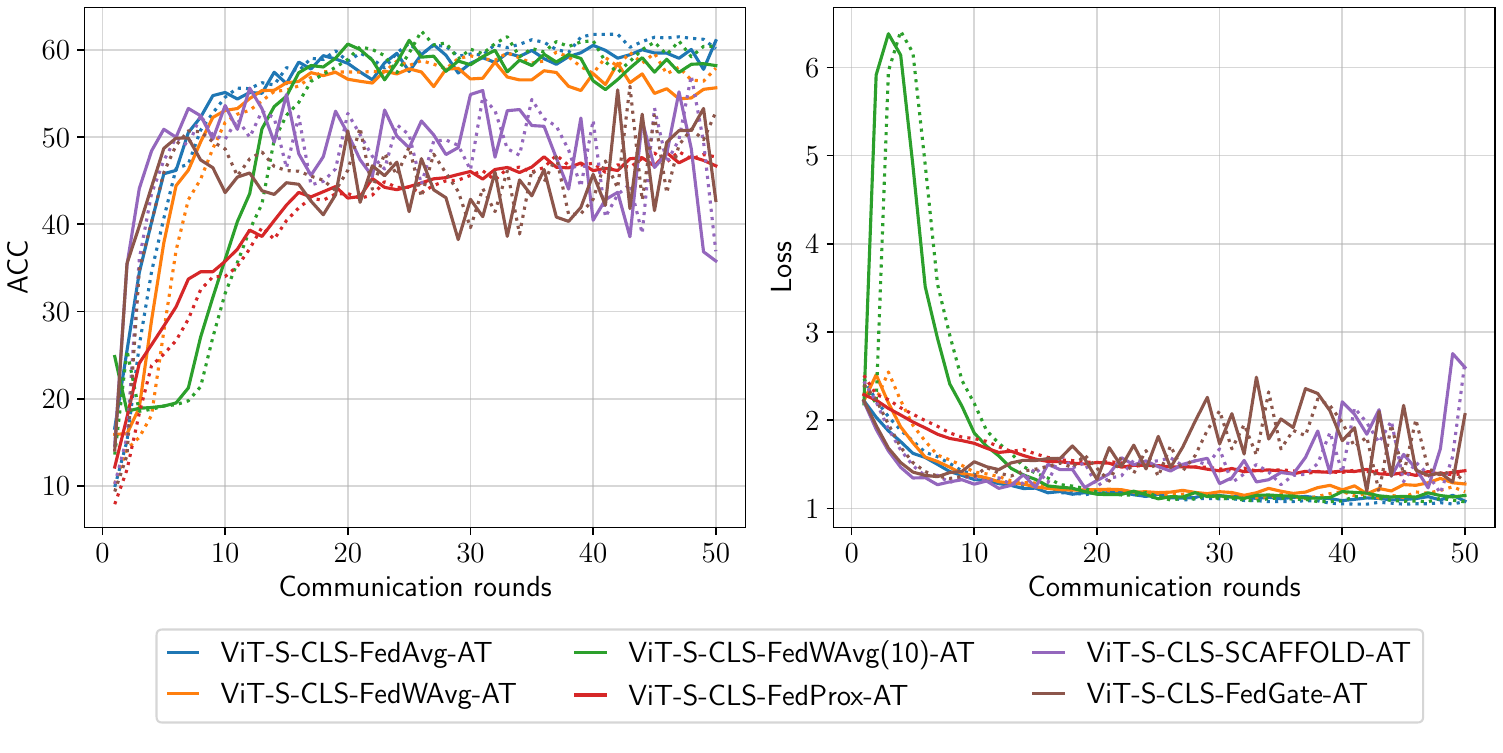}\label{fig:train_test_loss_vit_cls_niid2_nt}} \end{tabular} \\
			\end{tabular}
		}%
		\label{fig:train_test_loss_vit_cls}
	\end{figure*}

\begin{figure*}[!htbp]
		\centering
		\caption{The accuracy in training (dashed) and testing (solid) of \ac{tnt}-CLS+VIS model with loss values in the \ac{fat} process for different aggregation methods. The accuracies against the communication rounds under the \ac{at} process using \ac{niid}(4), and \ac{niid}(2) are shown in a), and b) respectively. For better visualization, the dashed line is shifted one round forward.}
		\resizebox{\textwidth}{!}{%
			\setlength\tabcolsep{1.5pt}
			\begin{tabular}{ccc}
				\begin{tabular}{l}\subfloat[]{\includegraphics[width=0.5\textwidth]{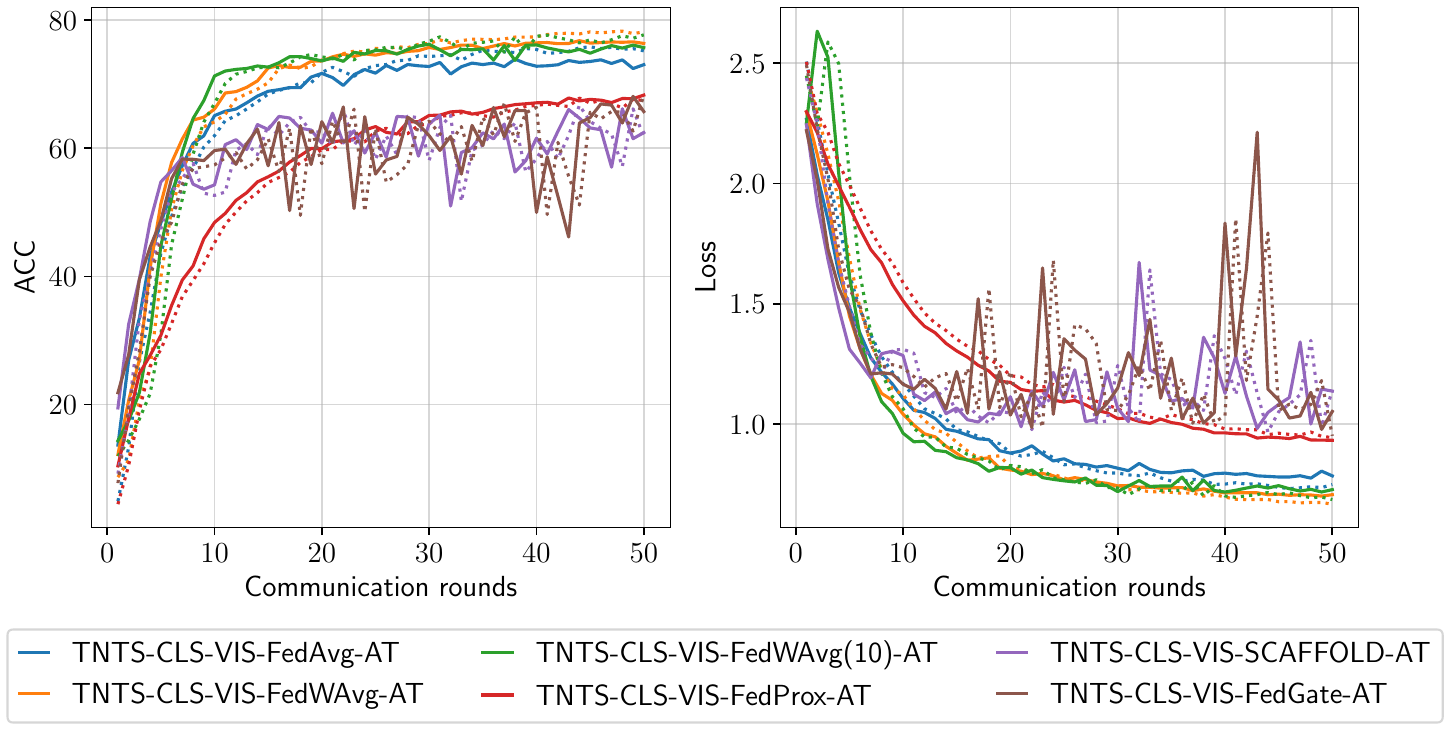}\label{fig:train_test_loss_tnt_cls_vis_niid4_nt}} \end{tabular} &
				\begin{tabular}{l}\subfloat[]{\includegraphics[width=0.5\textwidth]{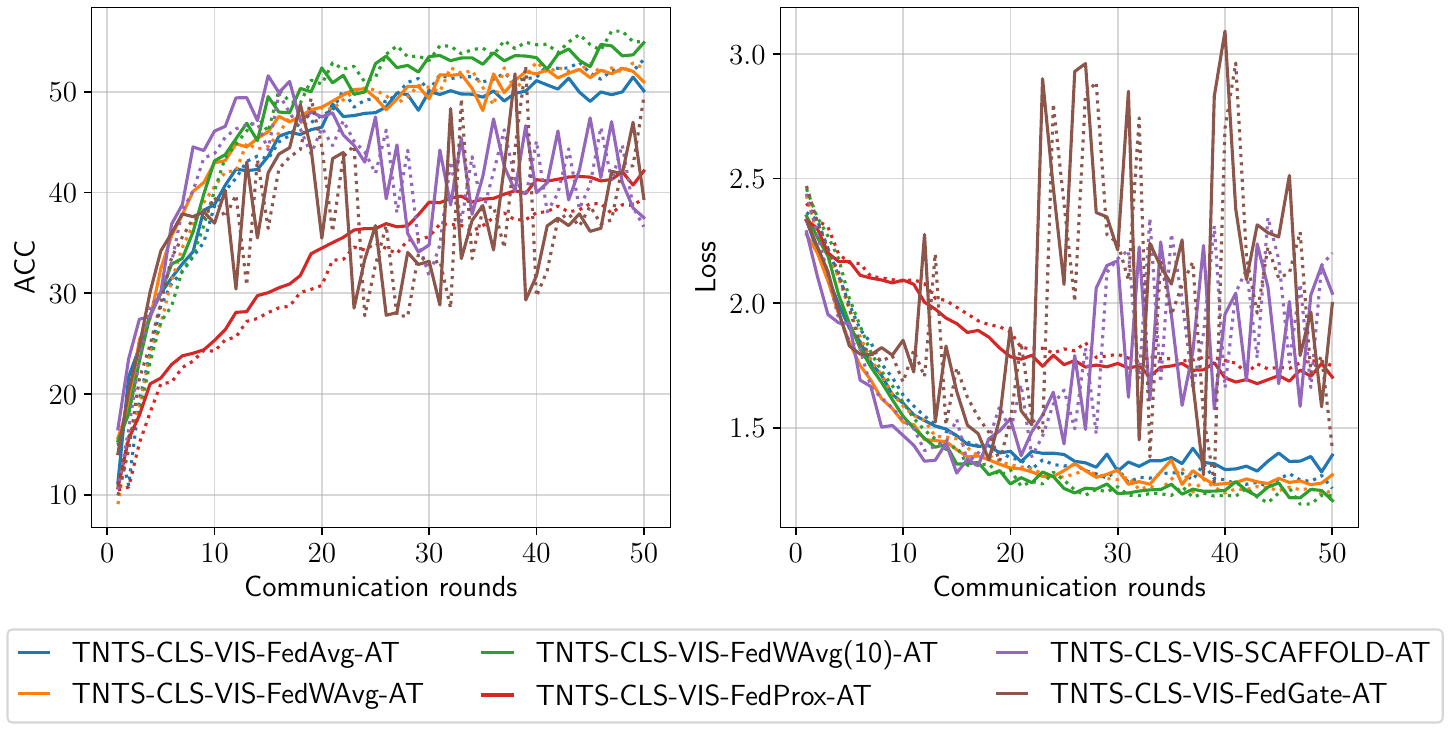}\label{fig:train_test_loss_tnt_cls_vis_niid2_nt}} \end{tabular} \\
			\end{tabular}
		}%
		\label{fig:train_test_loss_tnt_cls_vis}
	\end{figure*}

	\vspace{-1mm}
	More figures to show the convergence of FedWAvg and some of the tested \stateart are shown in \cref{fig:conv_vits_cls,fig:conv_t2t_cls,fig:conv_tnt_cls_vis}. Other figures that shows the convergence and the loss are shown in \cref{fig:conv_loss_tnt_cls_vis,fig:conv_loss_vits_cls,fig:conv_loss_t2t_cls}. Moreover, \cref{fig:train_test_loss_vit_cls,fig:train_test_loss_vit_cls,fig:train_test_loss_tnt_cls_vis} show the convergence during the training and the testing.

	\begin{figure*}[!htbp]
		\centering
		\caption{The accuracy in training (dashed) and testing (solid) of \ac{t2t}-VIS model with loss values in the \ac{fat} process for different aggregation methods. The accuracies against the communication rounds under the \ac{at} process using \ac{niid}(4), and \ac{niid}(2) are shown in a), and b) respectively. For better visualization, the dashed line is shifted one round forward.}
		\resizebox{\textwidth}{!}{%
			\setlength\tabcolsep{1.5pt}
			\begin{tabular}{ccc}
				\begin{tabular}{l}\subfloat[]{\includegraphics[width=0.5\textwidth]{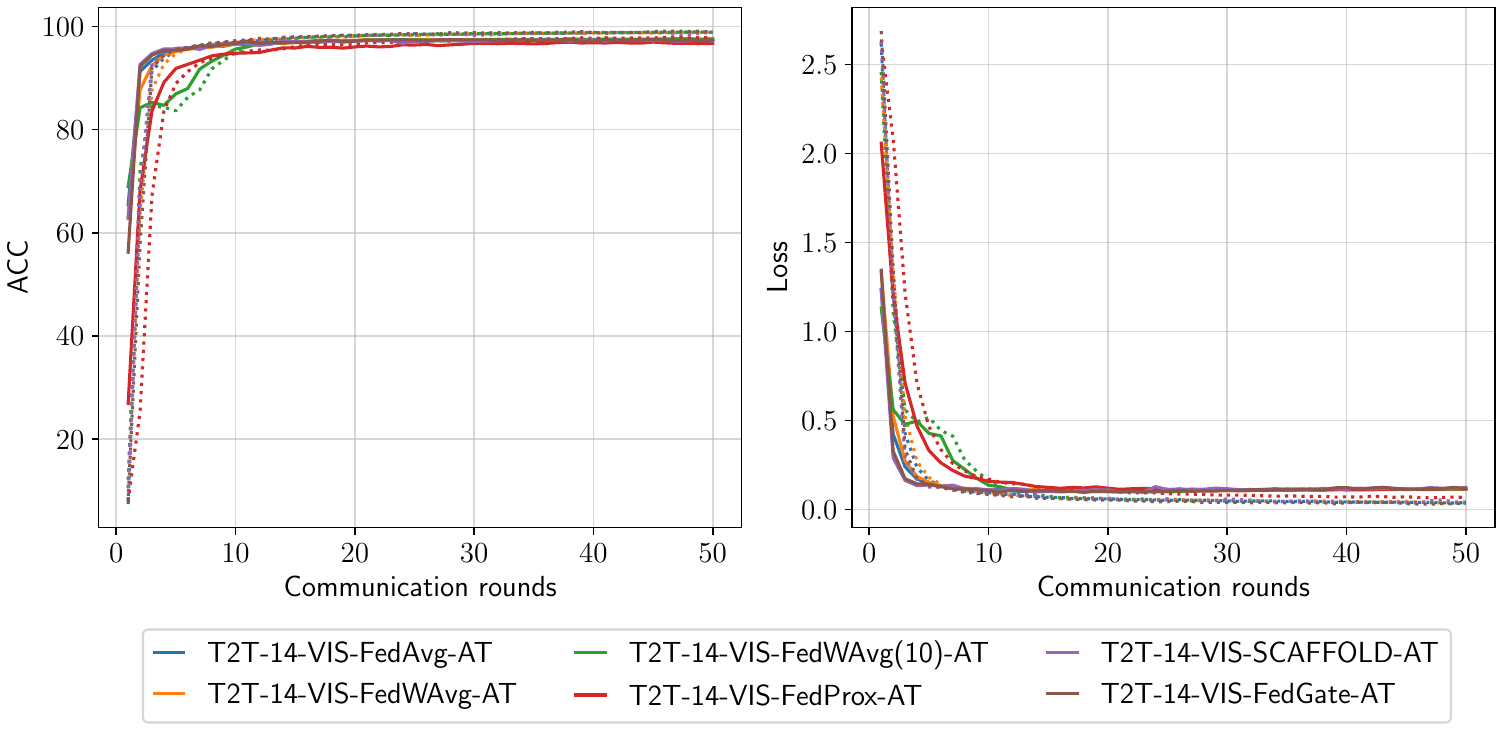}\label{fig:train_test_loss_t2t_vis_niid4_nt}} \end{tabular} &
				\begin{tabular}{l}\subfloat[]{\includegraphics[width=0.5\textwidth]{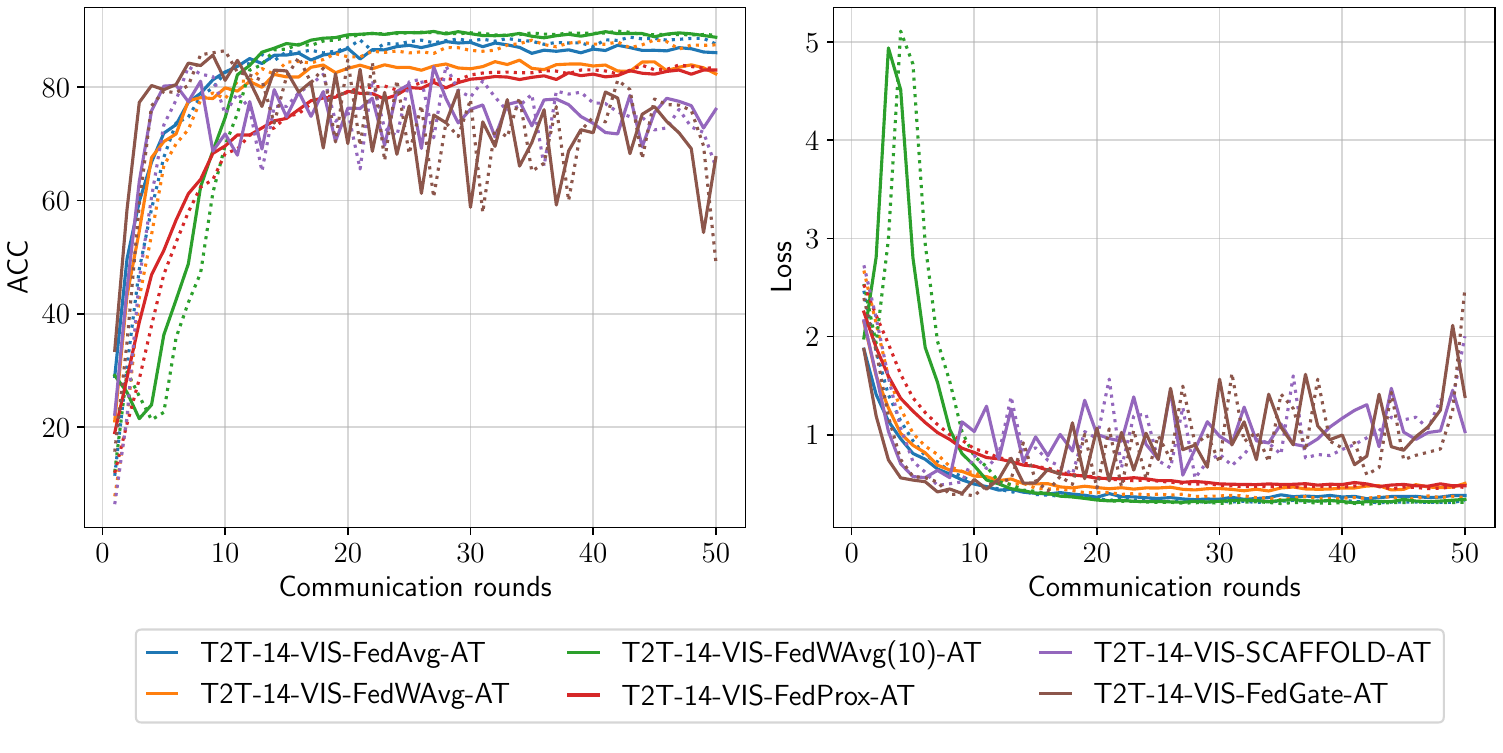}\label{fig:train_test_loss_t2t_vis_niid2_nt}} \end{tabular} \\
			\end{tabular}
		}%
		\label{fig:train_test_loss_t2t_vis}
	\end{figure*}
	
	\vspace{-5mm}
	\section{More figures for model drift}
	\label{app:more_model_drift}
\begin{figure*}[!t]
		\centering
		\caption{The \acs{svcca} for the first and the ninth layer of server, client 1, and client 4 models against communication rounds under \ac{fat} process using \ac{vit}-CLS model with \ac{niid}(4). a) using clean test samples and b) using adversarial test samples. The top row shows the \acs{svcca} between client 1 and client 2, the middle row shows the \acs{svcca} between the server and client 1, and the bottom row shows the \acs{svcca} between the server and client 4. }
		\resizebox{\textwidth}{!}{%
			\setlength\tabcolsep{1.5pt}
			\begin{tabular}{cc}
				\begin{tabular}{l}\subfloat[With clean samples]{\includegraphics[width=0.5\textwidth]{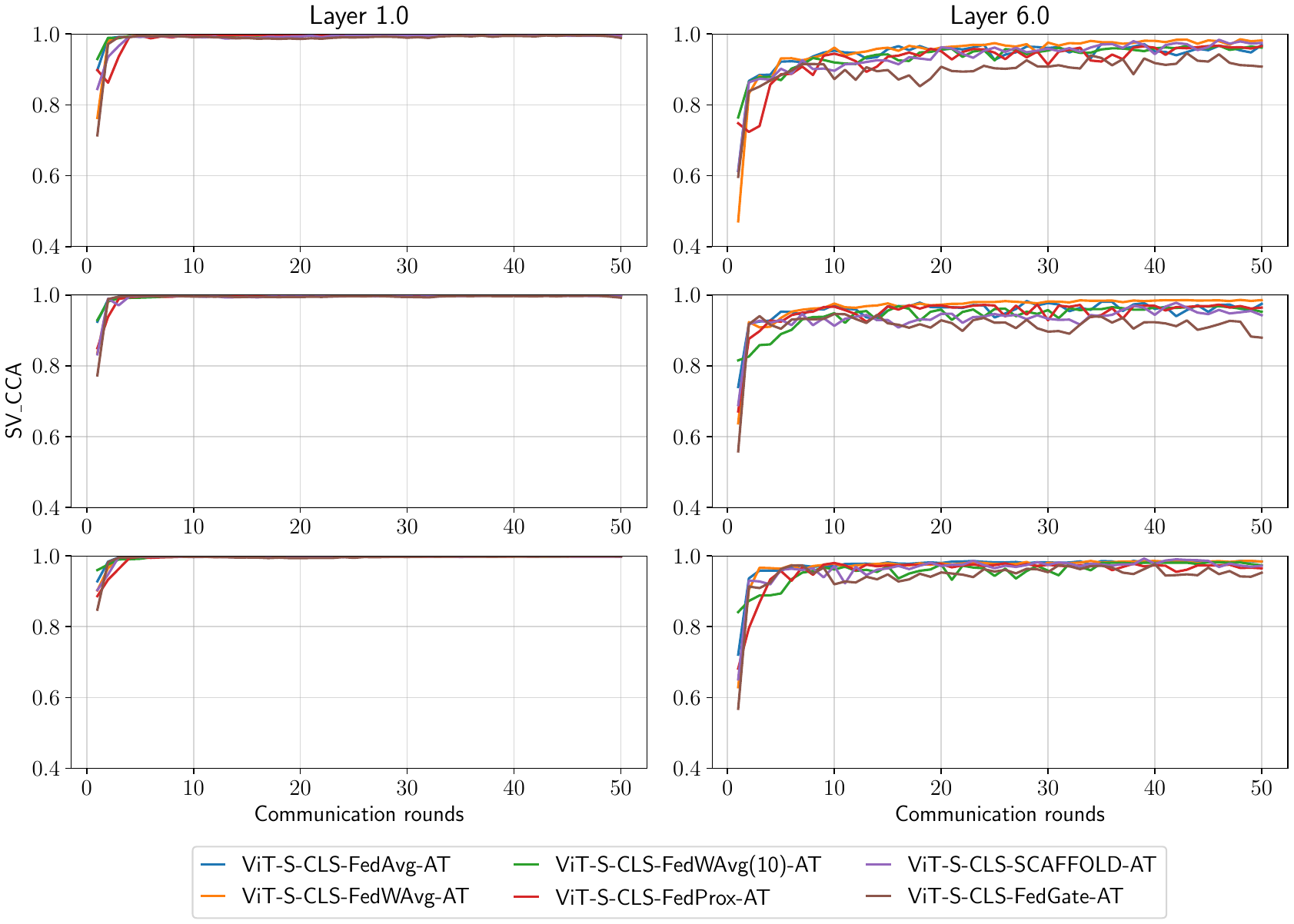}\label{fig:drift_vit_cls_iid4_at_clean}}\end{tabular} &
				\begin{tabular}{l}\subfloat[With adversarial samples]{\includegraphics[width=0.5\textwidth]{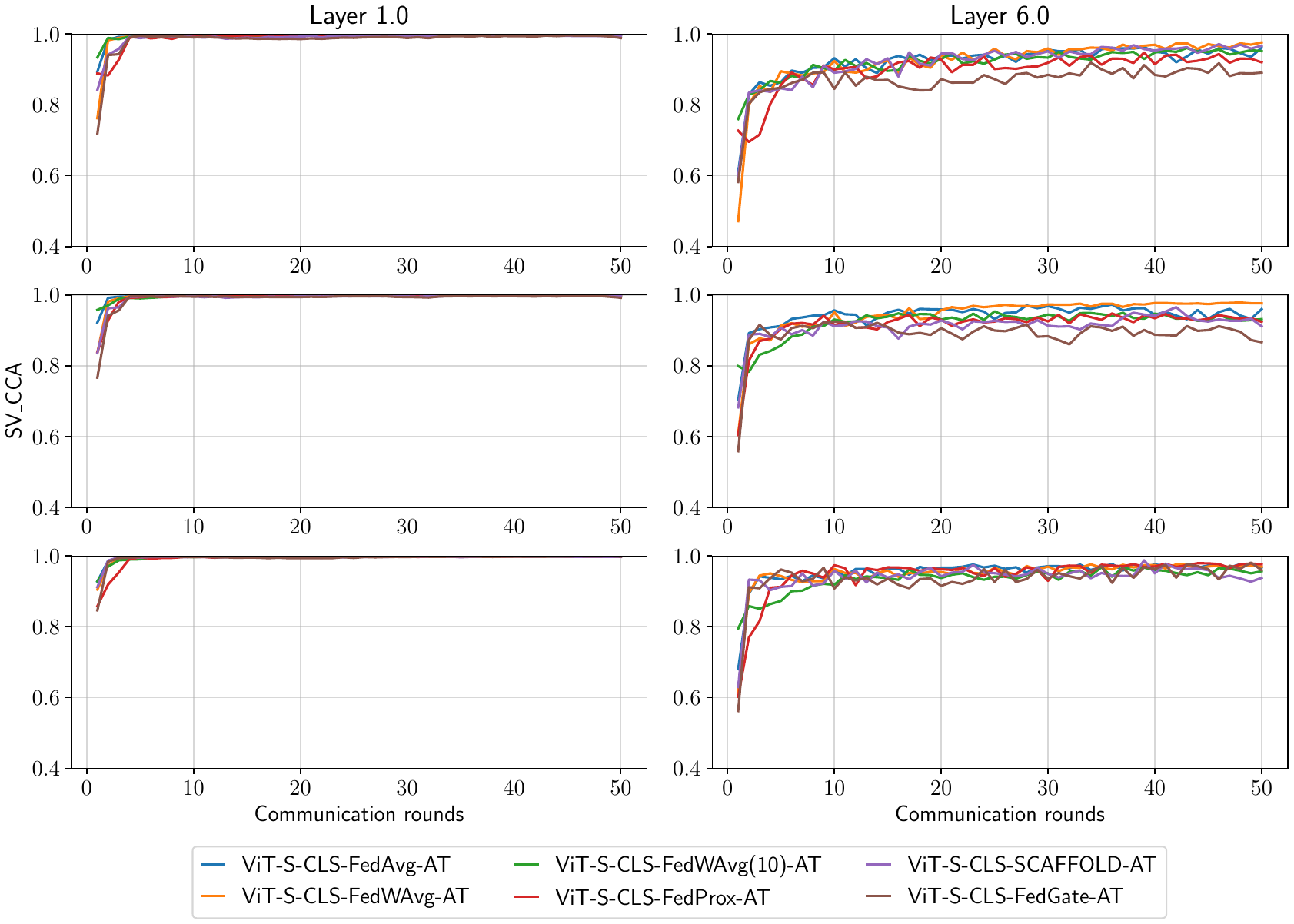}\label{fig:drift_vit_cls_iid4_at_adv}} \end{tabular} \\
			\end{tabular}
		}%

		\label{fig:drift_vit_cls_iid4_at}
	\end{figure*}

\begin{figure*}[!t]
		\centering
		\caption{The \acs{svcca} for the first and the ninth layer of server, client 1, and client 4 models against communication rounds under \ac{fat} process using \ac{vit}-CLS model with \ac{niid}(2). a) using clean test samples and b) using adversarial test samples. The top row shows the \acs{svcca} between client 1 and client 2, the middle row shows the \acs{svcca} between the server and client 1, and the bottom row shows the \acs{svcca} between the server and client 4. }
		\resizebox{\textwidth}{!}{%
			\setlength\tabcolsep{1.5pt}
			\begin{tabular}{cc}
				\begin{tabular}{l}\subfloat[With clean samples]{\includegraphics[width=0.5\textwidth]{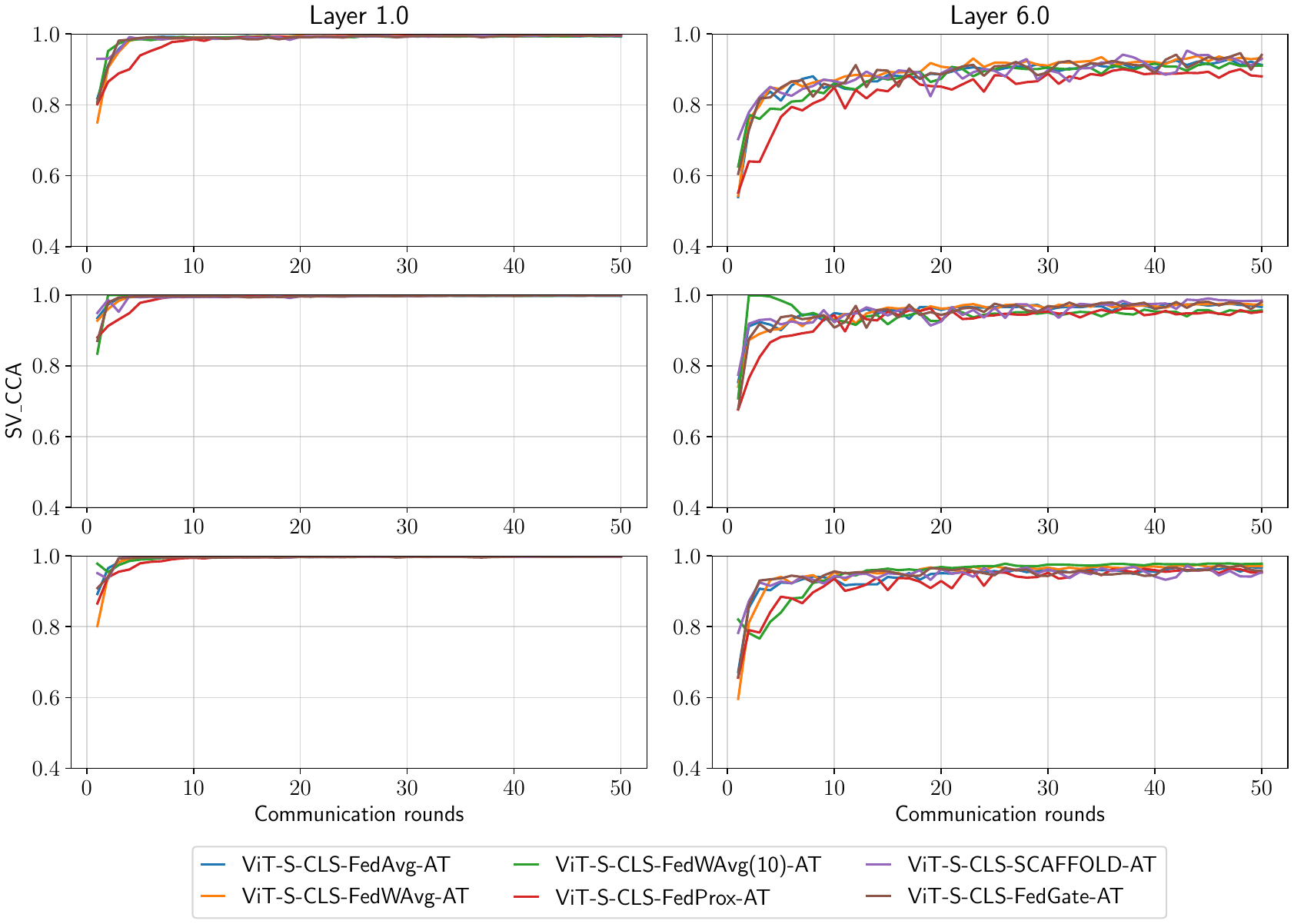}\label{fig:drift_vit_cls_iid2_at_clean}}\end{tabular} &
				\begin{tabular}{l}\subfloat[With adversarial samples]{\includegraphics[width=0.5\textwidth]{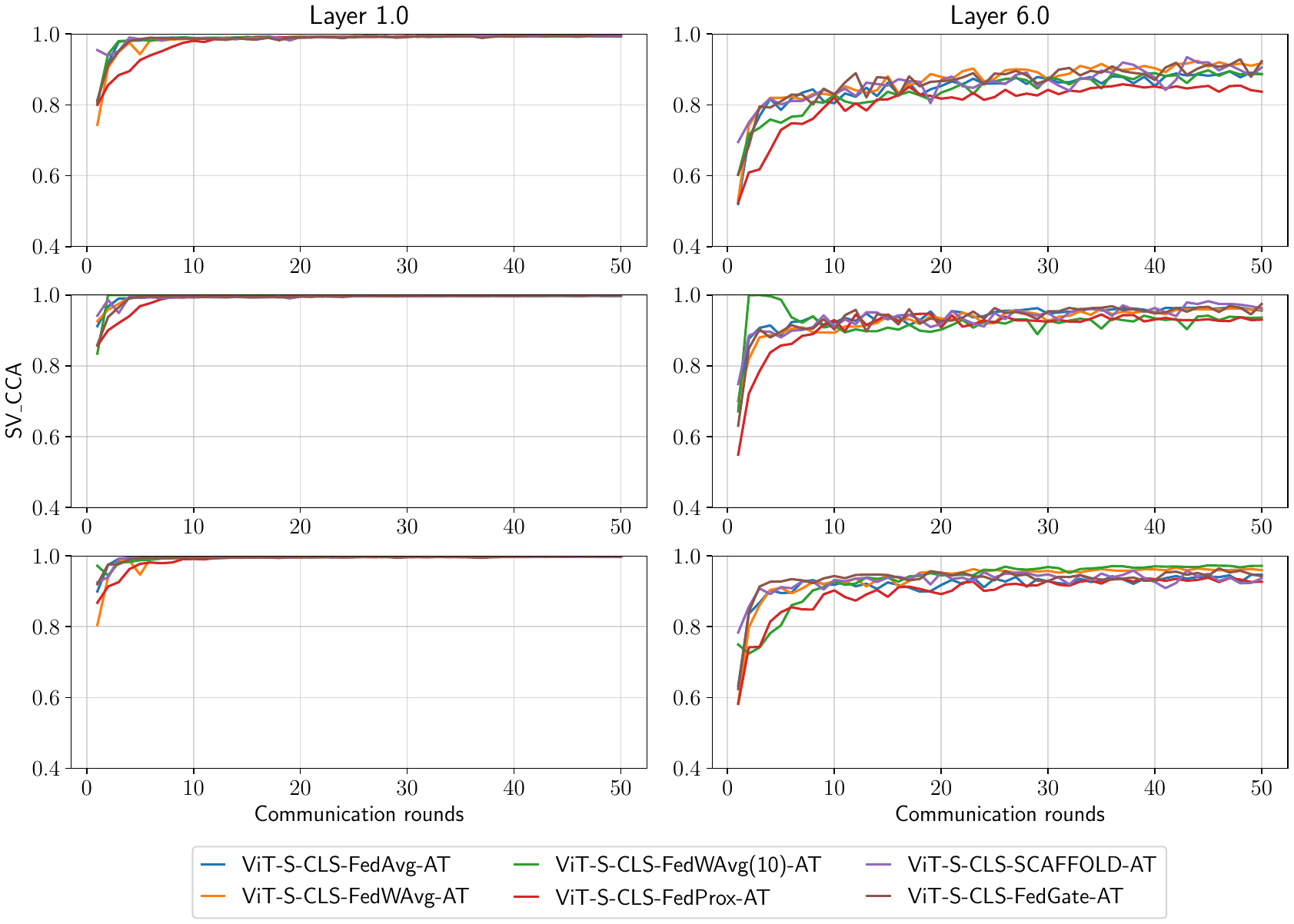}\label{fig:drift_vit_cls_iid2_at_adv}} \end{tabular} \\
			\end{tabular}
		}%

		\label{fig:drift_vit_cls_iid2_at}
	\end{figure*}

\begin{figure*}[!t]
		\centering
		\caption{The \acs{svcca} for the first and the ninth layer of server, client 1, and client 4 models against communication rounds under \ac{fat} process using \ac{t2t}-VIS model with \ac{niid}(4). a) using clean test samples and b) using adversarial test samples. The top row shows the \acs{svcca} between client 1 and client 2, the middle row shows the \acs{svcca} between the server and client 1, and the bottom row shows the \acs{svcca} between the server and client 4. }
		\resizebox{\textwidth}{!}{%
			\setlength\tabcolsep{1.5pt}
			\begin{tabular}{cc}
				\begin{tabular}{l}\subfloat[With clean samples]{\includegraphics[width=0.5\textwidth]{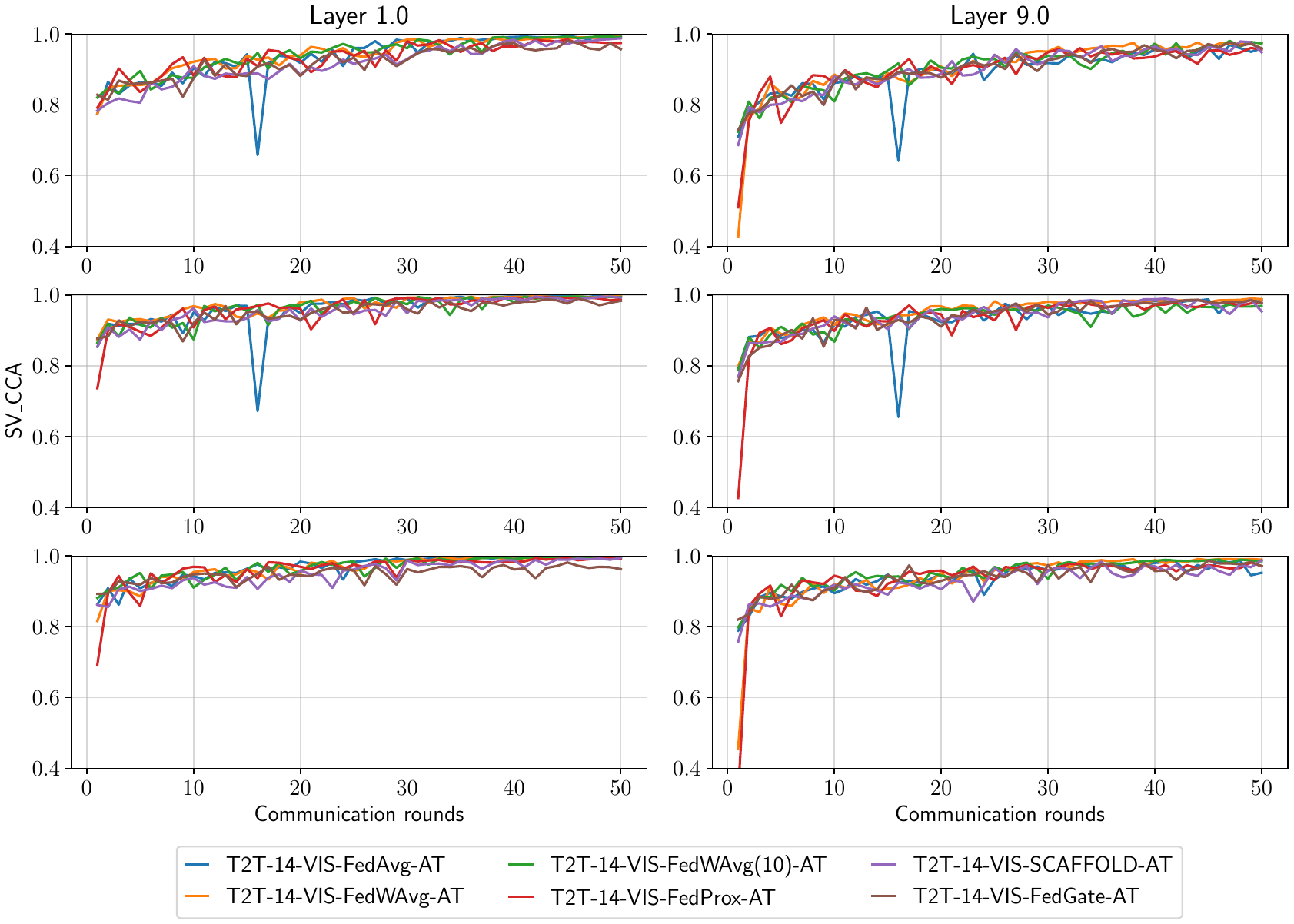}\label{fig:drift_t2t_vis_iid4_at_clean}}\end{tabular} &
				\begin{tabular}{l}\subfloat[With adversarial samples]{\includegraphics[width=0.5\textwidth]{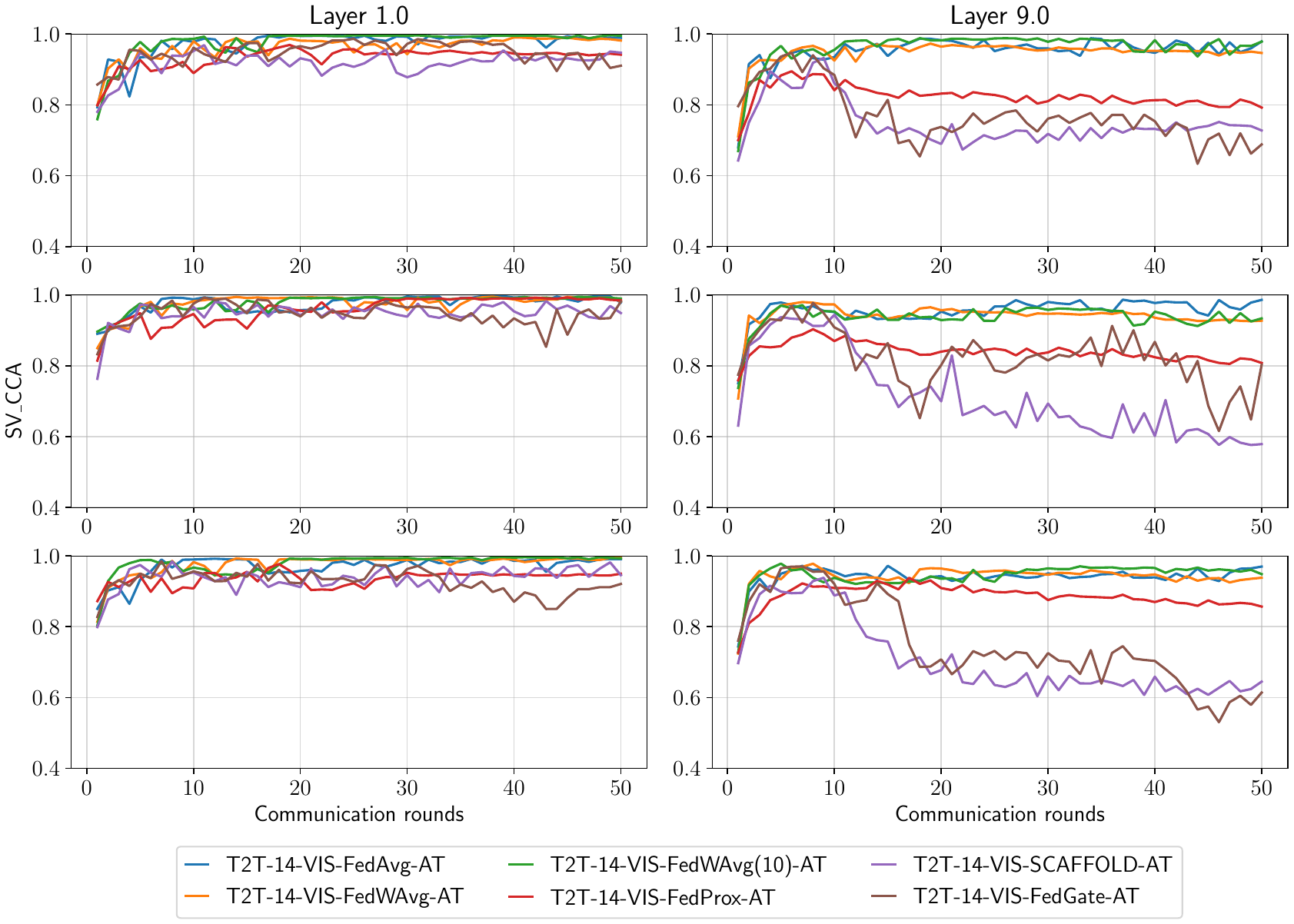}\label{fig:drift_t2t_vis_iid4_at_adv}} \end{tabular} \\
			\end{tabular}
		}%

		\label{fig:drift_t2t_vis_iid4_at}
	\end{figure*}

\begin{figure*}[!t]
		\centering
		\caption{The \acs{svcca} for the first and the ninth layer of server, client 1, and client 4 models against communication rounds under \ac{fat} process using \ac{t2t}-VIS model with \ac{niid}(2). a) using clean test samples and b) using adversarial test samples. The top row shows the \acs{svcca} between client 1 and client 2, the middle row shows the \acs{svcca} between the server and client 1, and the bottom row shows the \acs{svcca} between the server and client 4. }
		\resizebox{\textwidth}{!}{%
			\setlength\tabcolsep{1.5pt}
			\begin{tabular}{cc}
				\begin{tabular}{l}\subfloat[With clean samples]{\includegraphics[width=0.5\textwidth]{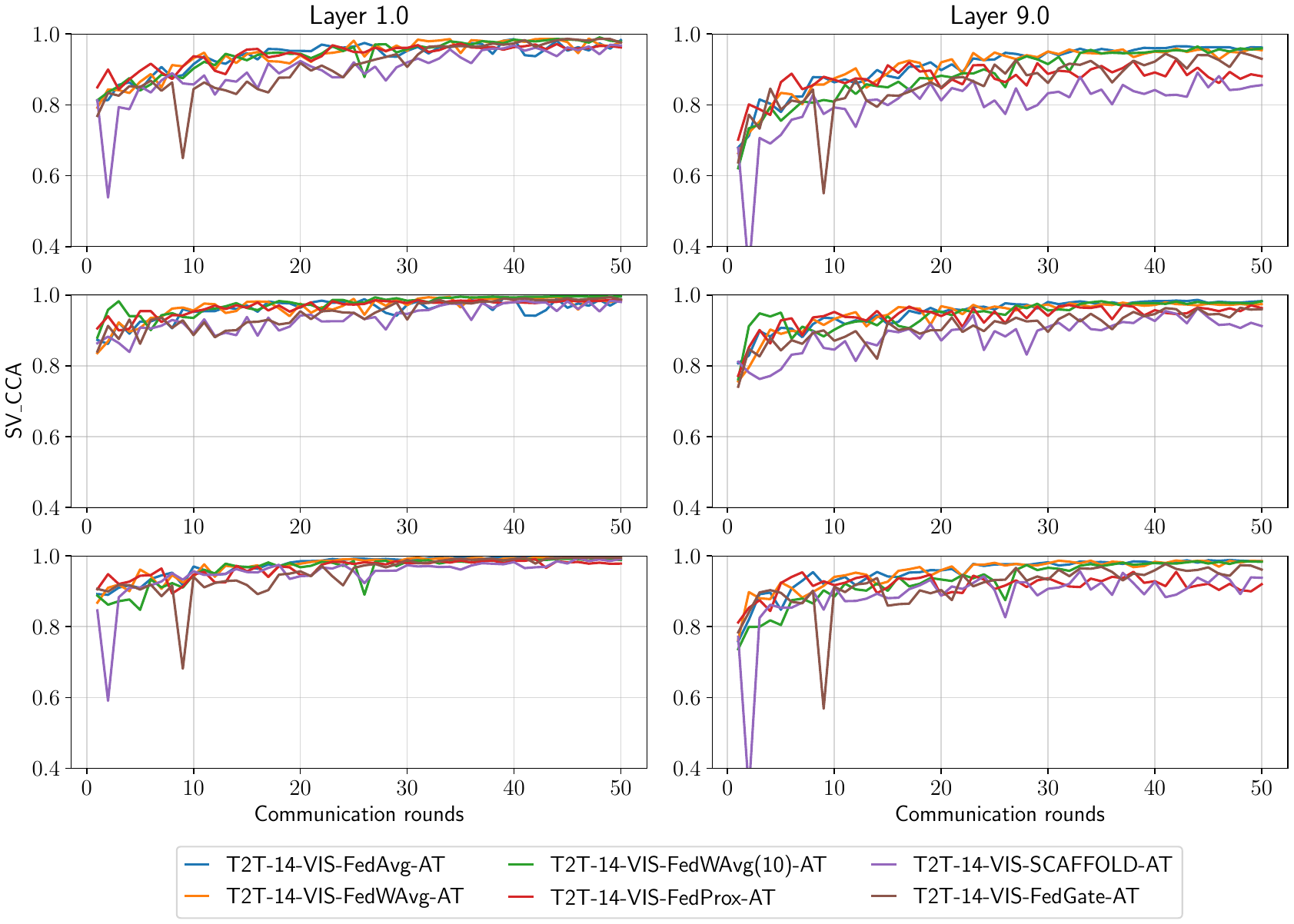}\label{fig:drift_t2t_vis_iid2_at_clean}}\end{tabular} &
				\begin{tabular}{l}\subfloat[With adversarial samples]{\includegraphics[width=0.5\textwidth]{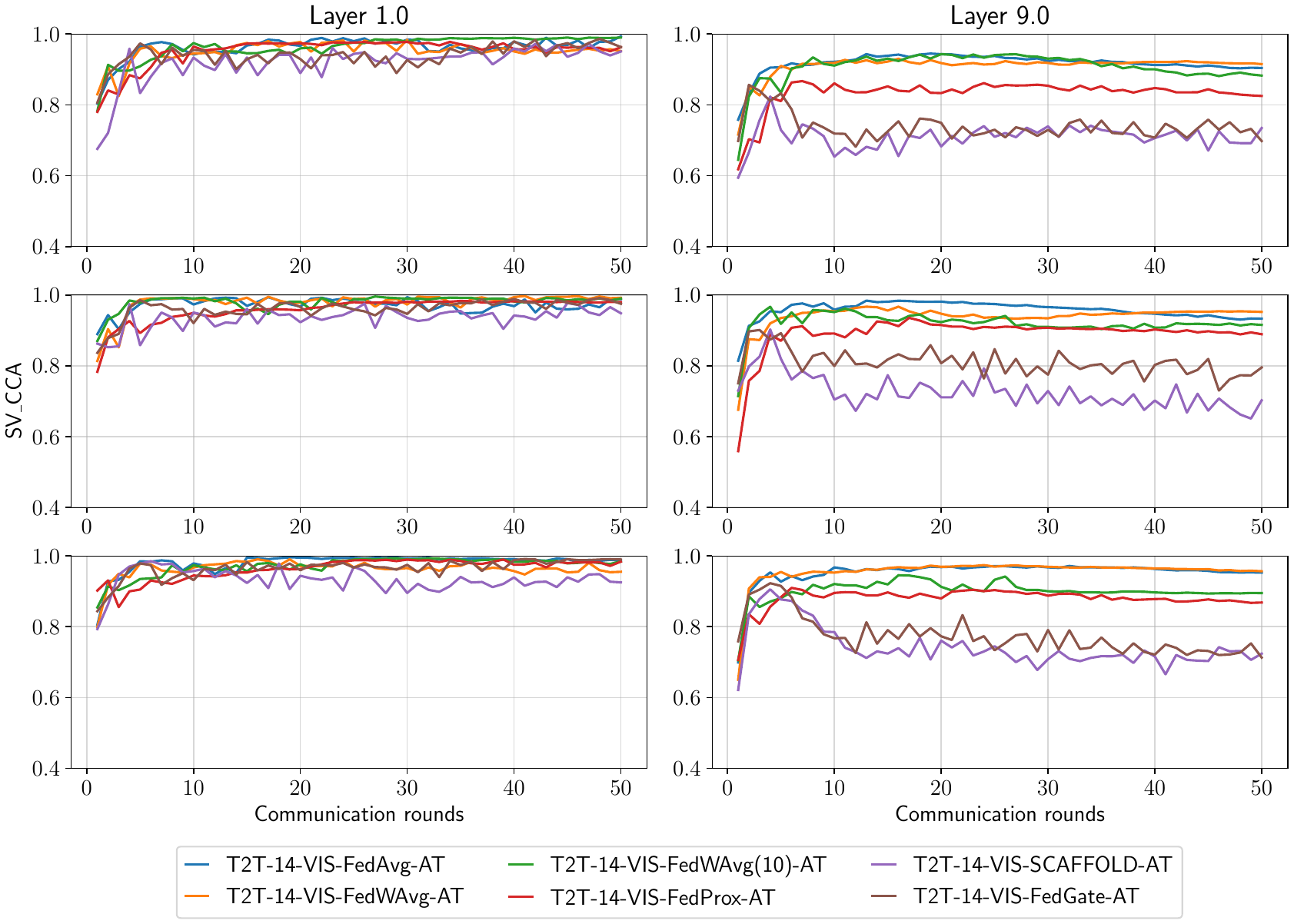}\label{fig:drift_t2t_vis_iid2_at_adv}} \end{tabular} \\
			\end{tabular}
		}%

		\label{fig:drift_t2t_vis_iid2_at}
	\end{figure*}

\begin{figure*}[!t]
		\centering
		\caption{The \acs{svcca} for the first and the ninth layer of server, client 1, and client 4 models against communication rounds under \ac{fat} process using \ac{tnt}-CLS-VIS model with \ac{niid}(4). a) using clean test samples and b) using adversarial test samples. The top row shows the \acs{svcca} between client 1 and client 2, the middle row shows the \acs{svcca} between the server and client 1, and the bottom row shows the \acs{svcca} between the server and client 4. }
		\resizebox{\textwidth}{!}{%
			\setlength\tabcolsep{1.5pt}
			\begin{tabular}{cc}
				\begin{tabular}{l}\subfloat[With clean samples]{\includegraphics[width=0.5\textwidth]{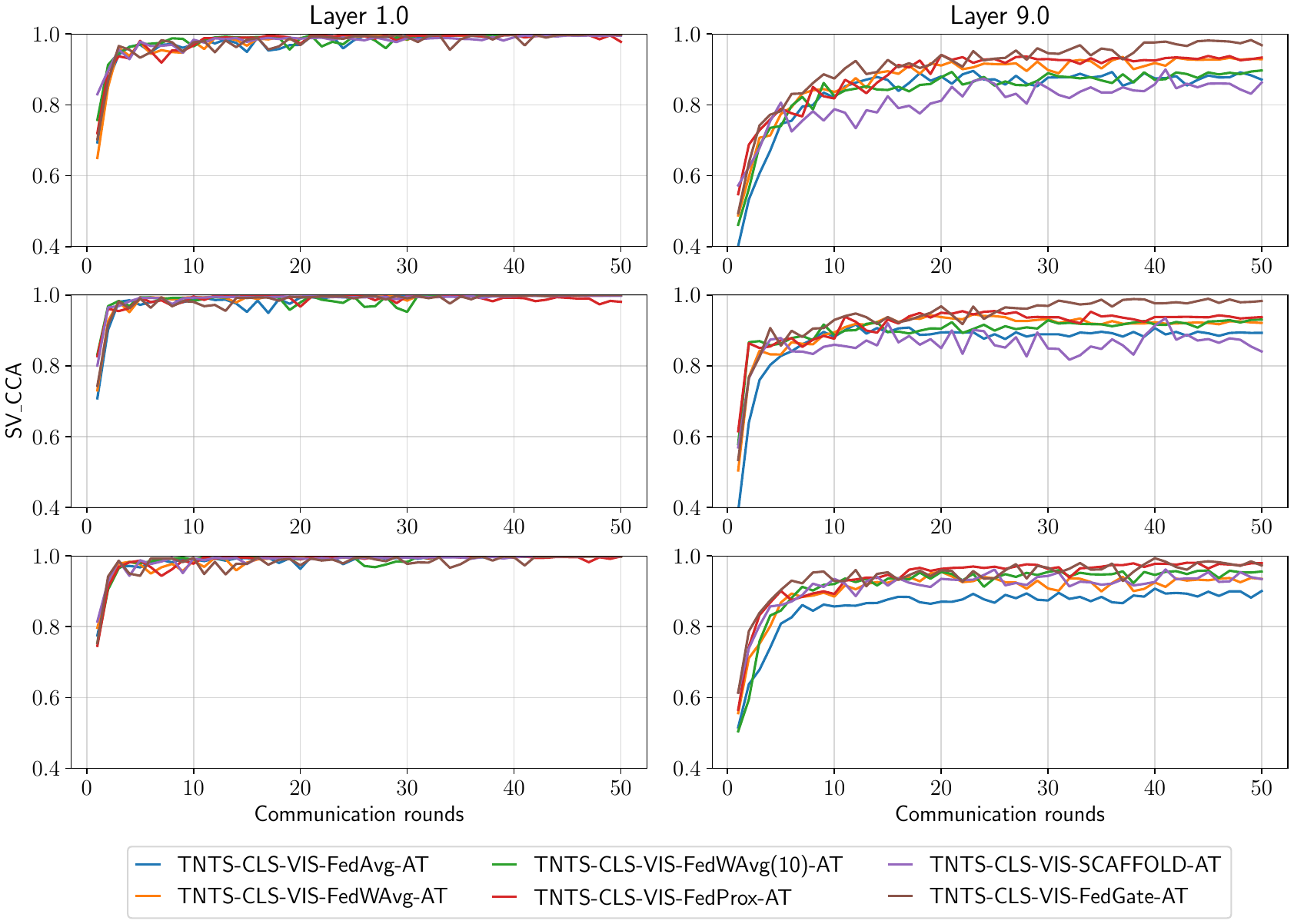}\label{fig:drift_tnt_cls_vis_iid4_at_clean}}\end{tabular} &
				\begin{tabular}{l}\subfloat[With adversarial samples]{\includegraphics[width=0.5\textwidth]{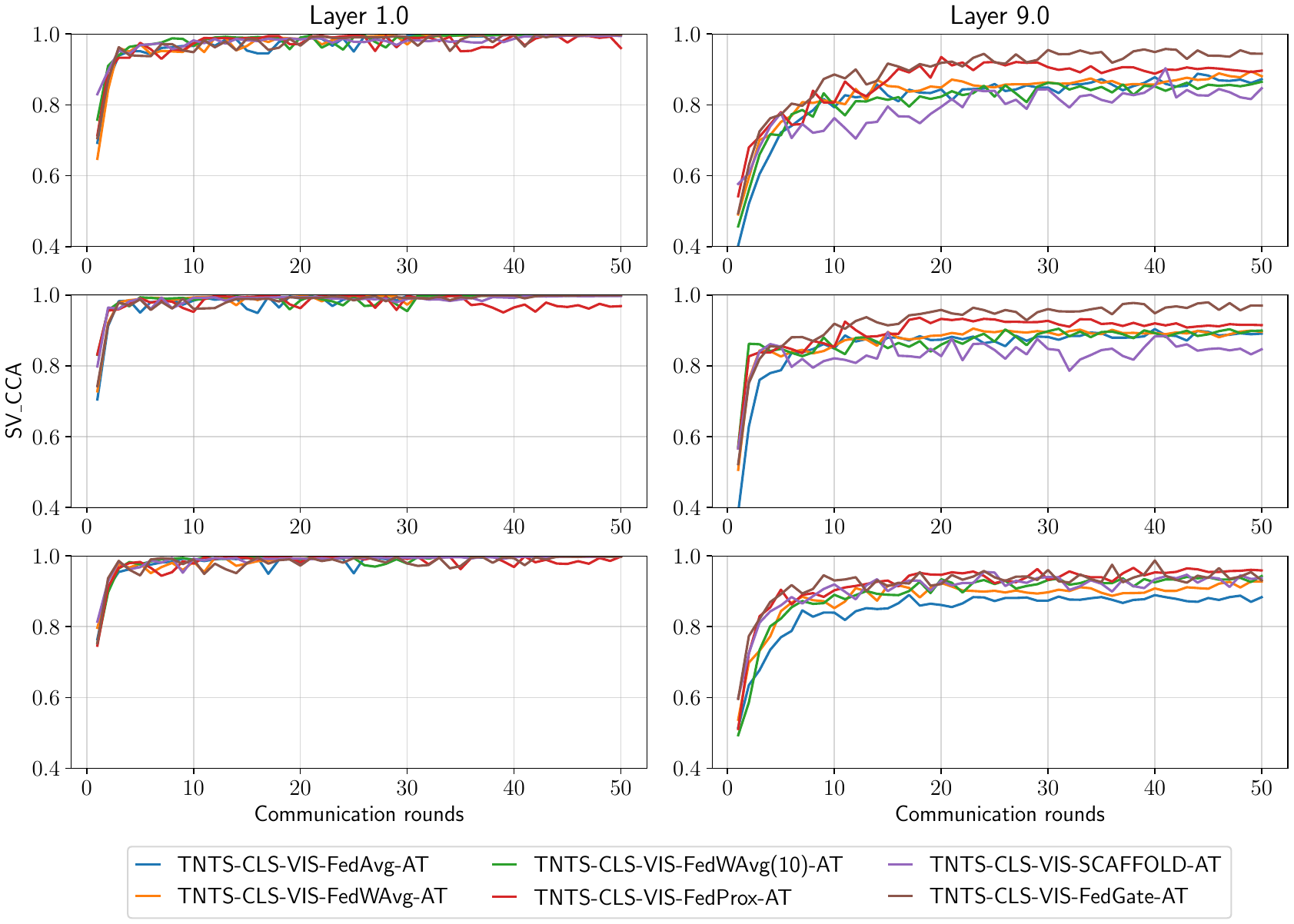}\label{fig:drift_tnt_cls_vis_iid4_at_adv}} \end{tabular} \\
			\end{tabular}
		}%

		\label{fig:drift_tnt_cls_vis_iid4_at}
	\end{figure*}

\begin{figure*}[!t]
		\centering
		\caption{The \acs{svcca} for the first and the ninth layer of server, client 1, and client 4 models against communication rounds under \ac{fat} process using \ac{tnt}-CLS-VIS model with \ac{niid}(2). a) using clean test samples and b) using adversarial test samples. The top row shows the \acs{svcca} between client 1 and client 2, the middle row shows the \acs{svcca} between the server and client 1, and the bottom row shows the \acs{svcca} between the server and client 4. }
		\resizebox{\textwidth}{!}{%
			\setlength\tabcolsep{1.5pt}
			\begin{tabular}{cc}
				\begin{tabular}{l}\subfloat[With clean samples]{\includegraphics[width=0.5\textwidth]{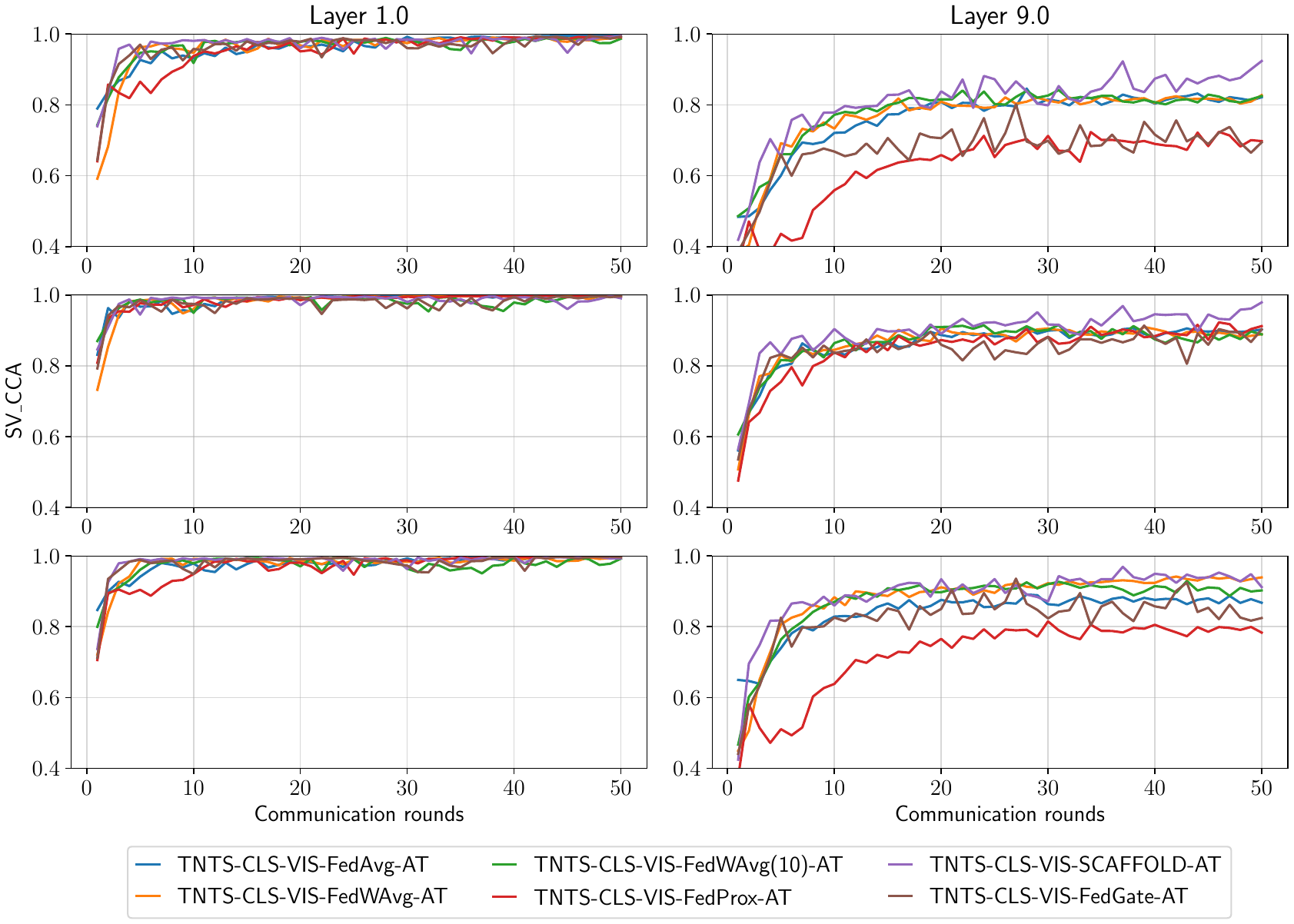}\label{fig:drift_tnt_cls_vis_iid2_at_clean}}\end{tabular} &
				\begin{tabular}{l}\subfloat[With adversarial samples]{\includegraphics[width=0.5\textwidth]{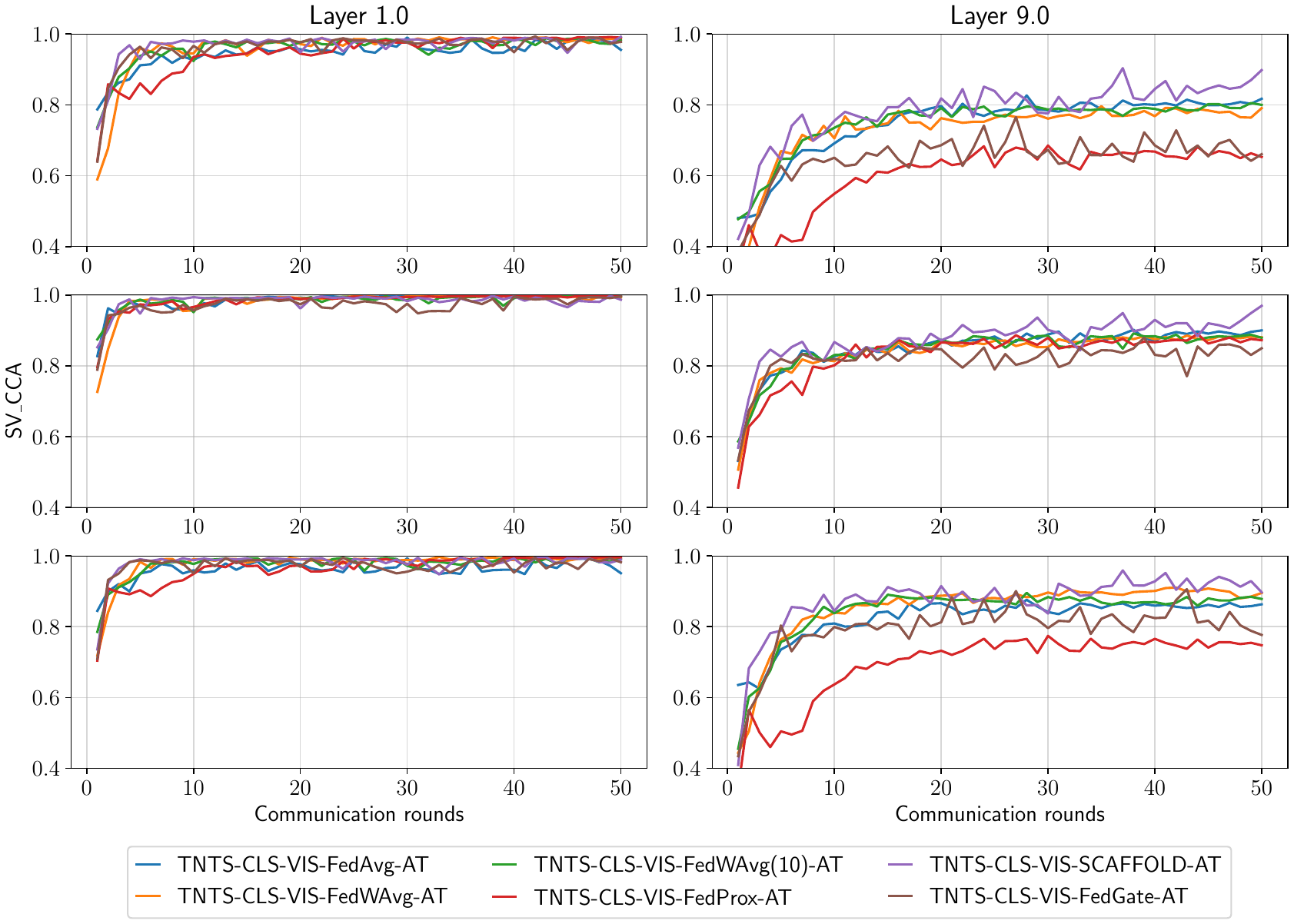}\label{fig:drift_tnt_cls_vis_iid2_at_adv}} \end{tabular} \\
			\end{tabular}
		}%

		\label{fig:drift_tnt_cls_vis_iid2_at}
	\end{figure*}

	\vspace{-1mm}
	\cref{fig:drift_vit_cls_iid4_at,fig:drift_vit_cls_iid2_at,fig:drift_t2t_vis_iid4_at,fig:drift_t2t_vis_iid2_at,fig:drift_tnt_cls_vis_iid4_at,fig:drift_tnt_cls_vis_iid2_at} show more model drift for some of the tested models. 
	
	\vspace{-5mm}
	\section{Notation}
	\label{app:notation}
	
	\begin{table}[!h]
		\centering
		\vspace{-3mm}
		\caption{Notation used throughout the paper.}
		\label{tab:notation}
		\resizebox{\linewidth}{!}{%
			\begin{tabular}{ll}
				\hline
				Symbol & Meaning \\
				\hline
				$K$, $k$ & number of clients; client index \\
				$C$, $m$, $S_t$ & client fraction; $m=\max(1,CK)$ sampled clients; sampled set at round $t$ \\
				$T$, $t$ & number of communication rounds; round index \\
				$E$, $\tau$ & local epochs per round; local SGD steps per round \\
				$B$, $b$ & local minibatch size; a minibatch \\
				$n$, $n_k$, $p_k$ & global and local sample counts; $p_k=n_k/n$ \\
				$D$, $D_k$ & global and local datasets \\
				$\mathcal{D}$, $\mathcal{D}_k$ & global and local data distributions \\
				$\theta_t$, $\theta_t^k$ & global and client-$k$ model parameters at round $t$ \\
				$\eta$ & learning rate \\
				$\mathcal{L}$, $F$, $F_k$ & loss; global and local robust risk, Eq.~\eqref{eq:fatobj} \\
				$F_w$ & $w$-weighted objective, $\sum_k w_k F_k$ \\
				$r$, $n_{adv}$ & adversarial ratio; adversarial samples per batch, $n_{adv}=rB$ \\
				$\mathcal{A}_{s,\varepsilon,\alpha}$ & PGD attack with $s$ steps, budget $\varepsilon$, step size $\alpha$ \\
				$\varepsilon_0$, $\zeta$ & perturbation budget; PGD inexactness (Assumption~\ref{as:inner}) \\
				$g$, $l_k$ & last-layer parameters of the global model and of client $k$ \\
				$\hat g$, $\hat l_k$ & the same, normalised to unit $\ell_2$ norm \\
				$c_k$, $w_k$, $q$ & cosine similarity; aggregation weight; softmax scale factor \\
				$d_k$, $\bar d(q)$ & directional drift, Eq.~\eqref{eq:drift}; its $w(q)$-weighted mean \\
				$\Delta_k$, $\Delta_k^{\perp}$ & client head update; its component orthogonal to $g$ \\
				$\Delta_c$ & cosine spread, $\max_k c_k - \min_k c_k$ \\
				$d_L$ & last-layer dimension \\
				$L$, $\sigma^2$, $\kappa^2$ & smoothness constant; gradient variance; gradient dissimilarity \\
				$u$ & uniform weighting, $(1/K)\mathbf{1}$ \\
				$K_{\mathrm{eff}}$ & effective client count, $1/\|w\|_2^2$ \\
				\hline
			\end{tabular}
		}%
	\end{table}
	
\end{document}